\documentclass[10pt,twocolumn,letterpaper]{article}

\usepackage[pagenumbers]{cvpr}

\definecolor{cvprblue}{rgb}{0.21,0.49,0.74}
\usepackage[pagebackref,breaklinks,colorlinks,linkcolor=blue,citecolor=blue]{hyperref}

\usepackage{todonotes}
\usepackage{subcaption}
\usepackage[percent]{overpic}  \newcommand{\myPara}[1]{\noindent\textbf{#1}}

\usepackage{tikz}
\usepackage{multirow}

\usepackage{siunitx} \usepackage{colortbl}
\usepackage{bm,bbm}

\usepackage[ruled,vlined,linesnumbered]{algorithm2e}

\SetKwInput{KwInit}{Initialize}
\SetKwInput{KwUpdate}{Update}
\SetKwInput{KwStage}{Stage}
\SetKwComment{Comment}{$\triangleright$\ }{}

\usepackage{makecell}

\newcommand\blfootnote[1]{\begingroup
  \renewcommand\thefootnote{}\footnote{#1}\addtocounter{footnote}{-1}\endgroup
}

\def\paperID{*****} \def\confName{CVPR}
\def\confYear{2026}

\title{
BriMA: Bridged Modality Adaptation \\ for Multi-Modal Continual Action Quality Assessment
}

\author{
Kanglei Zhou\textsuperscript{1},~
Chang Li\textsuperscript{1},~
Qingyi Pan\textsuperscript{2},~
Liyuan Wang\textsuperscript{1,$\dagger$} \\
{\normalsize
\textsuperscript{1} Department of Psychological and Cognitive Sciences, 
\textsuperscript{2} Department of Statistics and Data Science, Tsinghua University
}\\
}

\begin{document}
\maketitle

\begin{abstract}
\blfootnote{$^\dagger$ Corresponding author: {\tt liyuanwang@tsinghua.edu.cn}}Action Quality Assessment (AQA) aims to score how well an action is performed and is widely used in sports analysis, rehabilitation assessment, and human skill evaluation. Multi-modal AQA has recently achieved strong progress by leveraging complementary visual and kinematic cues, yet real-world deployments often suffer from non-stationary modality imbalance, where certain modalities become missing or intermittently available due to sensor failures or annotation gaps. Existing continual AQA methods overlook this issue and assume that all modalities remain complete and stable throughout training, which restricts their practicality. To address this challenge, we introduce \textbf{Bri}dged \textbf{M}odality \textbf{A}daptation (BriMA), an innovative approach to multi-modal continual AQA under modality-missing conditions. BriMA consists of a memory-guided bridging imputation module that reconstructs missing modalities using both task-agnostic and task-specific representations, and a modality-aware replay mechanism that prioritizes informative samples based on modality distortion and distribution drift.  
Experiments on three representative multi-modal AQA datasets (RG, Fis-V, and FS1000) show that BriMA consistently improves performance under different modality-missing conditions, achieving 6--8\% higher correlation and 12--15\% lower error on average.  
These results demonstrate a step toward robust multi-modal AQA systems under real-world deployment constraints.
Our code is available at \url{https://github.com/ZhouKanglei/BriMA}.
\end{abstract} \section{Introduction}
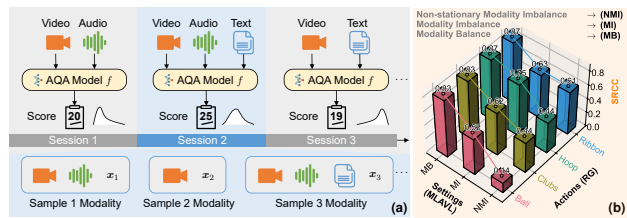
\begin{figure}[t]
    \centering
    \begin{overpic}[height=0.335\linewidth,trim=200 138 250 138,clip]{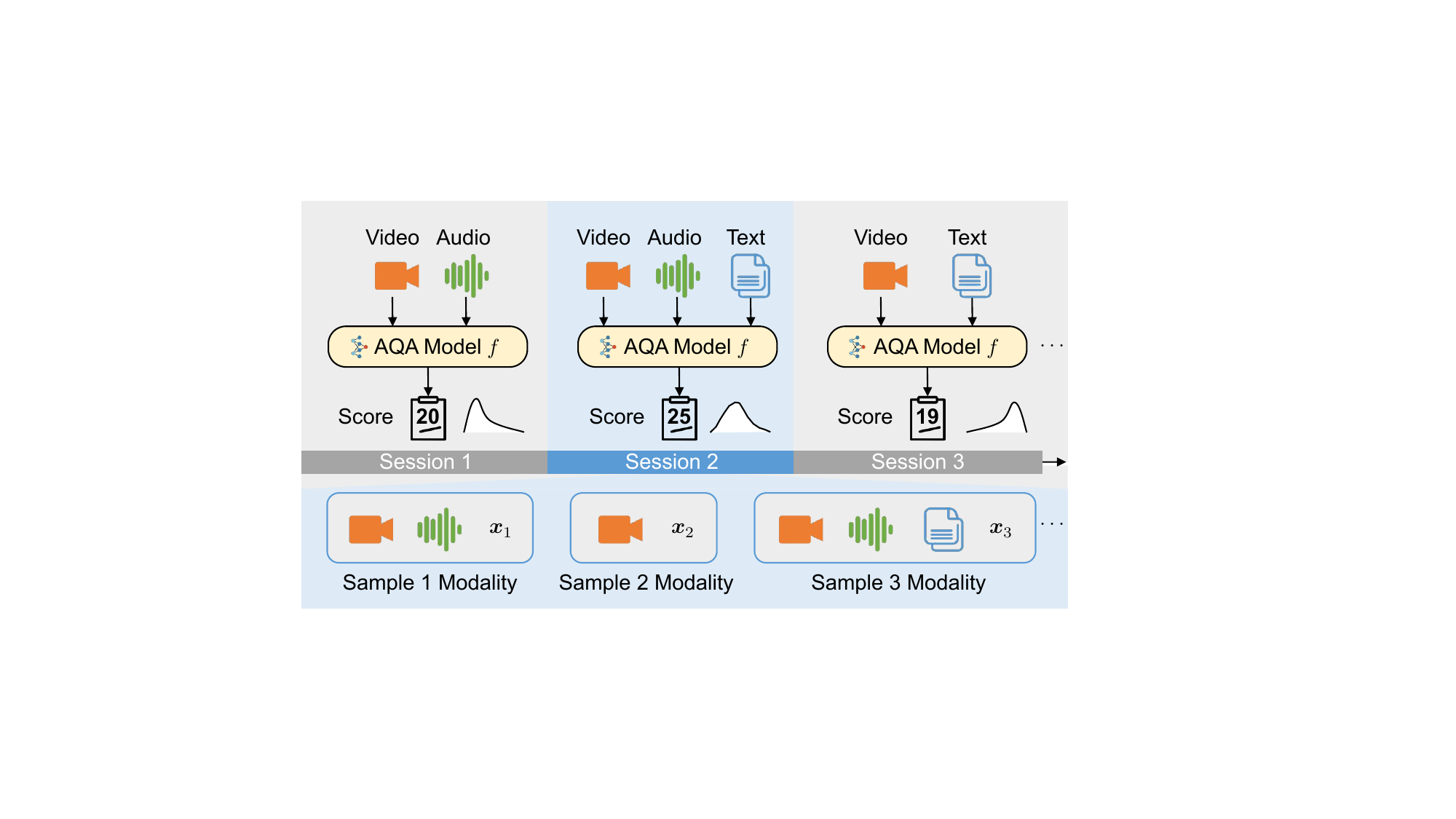}\put(92,0){
            \begin{tikzpicture}
                \node[fill=white, fill opacity=0, text opacity=1, inner sep=1pt] {\tiny\sf\textbf{(a)}};
            \end{tikzpicture}
        }\put(145.5,0){
            \begin{tikzpicture}
                \node[fill=white, fill opacity=0, text opacity=1, inner sep=1pt] {\tiny\sf\textbf{(b)}};
            \end{tikzpicture}
        }\end{overpic}\hspace{-0.1cm}\begin{overpic}[height=0.335\linewidth,clip,trim=10 0 10 0]{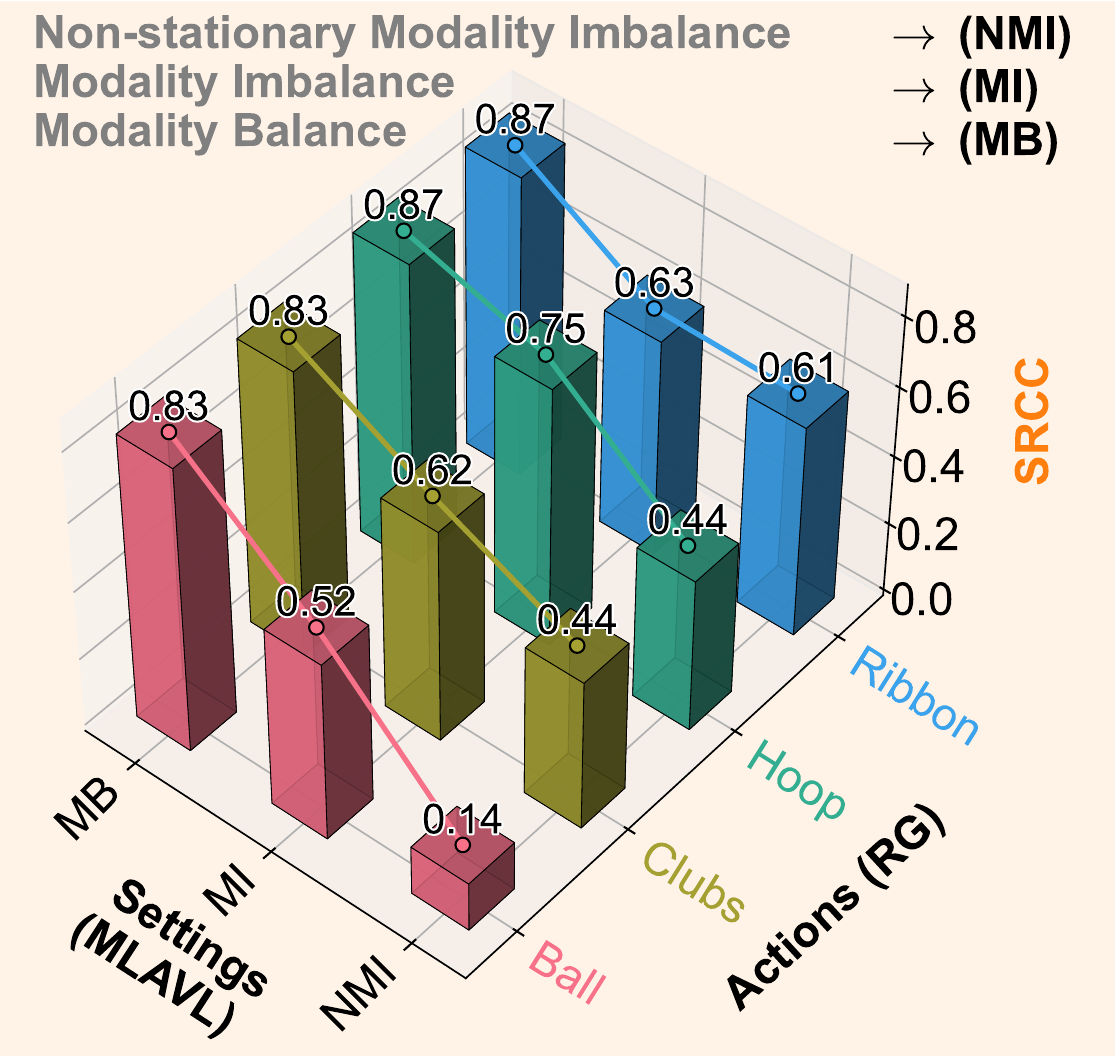}\end{overpic}
    \caption{Our motivation: non-stationary modality imbalance \textbf{(a)} significantly challenges continual AQA performance \textbf{(b)}.}
    \label{fig:teaser}
    \phantomsubcaption\label{fig:teaser-a}
    \phantomsubcaption\label{fig:teaser-b}
    \vspace{-0.5cm}
\end{figure}

Action Quality Assessment (AQA) \cite{wang2021tsa,wang2020assessing,parmar2022domain,zhou2024comprehensivesurveyactionquality,han2025caflow,han2025finecausal} aims to measure how well an action is performed, supporting applications in sports analysis \cite{xu2022finediving,xu2024fineparser}, rehabilitation assessment \cite{zhou2023video,bruce2024egcn++}, and skill evaluation \cite{huang2024egoexolearn,pan2025basket}. With the rise of multi-modal sensing, recent AQA models \cite{xia2023skating,xu2024vision,xu2025language,xu2025mcmoe} increasingly combine visual and other cues such as audio and motion flow, achieving notable progress in accuracy. However, real-world AQA rarely operates under fully reliable sensing conditions \cite{li2024toward,zeng2024missing,wu2024deep}. Cameras drop frames, sensors degrade over time, annotations are often missing, etc., leading to \textbf{non-stationary modality imbalance}, where modality availability varies over time (see \cref{fig:teaser-a}). This phenomenon naturally arises in practical AQA workflows and poses a significant obstacle to stable scoring.

Despite its practical prevalence, our empirical study reveals that current AQA efforts fail under such changing modality availability. Multi-modal AQA methods \cite{xu2025language,xu2024vision} assume fully observed inputs and therefore suffer noticeable failures once modalities become missing (supported by the performance decline in \cref{fig:teaser-b} and Appendix \cref{fig:supp-teaser}). Meanwhile, continual AQA approaches \cite{zhou2024magr,li2024continual,dadashzadeh2024pecop} mitigate task-level forgetting but also rely on complete and stable modalities during training, resulting in notable degradation under evolving modality imbalance (see the suboptimal performance of Fs-Aug \cite{li2024continual} and MAGR \cite{zhou2024magr} in \cref{tab:rg,tab:fisv,tab:fs1000}). These failures show that non-stationary modality imbalance is not a single missing-data issue. In AQA, modality changes reshape both the feature space and its temporal dynamics, forcing the model to cope with simultaneous input sparsity and shifting task distributions. This interaction destabilizes representation learning and amplifies forgetting, ultimately degrading scoring reliability.

The inherent complexity of AQA makes this problem fundamentally different from related challenges in recognition or general multi-modal learning \cite{wang2025hide,wang2022modality,zhou2024uncertainty}. AQA requires fine-grained score reasoning, where subtle temporal cues directly determine the score \cite{xu2025language,xu2024vision}. This tight coupling between modalities and score semantics breaks the assumptions underlying standard completion techniques. Simple imputation \cite{wu2024deep} introduces bias, retrieval-based reconstruction \cite{wang2022modality,lang2025retrieval} pulls mismatched contexts, and generative synthesis \cite{ma2021smil,du2018semi} fails to preserve score-critical geometry under limited supervision. While these methods can restore missing modalities at the representation level, they distort the scoring manifold and undermine the ranking consistency essential for AQA. Consequently, directly applying these techniques to multi-modal continual AQA  yields only suboptimal performance (see \cref{tab:ablation,fig:forgetting}), highlighting the need for a tailored solution that explicitly respects the score-sensitive nature of this task.

To address these challenges, we propose \textbf{Bri}dged \textbf{M}odality \textbf{A}daptation (BriMA), an innovative approach to handling evolving modality imbalance and distribution shifts in multi-modal continual AQA. Our core idea is to construct a stable bridging space that aligns missing modalities with shared task structure and memory, enabling reliable modality completion and drift-aware adaptation that existing imputation, retrieval, and generative strategies cannot provide. BriMA is instantiated with two components. The first is a memory-guided bridging module that retrieves structurally aligned exemplars from past tasks and predicts only a minimal residual correction instead of full-feature synthesis, ensuring stable and score-faithful modality completion. The second is a modality-aware replay module that maintains a curated buffer of representative samples with reliable modalities and balanced score coverage, and dynamically prioritizes replay by modality distortion and score drift, effectively counteracting distribution shifts over time.

Experiments on three representative multi-modal AQA datasets (RG \cite{zeng2020hybrid}, Fis-V \cite{parmar2017learning}, and FS1000 \cite{xia2023skating}) show that BriMA consistently improves scoring accuracy and mitigates forgetting under heterogeneous and incomplete modality conditions. On average, it boosts rank correlation by 6.1\%, 8.3\%, and 1.4\%; reduces error by 12.7\%, 15.3\%, and 6.4\%; and lowers relative error by 13.9\%, 14.1\%, and 5.2\% on the three datasets, establishing a strong baseline for multi-modal continual AQA under realistic scenarios.

Our contributions can be summarized as follows:
\begin{itemize}
    \item We identify non-stationary modality imbalance in multi-modal continual AQA and show its practical significance.
    \item We propose BriMA, consisting of memory-guided bridging imputation and modality-aware replay to stabilize modality completion and mitigate distribution shifts.
    \item Extensive experiments on RG, Fis-V, and FS1000 show consistent gains in correlation and error, establishing a strong baseline for this realistic setting. \end{itemize}
 \section{Related Work}

\myPara{Action Quality Assessment (AQA)} \cite{parmar2016measuring,parmar2017learning,parmar2019and,xu2025human,zhou2024cofinal,zhou2025phi,xu2025dancefix} aims to automatically evaluate how well an action is performed and has been widely applied in sports scoring \cite{liu2023figure,ashutoshlearning}, skill assessment \cite{ding2023sedskill,pan2025basket}, and rehabilitation \cite{zhou2023video,bruce2024egcn++,li2024egoexo}. Existing efforts evolve from hand-crafted representations to deep video models based on CNNs \cite{xiang2018s3d}, RNNs \cite{jain2020action}, 3D ConvNets \cite{zhou2023prior,xiang2018s3d}, and Transformers \cite{bai2022action}, as well as ranking and contrastive supervision \cite{joung2023contrastive}. Despite these advances, AQA remains challenging due to its fine-grained, score-sensitive nature, and large variations across action types \cite{zhou2024comprehensivesurveyactionquality,xu2022finediving,xu2024fineparser}.
Recent studies demonstrate that multi-modal cues such as pose \cite{hirosawa2023action,qi2025action,dong2025lucidaction}, flow \cite{zeng2024multimodal}, audio \cite{xia2023skating,xu2025language}, or gaze \cite{huang2024egoexolearn} can significantly enhance AQA. Fusion and alignment strategies, including cross-attention, graph modeling \cite{xu2024vision,xu2025quality}, and correlation-based alignment, have been proposed to leverage complementary information. However, real-world AQA data often exhibit modality imbalance, missing-modality cases, and inconsistent modality quality, whereas existing multi-modal AQA methods generally assume complete and synchronized inputs \cite{zhou2024comprehensivesurveyactionquality}. Although some studies \cite{zhou2024magr,li2024continual,dadashzadeh2024pecop} adopt continual learning techniques to mitigate task-wise non-stationarity, they are limited to single-modality settings and do not address the additional challenges introduced by modality heterogeneity with missing modalities in real-world scenarios.

\myPara{Continual Learning (CL)} \cite{wang2024comprehensive,wang2025hide,wang2023incorporating} seeks to incrementally learn new tasks without forgetting previously learned knowledge. Prior work spans regularization-based~\cite{james2017ewc,zenke2017continual}, replay-based~\cite{buzzega2020dark,riemer2019learning,yang2023neural}, and architecture-based~\cite{wang2023incorporating,serra2018overcoming} methods. Although effective on classification benchmarks, these methods struggle with the fine-grained regression, heterogeneous distributions, and modality inconsistencies inherent to AQA. Existing multi-modal CL methods \cite{liang2025boosting,zhou2025learning,huo2025continue} do not address modality imbalance issues, leaving multi-modal continual AQA largely unexplored.

\myPara{Incomplete Multi-Modal Learning} \cite{wu2024deep,zhou2024uncertainty} deals with scenarios where one or more modalities are missing during training or inference. Existing efforts typically rely on simple imputation strategies such as zero-filling, feature masking, or hallucination networks that reconstruct missing modalities from available ones. More advanced methods \cite{lang2025retrieval,xu2024leveraging,li2024toward,zeng2024missing} employ cross-modal alignment or shared latent spaces to mitigate modality gaps, yet these methods assume stationary data distributions and consistent cross-modal relationships. In contrast, our approach addresses incomplete multi-modal learning in a continual setting, where modality availability evolves across tasks and distribution shifts amplify modality imbalance.  \begin{figure*}
    \centering
    \includegraphics[width=\linewidth,clip,trim=0 100 0 100]{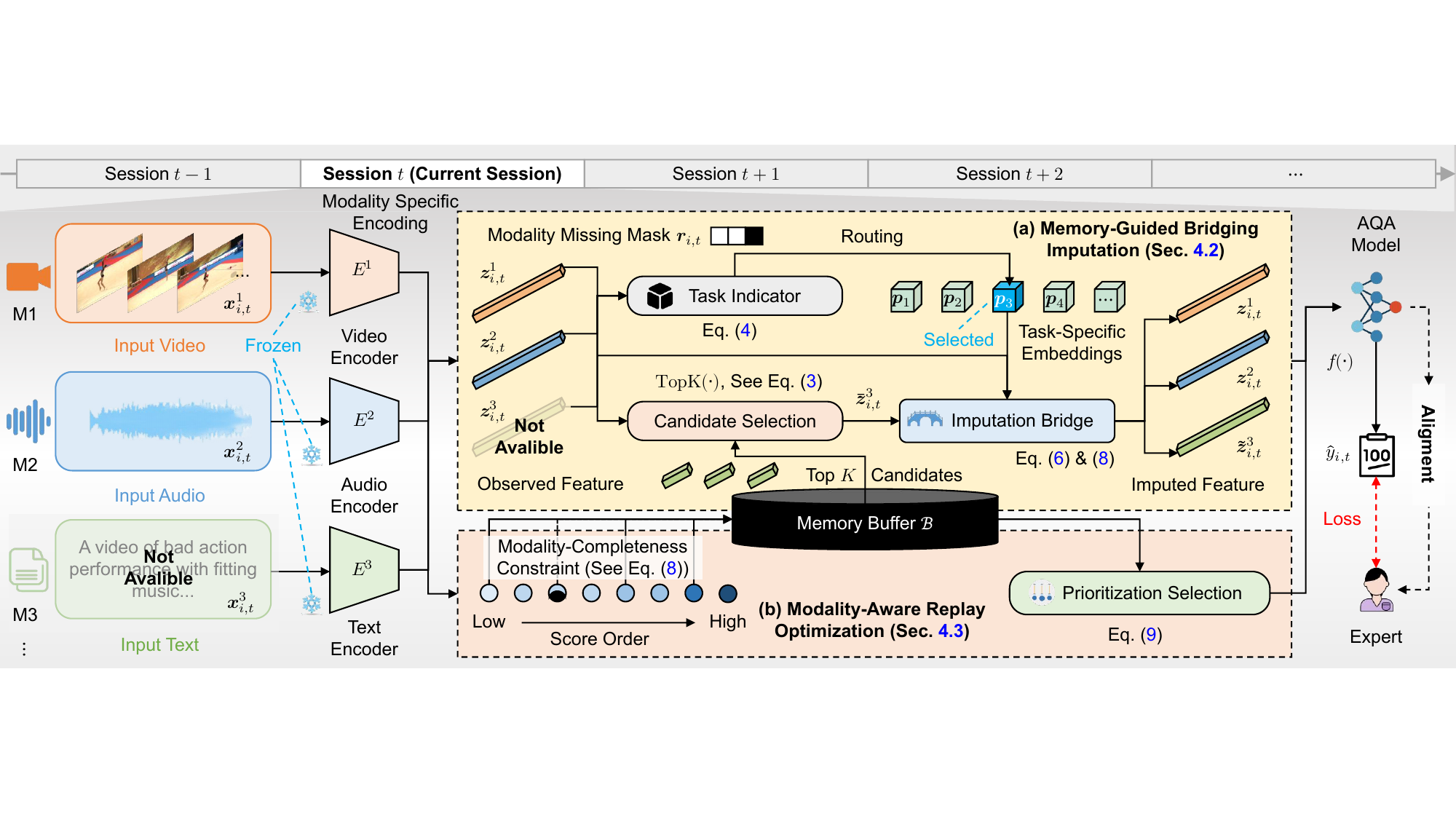}
    \caption{
    Overview of BriMA.  
    At each session, incomplete multi-modal inputs are encoded and completed via memory-guided bridging imputation (see \cref{sec:mbi}), then scored by the AQA model.  
    The modality-aware replay optimization (see \cref{sec:mro}) selects and updates representative prototypes in the memory bank to maintain consistency in continual adaptation across tasks.
    }
    \label{fig:framework}
    \vspace{-0.5cm}
\end{figure*}

\section{Problem Definition and Notations}
We define the multi-modal continual AQA setting with non-stationary modality imbalance and introduce the notations.

\myPara{Multi-Modal AQA.}
Traditional multi-modal AQA methods \cite{xu2024vision,xia2023skating,xu2025language} operate under a fully observed multi-modality setting. An action instance is represented by $M$ heterogeneous modalities $\{\boldsymbol{x}^1,\dots,\boldsymbol{x}^M\}$ with an associated score $y\in\mathbb{R}$. Each modality $\boldsymbol{x}^m$ is encoded as a feature vector $\boldsymbol{z}^m = E^m(\boldsymbol{x}^m)$, and the AQA model $f$ jointly performs feature fusion and score prediction as $\hat{y} = f(\boldsymbol{z}^1,\dots,\boldsymbol{z}^M)$.

\myPara{Continual AQA.}
Existing continual AQA methods \cite{zhou2024magr,li2024continual,dadashzadeh2024pecop} consider a sequence of tasks $\{\mathcal{T}_1,\dots,\mathcal{T}_T\}$, where each task $\mathcal{T}_t$ provides samples for score prediction. The model updates its parameters $\theta_f$ using only data from the current task, while being required to retain performance on all previously learned tasks. A standard formulation is
\begin{equation}
\label{eq:caqa_objective}
\min_{\theta_f}\;
\mathcal{L}_{\text{score}}
+\lambda_{\text{mem}}\,\mathcal{L}_{\text{mem}},
\end{equation}
where $\mathcal{L}_{\text{score}}$ is typically a mean-squared error loss on the current task, and $\mathcal{L}_{\text{mem}}$ (see \cref{eq:mem_loss}) leverages a memory buffer $\mathcal{B}_{t-1}$ to regularize updates and mitigate forgetting.

\myPara{Non-Stationary Modality Imbalance.}
The above formulations assume that all modalities remain fully observed across tasks, which rarely holds in real-world AQA.  
For a sample $(i,t)$ at task $t$, only a subset of modalities $\mathcal{O}_{i,t}\subseteq\{1,\dots,M\}$ may be available, while the remaining $\mathcal{M}_{i,t}=\{1,\dots,M\}\setminus\mathcal{O}_{i,t}$ are missing due to sensor failures or annotation gaps.  
When $m\in\mathcal{M}_{i,t}$, the modality input $\boldsymbol{x}_{i,t}^m$ and its feature $\boldsymbol{z}_{i,t}^m$ are absent, altering both the input structure and the underlying data distribution.  
As the missing pattern $\mathcal{M}_{i,t}$ evolves over tasks, the model must cope with simultaneous input sparsity and temporal distribution drift, posing a modality-driven CL challenge that existing multi-modal and continual AQA methods cannot handle.

\myPara{Multi-Modal Continual AQA.}
We consider multi-modal continual AQA under non-stationary modality imbalance, where each task 
$\mathcal{T}_t$ provides samples 
$\mathcal{D}_t=\{(\{\boldsymbol{x}_{i,t}^m\}_{m\in\mathcal{O}_{i,t}},\, y_{i,t})\}_{i=1}^{N_t}$ 
with only a subset of modalities $\mathcal{O}_{i,t}$ observed and the rest 
$\mathcal{M}_{i,t}$ missing.  
Observable modalities are encoded as $\boldsymbol{z}_{i,t}^m=E^m(\boldsymbol{x}_{i,t}^m)$,  
while missing ones require reconstructed features $\tilde{\boldsymbol{z}}_{i,t}^m$.
Compared with continual AQA, multi-modal continual AQA additionally learns both the scoring parameters $\theta_f$ and the reconstruction parameters $\theta_g$, which are jointly optimized to produce accurate score predictions while maintaining cross-task stability.
This leads to the optimization objective
\begin{equation}
\label{eq:objective}
\min_{\theta_f,\theta_g}\;
\mathcal{L}_{\text{score}}
+\lambda_{\text{mem}}\,\mathcal{L}_{\text{mem}}
+\lambda_{\text{rec}}\,\mathcal{L}_{\text{rec}},
\end{equation}
where $\mathcal{L}_{\text{score}}$ and $\mathcal{L}_{\text{mem}}$ follow \cref{eq:caqa_objective}, 
and $\mathcal{L}_{\text{rec}}$ (see \cref{eq:rec_loss}) enforces feature-level reconstruction for missing modalities.  
Unlike raw-data imputation, we store and reconstruct lightweight feature embeddings for efficiency and privacy.

 \section{BriMA: Bridged Modality Adaptation} \label{sec:method}

We propose \textbf{Bri}dged \textbf{M}odality \textbf{A}daptation (BriMA) for multi-modal continual AQA under practical modality imbalance, outlining its core idea and two main components.

\subsection{Core Idea and Framework Overview}

\myPara{Challenges.}
Multi-modal continual AQA is challenging because evolving modality availability reshapes the feature space, while score prediction is highly sensitive to these shifts, leading to representation drift that interacts with continual forgetting. Existing strategies fall short: simple imputation \cite{wu2024deep} amplifies noise in score estimation, retrieval-based completion \cite{wang2022modality,lang2025retrieval} often pulls mismatched scoring contexts that disrupt representation alignment, and generative synthesis \cite{ma2021smil,du2018semi} cannot maintain score fidelity under limited supervision common in AQA. These errors accumulate across tasks and interact with CL dynamics, making it difficult to jointly achieve accurate scoring in~\cref{eq:objective}.

\myPara{Core Idea.}
BriMA introduces a stable bridging space that anchors incomplete modalities to shared task structure and cross-task memory. By reconstructing missing features through exemplar-aligned residuals and reinforcing temporal consistency via modality-aware replay, BriMA produces robust, drift-resistant scoring. Grounding completion in this structured space avoids the noise of imputation and the instability of generative synthesis, preventing error accumulation and maintaining reliable performance across tasks.

\myPara{Framework Overview.}
The framework of BriMA is illustrated in \cref{fig:framework}.  
At session $t$, observed modalities produce features $\boldsymbol{z}_{i,t}^m = E^m(\boldsymbol{x}_{i,t}^m)$, while missing modalities $\mathcal{M}_{i,t}$ require reconstructed features $\tilde{\boldsymbol{z}}_{i,t}^m$.  
The Memory-guided Bridging Imputation (MBI) module (see \cref{sec:mbi}) generates these reconstructed features by retrieving structurally aligned exemplars from the memory $\mathcal{B}_{t-1}$ and predicting a residual correction that anchors each modality to a stable, task-consistent bridging space.  
The scoring model $f$ then performs multi-modal fusion and prediction jointly as  
$\hat{y}_{i,t} = f(\{\boldsymbol{z}_{i,t}^m\}_{m\in\mathcal{O}_{i,t}},\,\{\tilde{\boldsymbol{z}}_{i,t}^m\}_{m\in\mathcal{M}_{i,t}})$.  
To maintain temporal stability, the Modality-aware Replay Optimization (MRO) module (see \cref{sec:mro}) selects informative samples from $\mathcal{B}_{t-1}$ based on modality distortion and score drift, replaying them to counteract distribution shift and forgetting.  
After learning on task $t$, representative and reliable samples are updated into the memory. A complete description of the training pipeline is given in \textbf{Appendix} \cref{alg:training}.

\subsection{MBI: Memory-Guided Bridging Imputation} \label{sec:mbi}

\myPara{Design Idea.}
MBI targets the core difficulty of modality absence in multi-modal continual AQA, where 
direct imputation \cite{wu2024deep,zeng2024missing} amplifies noise, 
retrieval-based filling \cite{wang2022modality,lang2025retrieval} retrieves mismatched scoring contexts,  
and generative synthesis \cite{ma2021smil,du2018semi} is unreliable under limited supervision.  
To overcome these limitations, MBI constructs a task-consistent bridging space that anchors incomplete inputs to stable cross-task structure.  
It operates in three steps: (1) retrieve structurally aligned exemplars from memory,  
(2) estimate a residual correction rather than full-feature synthesis, and  
(3) combine task-specific cues to regularize reconstruction.  
This design yields noise-resistant and score-preserving modality completion suitable for continual adaptation.

\myPara{Candidate Selection.}
For each missing modality $m \in \mathcal{M}_{i,t}$, MBI retrieves a set of $K$ exemplar features 
$\{\boldsymbol{z}_{j,t'}^{m}\}_{j=1}^{K}$ from the memory buffer $\mathcal{B}_{t-1}$, 
where $t' < t$ indexes previous continual sessions. 
Retrieval is performed by computing the cosine similarity between the current observed representation 
$\boldsymbol{z}_{i,t}^{\mathcal{O}}$ and stored cross-modal embeddings:
\begin{equation}
\label{eq:retrieval}
s_{j,t'} =
\frac{
\langle \boldsymbol{z}_{i,t}^{\mathcal{O}},\, \boldsymbol{z}_{j,t'}^{\mathcal{O}} \rangle
}{
\|\boldsymbol{z}_{i,t}^{\mathcal{O}}\|
\,
\|\boldsymbol{z}_{j,t'}^{\mathcal{O}}\|
},
~~
\{\boldsymbol{z}_{j,t'}^{m}\}_{j=1}^{K}
=
\operatorname{TopK}_{j}(s_{j,t'}).
\end{equation}
This selects semantically aligned exemplars from earlier tasks, ensuring that modality reconstruction is anchored to contextually consistent cross-modal structure.

\myPara{Task Indicator.}
We introduce a task indicator to specify which modalities are missing and to provide lightweight task-specific conditioning for the bridging network.  
In this study, we follow the standard assumption that the missing-modality set $\mathcal{M}_{i,t}$ is known during training, so no detector needs to be learned.  
Accordingly, we define a binary mask
\begin{equation}
\label{eq:mask}
\boldsymbol{r}_{i,t} \in \{0,1\}^M, 
\quad 
r_{i,t}^{m} = \mathbbm{1}[m \in \mathcal{M}_{i,t}],
\end{equation}
and use it to select a task-specific embedding $\boldsymbol{p}_t^{m}$ from a learnable pool  
$\mathcal{P} = \{\boldsymbol{p}_t^{1},\dots,\boldsymbol{p}_t^{M}\}$.  
The conditioning vector
\begin{equation}
\label{eq:task_indicator}
\boldsymbol{c}_{t}^{m} = \mathrm{MLP}([\boldsymbol{p}_t^{m}])
\end{equation}
modulates the bridging network for modality $m$, ensuring that reconstruction remains consistent with the task domain under evolving modality imbalance.

\myPara{Imputation Bridge.}
Given the retrieved exemplar set 
$\{\boldsymbol{z}_{j,t'}^{m}\}_{j=1}^{K}$ with similarity scores 
$\{s_{j,t'}\}$, we compute normalized exemplar weights  
$w_{j}=\mathrm{softmax}(s_{j,t'})$ and form an initial estimate
$\bar{\boldsymbol{z}}_{i,t}^{m}=\sum_{j=1}^{K} w_{j}\,\boldsymbol{z}_{j,t'}^{m}$.
Conditioned on $\boldsymbol{z}_{i,t}^{\mathcal{O}}$ and the task-specific vector 
$\boldsymbol{c}_{t}^{m}$, the bridging network learns a residual refinement:
\begin{equation}
\label{eq:bridge}
\Delta\boldsymbol{z}_{i,t}^{m}
= B_{\Theta}\!\left(\boldsymbol{z}_{i,t}^{\mathcal{O}},\, 
\bar{\boldsymbol{z}}_{i,t}^{m},\, \boldsymbol{c}_{t}^{m}\right),
\quad
\tilde{\boldsymbol{z}}_{i,t}^{m}
= \bar{\boldsymbol{z}}_{i,t}^{m} + \Delta\boldsymbol{z}_{i,t}^{m}.
\end{equation}
The reconstruction loss enforces alignment between reconstructed and ground-truth features:
\begin{equation}
\label{eq:rec_loss}
\mathcal{L}_{\text{rec}}
=\left\|\tilde{\boldsymbol{z}}_{i,t}^{m}-\boldsymbol{z}_{i,t}^{m}\right\|_2^2,
\end{equation}
where the weighted exemplar prior offers a well-justified starting point and the residual correction adapts to task-dependent modality shifts, yielding semantically consistent imputation under non-stationary modality imbalance.

\begin{table*}[]
    \centering
    \caption{Performance comparison on the RG dataset. 
    \textbf{Bold} values indicate the best results. 
    {\color{violet}Joint Training (JT)} and {\color{teal}Sequential Training (ST)} denote the upper and lower bounds, while {\color{orange}rehearsal-free} and {\color{yellow!80!black}rehearsal-based} methods represent different CL strategies. 
    $\uparrow$: higher is better; $\downarrow$: lower is better. 
Average SRCC is computed using Fisher-$z$ transformation.}
    \label{tab:rg}
    \small
    \setlength{\tabcolsep}{6pt}
    \resizebox{\linewidth}{!}{
    \begin{tabular}{rr
    S[table-format=1.3, input-symbols={\textbf}]
    S[table-format=1.3, input-symbols={\textbf}]
    S[table-format=1.3, input-symbols={\textbf}]
    S[table-format=1.3, input-symbols={\textbf}]
    S[table-format=1.3, input-symbols={\textbf}]
    S[table-format=2.2, input-symbols={\textbf}]
    S[table-format=2.2, input-symbols={\textbf}]
    S[table-format=2.2, input-symbols={\textbf}]
    S[table-format=2.2, input-symbols={\textbf}]
    S[table-format=2.2, input-symbols={\textbf}]
    S[table-format=1.3, input-symbols={\textbf}]
    S[table-format=1.3, input-symbols={\textbf}]
    S[table-format=1.3, input-symbols={\textbf}]
    S[table-format=1.3, input-symbols={\textbf}]
    S[table-format=1.3, input-symbols={\textbf}]
    c
    c
    }
    \toprule
    \multirow{2.5}{*}{\textbf{Method}} & \multirow{2.5}{*}{\textbf{Publisher}} & \multicolumn{5}{c}{\textbf{SRCC ($\uparrow$)}} & \multicolumn{5}{c}{\textbf{MSE ($\downarrow$)}}  & \multicolumn{5}{c}{\textbf{RL2 ($\downarrow$)}}  \\
    \cmidrule(lr){3-7}
    \cmidrule(lr){8-12}
    \cmidrule(lr){13-17}
    & & 
    \multicolumn{1}{c}{Ball} & \multicolumn{1}{c}{Clubs} & \multicolumn{1}{c}{Hoop} & \multicolumn{1}{c}{Ribbon} & \multicolumn{1}{c}{Avg.} & 
    \multicolumn{1}{c}{Ball} & \multicolumn{1}{c}{Clubs} & \multicolumn{1}{c}{Hoop} & \multicolumn{1}{c}{Ribbon} & \multicolumn{1}{c}{Avg.} & 
    \multicolumn{1}{c}{Ball} & \multicolumn{1}{c}{Clubs} & \multicolumn{1}{c}{Hoop} & \multicolumn{1}{c}{Ribbon} & \multicolumn{1}{c}{Avg.}  \\
    \midrule
    \rowcolor{violet!10} JT-MLAVL\cite{xu2025language}   & CVPR'25 & 0.826 & 0.829 & 0.871 & 0.866 & 0.849 & 5.57 & 4.20 & 4.11 & 3.99 & 4.47 & {N/A} & {N/A} & {N/A} & {N/A} & {N/A}  \\
    \midrule
    \multicolumn{17}{l}{\it Modality Missing Rate $\beta=10\%$} \\
    \midrule
\rowcolor{teal!10} ST-MLAVL \cite{xu2025language}   &    CVPR'25 &  0.480 &  0.575 &  0.659 &  0.664 &  0.599 &  14.34 &   9.19 &   7.26 &   8.98 &   9.94 &  3.829 &  4.311 &  2.734 &  3.358 &  3.558 \\
\rowcolor{orange!10} SI \cite{zenke2017continual}   &    ICML'17 &  0.481 &  0.597 &  0.605 &  0.685 &  0.597 &  14.61 &   8.70 &   8.63 &   8.74 &  10.17 &  3.903 &  4.083 &  3.249 &  3.271 &  3.626 \\
\rowcolor{orange!10} EWC \cite{james2017ewc}        &    PNAS'17 &  0.524 &  0.594 &  0.597 &  0.692 &  0.605 &  13.35 &   9.41 &   9.16 &   9.11 &  10.26 &  3.565 &  4.415 &  3.448 &  3.409 &  3.709 \\
\rowcolor{orange!10} LwF \cite{li2017learning}      &   TPAMI'17 &  0.484 &  0.468 &  0.542 &  0.664 &  0.545 &  16.43 &  10.36 &  16.49 &   8.17 &  12.86 &  4.387 &  4.861 &  6.206 &  3.056 &  4.628 \\
\rowcolor{yellow!10} MER \cite{riemer2019learning}  &    ICLR'19 &  0.594 &  0.773 &  0.704 &  0.786 &  0.722 &  \bf 8.33 &  6.06 &   8.22 &   5.00 &   6.77 & \bf 2.223 & 2.843 &  3.093 &  1.870 &  2.473 \\
\rowcolor{yellow!10} DER++ \cite{buzzega2020dark}   & NeurIPS'20 &  0.543 &  0.605 & \bf 0.727 &  0.753 &  0.665 &  12.92 &   8.29 &   9.86 &   7.31 &   9.60 &  3.451 &  3.890 &  3.712 &  2.736 &  3.447 \\
\rowcolor{yellow!10} NC-FSCIL \cite{yang2023neural} &    ICLR'23 &  0.457 &  0.620 &  0.664 &  0.731 &  0.628 &  14.10 &   9.06 &   7.70 &   9.40 &  10.06 &  3.766 &  4.248 &  2.899 &  3.515 &  3.607 \\
\rowcolor{yellow!10} SLCA \cite{zhang2023slca}      &    ICCV'23 &  0.467 &  0.569 &  0.695 &  0.744 &  0.631 &  18.52 &   8.31 &  23.85 &   8.52 &  14.80 &  4.946 &  3.898 &  8.976 &  3.186 &  5.251 \\
\rowcolor{yellow!10} Fs-Aug \cite{li2024continual}  &   TCSVT'24 &  0.463 &  0.585 &  0.703 &  0.703 &  0.623 &  14.89 &   8.46 &  12.56 &   8.51 &  11.10 &  3.976 &  3.970 &  4.728 &  3.182 &  3.964 \\
\rowcolor{yellow!10} MAGR \cite{zhou2024magr}       &    ECCV'24 &  0.460 &  0.606 &  0.672 &  0.720 &  0.623 &  13.85 &   8.52 &   9.23 &   8.32 &   9.98 &  3.700 &  3.996 &  3.475 &  3.113 &  3.571 \\
\rowcolor{yellow!10} ASAL \cite{zhou2025adaptive}   &    TVCG'25 &  0.552 &  0.629 &  0.693 &  0.678 &  0.641 &  13.37 &  10.02 &   9.30 &   8.35 &  10.26 &  3.571 &  4.699 &  3.501 &  3.123 &  3.724 \\
\rowcolor{yellow!10} \bf BriMA (Ours) & -- &
\bf 0.648 & \bf 0.788 & 0.710 & \bf 0.836 & \bf 0.746 &
8.65 &  \bf 4.88 & \bf 7.12 & \bf 4.39 & \bf 6.26 &
2.310 & \bf 2.291 & \bf 2.681 & \bf 1.641 & \bf 2.231 \\
    \midrule
    \multicolumn{17}{l}{\it Modality Missing Rate $\beta=25\%$} \\
    \midrule
\rowcolor{teal!10} ST-MLAVL \cite{xu2025language}   &    CVPR'25 &  0.406 &  0.576 &  0.549 &  0.674 &  0.558 &  16.89 &   8.42 &   9.39 &   9.15 &  10.96 &  4.512 &  3.949 &  3.536 &  3.423 &  3.855 \\
\rowcolor{orange!10} SI \cite{zenke2017continual}   &    ICML'17 &  0.409 &  0.589 &  0.584 &  0.649 &  0.564 &  17.18 &   8.65 & \bf 7.81 &   9.33 &  10.74 &  4.587 &  4.057 & \bf 2.940 &  3.489 &  3.769 \\
\rowcolor{orange!10} EWC \cite{james2017ewc}        &    PNAS'17 &  0.379 &  0.580 &  0.638 &  0.652 &  0.571 &  16.85 &   9.13 &   8.91 &   9.60 &  11.12 &  4.499 &  4.284 &  3.354 &  3.592 &  3.932 \\
\rowcolor{orange!10} LwF \cite{li2017learning}      &   TPAMI'17 &  0.448 &  0.586 &  0.487 &  0.642 &  0.545 &  13.55 &   7.55 &   9.98 &   9.05 &  10.03 &  3.618 &  3.541 &  3.757 &  3.384 &  3.575 \\
\rowcolor{yellow!10} MER \cite{riemer2019learning} & ICLR'19 &
0.527 & 0.766 & 0.519 & 0.761 & 0.660 &
\bf 9.39 & 7.10 & 13.36 & 5.29 & 8.79 &
\bf 2.509 & 3.329 & 5.028 & 1.980 & 3.212 \\
\rowcolor{yellow!10} DER++ \cite{buzzega2020dark}   & NeurIPS'20 &  0.528 &  0.574 &  0.683 &  0.739 &  0.639 &  14.97 &   8.33 &  10.94 &   7.91 &  10.54 &  3.999 &  3.910 &  4.119 &  2.958 &  3.746 \\
\rowcolor{yellow!10} NC-FSCIL \cite{yang2023neural} &    ICLR'23 &  0.422 &  0.608 &  0.615 &  0.721 &  0.601 &  15.04 &   8.85 &   9.87 &   9.07 &  10.71 &  4.016 &  4.150 &  3.717 &  3.394 &  3.819 \\
\rowcolor{yellow!10} SLCA \cite{zhang2023slca}      &    ICCV'23 &  0.397 &  0.551 & \bf 0.720 &  0.715 &  0.612 &  18.92 &   8.44 &  24.82 &   9.30 &  15.37 &  5.054 &  3.962 &  9.343 &  3.480 &  5.460 \\
\rowcolor{yellow!10} Fs-Aug \cite{li2024continual}  &   TCSVT'24 &  0.472 &  0.565 &  0.607 &  0.699 &  0.592 &  17.04 &   8.56 &  16.65 &   9.23 &  12.87 &  4.551 &  4.016 &  6.268 &  3.453 &  4.572 \\
\rowcolor{yellow!10} MAGR \cite{zhou2024magr}       &    ECCV'24 &  0.364 &  0.542 &  0.589 &  0.695 &  0.558 &  15.11 &   8.63 &  11.93 &   8.39 &  11.02 &  4.035 &  4.051 &  4.490 &  3.140 &  3.929 \\
\rowcolor{yellow!10} ASAL \cite{zhou2025adaptive}   &    TVCG'25 &  0.505 &  0.609 &  0.676 &  0.682 &  0.623 &  14.36 &  10.15 &  10.31 &   8.24 &  10.77 &  3.835 &  4.762 &  3.880 &  3.084 &  3.890 \\
\rowcolor{yellow!10} \bf BriMA (Ours)                &         -- & \bf 0.592 & \bf 0.771 &  0.522 & \bf 0.794 & \bf 0.686 &   9.43 &   \bf 6.10 &  10.11 & \bf 4.92 & \bf 7.64 &  2.519 & \bf 2.860 &  3.806 & \bf 1.841 & \bf 2.757 \\
    \midrule
    \multicolumn{17}{l}{\it Modality Missing Rate $\beta=50\%$} \\
    \midrule
\rowcolor{teal!10} ST-MLAVL \cite{xu2025language}   &    CVPR'25 &  0.142 &  0.436 &  0.444 &  0.613 &  0.422 &  23.42 &   8.67 &  12.21 &  10.08 &  13.59 &  6.255 &  4.067 &  4.595 &  3.770 &  4.672 \\
\rowcolor{orange!10} SI \cite{zenke2017continual}   &    ICML'17 &  0.163 &  0.441 &  0.463 &  0.622 &  0.436 &  23.39 &   8.69 & \bf 9.31 &  10.30 &  12.92 &  6.248 &  4.078 & \bf 3.503 &  3.854 &  4.421 \\
\rowcolor{orange!10} EWC \cite{james2017ewc}        &    PNAS'17 &  0.046 &  0.410 &  0.370 &  0.528 &  0.349 &  25.13 &   8.95 &  12.72 &  12.47 &  14.82 &  6.712 &  4.201 &  4.786 &  4.664 &  5.091 \\
\rowcolor{orange!10} LwF \cite{li2017learning}      &   TPAMI'17 &  0.136 &  0.227 &  0.310 &  0.628 &  0.342 &  24.18 &  14.04 &  12.91 &   9.02 &  15.04 &  6.458 &  6.584 &  4.858 &  3.373 &  5.318 \\
\rowcolor{yellow!10} MER \cite{riemer2019learning}  &    ICLR'19 &  0.499 & \bf 0.752 &  0.389 &  0.662 &  0.594 &  10.55 &   5.31 &  15.15 &   8.69 &   9.92 &  2.818 &  2.493 &  5.701 &  3.249 &  3.565 \\
\rowcolor{yellow!10} DER++ \cite{buzzega2020dark}   & NeurIPS'20 &  0.225 &  0.510 &  0.467 &  0.571 &  0.452 &  19.68 &   8.74 &  11.32 &   9.73 &  12.37 &  5.257 &  4.100 &  4.262 &  3.639 &  4.315 \\
\rowcolor{yellow!10} NC-FSCIL \cite{yang2023neural} &    ICLR'23 &  0.226 &  0.561 &  0.499 &  0.625 &  0.490 &  19.37 &   9.44 &  12.41 &  11.24 &  13.11 &  5.172 &  4.428 &  4.669 &  4.205 &  4.619 \\
\rowcolor{yellow!10} SLCA \cite{zhang2023slca}      &    ICCV'23 &  0.198 &  0.508 &  0.406 &  0.658 &  0.458 &  21.03 &   8.52 &  24.52 &   9.72 &  15.95 &  5.618 &  3.997 &  9.230 &  3.637 &  5.620 \\
\rowcolor{yellow!10} Fs-Aug \cite{li2024continual}  &   TCSVT'24 &  0.216 &  0.508 &  0.535 &  0.651 &  0.492 &  20.09 &   8.95 &  17.78 &   9.84 &  14.17 &  5.366 &  4.199 &  6.692 &  3.680 &  4.984 \\
\rowcolor{yellow!10} MAGR \cite{zhou2024magr}       &    ECCV'24 &  0.210 &  0.520 &  0.423 &  0.668 &  0.471 &  16.45 &   8.64 &  12.99 &   9.92 &  12.00 &  4.393 &  4.054 &  4.888 &  3.712 &  4.262 \\
\rowcolor{yellow!10} ASAL \cite{zhou2025adaptive}   &    TVCG'25 &  0.305 &  0.543 &  0.544 &  0.531 &  0.486 &  17.30 &  10.23 &  10.90 &  10.34 &  12.19 &  4.620 &  4.801 &  4.104 &  3.868 &  4.348 \\
\rowcolor{yellow!10} \bf BriMA (Ours)                &         -- & \bf 0.569 &  0.739 & \bf 0.583 & \bf 0.716 & \bf 0.659 & \bf 9.77 & \bf 5.13 &  10.85 & \bf 6.99 & \bf 8.18 & \bf 2.609 & \bf 2.407 &  4.085 & \bf 2.614 & \bf 2.928 \\

    \bottomrule
    \end{tabular}
    }
\vspace{-0.25cm}
\end{table*}

\subsection{MRO: Modality-Aware Replay Optimization} \label{sec:mro}
\myPara{Design Idea.}
MRO is designed to address the combined challenges of catastrophic forgetting and evolving modality imbalance in continual AQA.  
Unlike uniform replay \cite{zhou2024magr,li2024continual}, which indiscriminately reuses all stored samples and propagates noisy or distorted modalities,  
MRO performs reliability-aware replay by (1) maintaining a clean, modality-balanced memory,  
(2) selecting past samples based on modality distortion and score drift, and  
(3) prioritizing those most critical for stabilizing prediction under shifting modality availability.  
This strategy ensures that only informative exemplars contribute to updating the model,  
resulting in more stable cross-task scoring.

\myPara{Sample Selection.}
After completing task $\mathcal{T}_t$, we construct a compact and reliable memory buffer $\mathcal{B}_t$ for future replay.  
All training samples are first ranked by their predicted scores $\hat{y}_{i,t}$ and partitioned into $Q$ quantile bins.  
From each bin, we aim to uniformly select instances that satisfy the modality-completeness constraint:
\begin{equation}
\label{eq:selection}
\mathcal{B}_t = 
\big\{(\boldsymbol{x}_{i,t}, y_{i,t})~
\big|~
i \in \mathcal{S}_t,\;
\sum_{m=1}^{M} \mathbbm{1}[m \in \mathcal{M}_{i,t}] = 0
\big\},
\end{equation}
where $\mathcal{S}_t$ denotes selected indices across bins.  
Suppose a bin contains no samples satisfying the constraint, we iteratively re-rank the unselected samples, re-partition them, and repeat the selection until the buffer is filled.

\myPara{Replay Prioritization.}
When learning the next task $\mathcal{T}_{t+1}$, replay samples are prioritized according to their vulnerability to continual drift.  
We compute (i) the modality distortion  
$d_i=\|\tilde{\boldsymbol{z}}_{i,t}^{m}-\boldsymbol{z}_{i,t}^{m}\|_2^2$  
and (ii) the score drift  
$\Delta y_i=\big|f_{\theta_f^{t+1}}(\{\boldsymbol{z}_{i,t}^{m}\})-
f_{\theta_f^{t}}(\{\boldsymbol{z}_{i,t}^{m}\})\big|$,  
where $\theta_f^{t}$ and $\theta_f^{t+1}$ denote the scoring-network parameters before and after learning $\mathcal{T}_{t+1}$.  
Each sample’s priority is defined as
\begin{equation}
\label{eq:priority}
q_i = \alpha\, d_i + (1-\alpha)\, \Delta y_i,
\end{equation}
where $\alpha$ controls the trade-off between modality discrepancy and scoring stability.  
Samples with larger $q_i$ are replayed more frequently, enabling targeted correction for instances most susceptible to drift.

\myPara{Score Consistency.}
To further stabilize continual adaptation, we regularize the scoring network against its previous snapshot on stored memory features.  
Given a replay exemplar with multi-modality embeddings $\{\boldsymbol{z}_{i,t}^{m}\}$,  
the loss is
\begin{equation}
\label{eq:mem_loss}
\mathcal{L}_{\text{mem}}
=\mathbb{E}_{i\in\mathcal{B}_t}
\Big[
\big\|
f_{\theta_f^{\,t+1}}(\{\boldsymbol{z}_{i,t}^{m}\})
-
f_{\theta_f^{\,t}}(\{\boldsymbol{z}_{i,t}^{m}\})
\big\|_2^{2}
\Big].
\end{equation}
This term penalizes temporal prediction drift and complements prioritized replay, yielding a more stable continual AQA process under evolving modality imbalance.
 \section{Experiments}
In this section, we present the main results and analyses. Additional results and analyses are provided in \textbf{Appendix}.

\begin{table}[]
    \centering
    \small
    \setlength{\tabcolsep}{3pt}
    \caption{Performance comparison on the Fis-V dataset.}
    \label{tab:fisv}
    \resizebox{\linewidth}{!}{
    \begin{tabular}{rr
    S[table-format=1.3, input-symbols={\textbf}]
    S[table-format=1.3, input-symbols={\textbf}]
    S[table-format=1.3, input-symbols={\textbf}]
    S[table-format=2.2, input-symbols={\textbf}]
    S[table-format=2.2, input-symbols={\textbf}]
    S[table-format=2.2, input-symbols={\textbf}]
    S[table-format=1.3, input-symbols={\textbf}]
    S[table-format=1.3, input-symbols={\textbf}]
    S[table-format=1.3, input-symbols={\textbf}]
    c
    c
    }
    \toprule
    \multirow{2.5}{*}{\textbf{Method}} & \multirow{2.5}{*}{\textbf{Publisher}} & 
    \multicolumn{3}{c}{\textbf{SRCC ($\uparrow$)}} & 
    \multicolumn{3}{c}{\textbf{MSE ($\downarrow$)}} & 
    \multicolumn{3}{c}{\textbf{RL2 ($\downarrow$)}}  \\
    \cmidrule(lr){3-5}
    \cmidrule(lr){6-8}
    \cmidrule(lr){9-11}
    & & 
    \multicolumn{1}{c}{PCS} & \multicolumn{1}{c}{TES} & \multicolumn{1}{c}{Avg.} &
    \multicolumn{1}{c}{PCS} & \multicolumn{1}{c}{TES} & \multicolumn{1}{c}{Avg.} &
    \multicolumn{1}{c}{PCS} & \multicolumn{1}{c}{TES} & \multicolumn{1}{c}{Avg.} \\
    \midrule
    \rowcolor{violet!10} JT-MLAVL\cite{xu2025language}   & CVPR'25 & 0.863 & 0.766 & 0.823 & 7.17 & 19.44 & 13.31 & {N/A} & {N/A} & {N/A} \\
    \midrule
    \multicolumn{11}{l}{\it Modality Missing Rate $\beta=10\%$} \\
    \midrule
\rowcolor{teal!10} ST-MLAVL \cite{xu2025language}   &    CVPR'25 &  0.648 &  0.566 &  0.609 &  17.03 &  33.31 &  25.17 &  2.721 &  2.795 &  2.758 \\
\rowcolor{orange!10} SI \cite{zenke2017continual}   &    ICML'17 &  0.630 &  0.556 &  0.594 &  16.81 &  28.77 &  22.79 &  2.685 &  2.414 &  2.550 \\
\rowcolor{orange!10} EWC \cite{james2017ewc}        &    PNAS'17 &  0.584 &  0.551 &  0.568 &  18.36 &  33.43 &  25.89 &  2.933 &  2.805 &  2.869 \\
\rowcolor{orange!10} LwF \cite{li2017learning}      &   TPAMI'17 &  0.369 &  0.165 &  0.270 &  25.43 &  52.97 &  39.20 &  4.062 &  4.446 &  4.254 \\
\rowcolor{yellow!10} MER \cite{riemer2019learning}  &    ICLR'19 &  0.624 &  0.416 &  0.528 &  19.86 &  43.13 &  31.50 &  3.173 &  3.620 &  3.396 \\
\rowcolor{yellow!10} DER++ \cite{buzzega2020dark}   & NeurIPS'20 &  0.682 &  0.539 &  0.616 &  14.05 &  33.56 &  23.81 &  2.244 &  2.816 &  2.530 \\
\rowcolor{yellow!10} NC-FSCIL \cite{yang2023neural} &    ICLR'23 &  0.702 &  0.540 &  0.628 &  15.34 &  30.08 &  22.71 &  2.451 &  2.524 &  2.487 \\
\rowcolor{yellow!10} SLCA \cite{zhang2023slca}      &    ICCV'23 &  0.671 &  0.501 &  0.592 &  24.25 &  36.53 &  30.39 &  3.874 &  3.066 &  3.470 \\
\rowcolor{yellow!10} Fs-Aug \cite{li2024continual}  &   TCSVT'24 &  0.664 &  0.514 &  0.594 &  14.65 &  36.34 &  25.50 &  2.341 &  3.049 &  2.695 \\
\rowcolor{yellow!10} MAGR \cite{zhou2024magr}       &    ECCV'24 & \bf 0.717 &  0.488 &  0.615 &  17.39 &  38.18 &  27.79 &  2.778 &  3.204 &  2.991 \\
\rowcolor{yellow!10} ASAL \cite{zhou2025adaptive}   &    TVCG'25 &  0.695 &  0.457 &  0.589 &  14.53 &  39.12 &  26.83 &  2.322 &  3.283 &  2.802 \\
\rowcolor{yellow!10} \bf BriMA (Ours)                          &         -- &  0.705 & \bf 0.594 & \bf 0.653 & \bf 13.60 & \bf 25.78 & \bf 19.69 & \bf 2.173 & \bf 2.164 & \bf 2.168 \\
    \midrule
    \multicolumn{11}{l}{\it Modality Missing Rate $\beta=25\%$} \\
    \midrule
\rowcolor{teal!10} ST-MLAVL \cite{xu2025language}   &    CVPR'25 &  0.622 &  0.572 &  0.598 &  17.98 &  30.11 &  24.05 &  2.873 &  2.527 &  2.700 \\
\rowcolor{orange!10} SI \cite{zenke2017continual}   &    ICML'17 &  0.594 &  0.544 &  0.569 &  18.30 &  31.37 &  24.83 &  2.924 &  2.632 &  2.778 \\
\rowcolor{orange!10} EWC \cite{james2017ewc}        &    PNAS'17 &  0.625 &  0.573 &  0.599 &  17.69 &  30.90 &  24.29 &  2.826 &  2.593 &  2.709 \\
\rowcolor{orange!10} LwF \cite{li2017learning}      &   TPAMI'17 &  0.216 &  0.136 &  0.177 &  33.66 &  53.09 &  43.37 &  5.377 &  4.455 &  4.916 \\
\rowcolor{yellow!10} MER \cite{riemer2019learning}  &    ICLR'19 &  0.545 &  0.459 &  0.503 &  21.57 &  38.43 &  30.00 &  3.446 &  3.225 &  3.336 \\
\rowcolor{yellow!10} DER++ \cite{buzzega2020dark}   & NeurIPS'20 &  0.649 &  0.537 &  0.596 &  16.85 &  36.47 &  26.66 &  2.691 &  3.060 &  2.876 \\
\rowcolor{yellow!10} NC-FSCIL \cite{yang2023neural} &    ICLR'23 &  0.617 &  0.576 &  0.597 &  18.58 &  31.96 &  25.27 &  2.968 &  2.682 &  2.825 \\
\rowcolor{yellow!10} SLCA \cite{zhang2023slca}      &    ICCV'23 &  0.597 &  0.569 &  0.583 &  20.93 &  41.93 &  31.43 &  3.344 &  3.519 &  3.431 \\
\rowcolor{yellow!10} Fs-Aug \cite{li2024continual}  &   TCSVT'24 &  0.649 &  0.551 &  0.602 &  21.61 &  35.30 &  28.45 &  3.451 &  2.962 &  3.207 \\
\rowcolor{yellow!10} MAGR \cite{zhou2024magr}       &    ECCV'24 &  0.667 &  0.513 &  0.596 &  16.04 &  33.30 &  24.67 &  2.563 &  2.795 &  2.679 \\
\rowcolor{yellow!10} ASAL \cite{zhou2025adaptive}   &    TVCG'25 &  0.664 &  0.519 &  0.596 &  20.24 &  36.20 &  28.22 &  3.233 &  3.037 &  3.135 \\
\rowcolor{yellow!10} \bf BriMA (Ours)                           &         -- & \bf 0.688 & \bf 0.632 & \bf 0.661 & \bf 15.37 & \bf 26.94 & \bf 21.15 & \bf 2.455 & \bf 2.261 & \bf 2.358  \\
    \midrule
    \multicolumn{11}{l}{\it Modality Missing Rate $\beta=50\%$} \\
    \midrule
\rowcolor{teal!10} ST-MLAVL \cite{xu2025language}   &    CVPR'25 &  0.386 &  0.569 &  0.483 &  23.41 &  29.33 &  26.37 &  3.740 &  2.461 &  3.100 \\
\rowcolor{orange!10} SI \cite{zenke2017continual}   &    ICML'17 &  0.493 &  0.551 &  0.523 &  21.82 &  31.65 &  26.73 &  3.486 &  2.656 &  3.071 \\
\rowcolor{orange!10} EWC \cite{james2017ewc}        &    PNAS'17 &  0.442 &  0.479 &  0.461 &  31.39 &  48.39 &  39.89 &  5.014 &  4.061 &  4.537 \\
\rowcolor{orange!10} LwF \cite{li2017learning}      &   TPAMI'17 &  0.435 &  0.257 &  0.349 &  25.72 &  46.07 &  35.90 &  4.109 &  3.866 &  3.988 \\
\rowcolor{yellow!10} MER \cite{riemer2019learning}  &    ICLR'19 &  0.494 &  0.351 &  0.425 &  26.59 &  43.90 &  35.25 &  4.248 &  3.684 &  3.966 \\
\rowcolor{yellow!10} DER++ \cite{buzzega2020dark}   & NeurIPS'20 &  0.592 &  0.530 &  0.562 &  18.88 &  31.12 &  25.00 &  3.016 &  2.611 &  2.814 \\
\rowcolor{yellow!10} NC-FSCIL \cite{yang2023neural} &    ICLR'23 &  0.630 &  0.544 &  0.588 &  17.58 &  37.10 &  27.34 &  2.808 &  3.114 &  2.961 \\
\rowcolor{yellow!10} SLCA \cite{zhang2023slca}      &    ICCV'23 &  0.441 &  0.564 &  0.505 &  35.12 &  46.28 &  40.70 &  5.610 &  3.884 &  4.747 \\
\rowcolor{yellow!10} Fs-Aug \cite{li2024continual}  &   TCSVT'24 &  0.602 &  0.488 &  0.547 &  24.22 &  41.61 &  32.92 &  3.869 &  3.492 &  3.681 \\
\rowcolor{yellow!10} MAGR \cite{zhou2024magr}       &    ECCV'24 &  0.634 &  0.446 &  0.547 &  18.31 &  35.12 &  26.72 &  2.925 &  2.947 &  2.936 \\
\rowcolor{yellow!10} ASAL \cite{zhou2025adaptive}   &    TVCG'25 &  0.407 &  0.494 &  0.451 &  37.05 &  38.71 &  37.88 &  5.918 &  3.249 &  4.583 \\
\rowcolor{yellow!10} \bf BriMA (Ours)                           &         -- & \bf 0.648 & \bf 0.620 & \bf 0.634 & \bf 15.70 & \bf 28.03 & \bf 21.86 & \bf 2.508 & \bf 2.352 & \bf 2.430 \\
    \bottomrule
    \end{tabular}
    }
\vspace{-0.25cm}
\end{table}

\subsection{Experimental Setting}

\myPara{Datasets.}
We evaluate BriMA on three representative multi-modal AQA benchmarks, considering three commonly available modalities: video, audio, and textual commentary. 
The \textbf{Rhythmic Gymnastics (RG)} dataset \cite{zeng2020hybrid} contains 1,000 videos with a relatively small sample size but diverse action types (Ball, Clubs, Hoop, Ribbon), with 200 training and 50 evaluation samples per type. 
In contrast, the \textbf{Figure Skating Video (Fis-V)} dataset \cite{parmar2017learning} offers a moderate-scale collection of 500 videos but includes only two score types: Total Element Score (TES) and Program Component Score (PCS). 
The \textbf{FS1000} dataset \cite{xia2023skating} is a large-scale benchmark with 1,000 training and 247 validation videos, providing seven component scores (TES, PCS, SS, TR, PE, CO, IN). 
Following prior works \cite{xu2025language,xu2024vision}, we train separate models for each action or score type and split each action or subscore into $T=5$ sequential tasks.

\myPara{Evaluation Metrics.}
Existing works adopt different evaluation metrics, making direct comparison difficult. 
To ensure fairness and comprehensiveness, we report all three existing metrics: 
Spearman’s Rank Correlation Coefficient (SRCC), MSE, and RL2.  
SRCC measures the monotonic relationship between predicted scores $\hat{y}_i$ 
and ground-truth scores $y_i$ based on their rank correlation, defined as
\begin{equation}
    \text{SRCC} = 
    \frac{\sum_{i=1}^{N} (r_i - \bar{r})(\hat{r}_i - \bar{\hat{r}})}
    {\sqrt{\sum_{i=1}^{N} (r_i - \bar{r})^2} 
     \sqrt{\sum_{i=1}^{N} (\hat{r}_i - \bar{\hat{r}})^2}},
\end{equation}
where $r_i$ and $\hat{r}_i$ denote the rank values of the ground-truth 
and predicted scores, respectively.  
MSE quantifies the average squared prediction error between $\hat{y}_i$ and $y_i$, 
while RL2 represents the normalized mean squared error, defined as
\begin{equation}
    \text{RL2} = 
    \frac{1}{N} \sum_{i=1}^{N} 
    \frac{(y_i - \hat{y}_i)^2}
    {(\max(\bm{y}) - \min(\bm{y}))^2} \times 100.
\end{equation}

\begin{table}[]
    \centering
    \small
    \setlength{\tabcolsep}{3pt}
    \caption{Performance comparison on the FS1000 dataset. 
        \textbf{Subscore} results are shown in Appendix \cref{tab:supp-fs1000}.}
    \label{tab:fs1000}
    \resizebox{\linewidth}{!}{
    \begin{tabular}{rr
    S[table-format=1.3, input-symbols={\textbf}]
    S[table-format=1.3, input-symbols={\textbf}]
    S[table-format=1.3, input-symbols={\textbf}]
    S[table-format=2.2, input-symbols={\textbf}]
    S[table-format=2.2, input-symbols={\textbf}]
    S[table-format=2.2, input-symbols={\textbf}]
    S[table-format=1.3, input-symbols={\textbf}]
    S[table-format=1.3, input-symbols={\textbf}]
    S[table-format=1.3, input-symbols={\textbf}]
    c
    c
    }
    \toprule
    \multirow{2.5}{*}{\textbf{Method}} & \multirow{2.5}{*}{\textbf{Publisher}} & 
    \multicolumn{3}{c}{\textbf{SRCC ($\uparrow$)}} & 
    \multicolumn{3}{c}{\textbf{MSE ($\downarrow$)}} & 
    \multicolumn{3}{c}{\textbf{RL2 ($\downarrow$)}}  \\
    \cmidrule(lr){3-5}
    \cmidrule(lr){6-8}
    \cmidrule(lr){9-11}
    & & 
    \multicolumn{1}{c}{10\%} & \multicolumn{1}{c}{25\%} & \multicolumn{1}{c}{50\%} &
    \multicolumn{1}{c}{10\%} & \multicolumn{1}{c}{25\%} & \multicolumn{1}{c}{50\%} &
    \multicolumn{1}{c}{10\%} & \multicolumn{1}{c}{25\%} & \multicolumn{1}{c}{50\%} \\
    \midrule
    \rowcolor{violet!10} JT-MLAVL\cite{xu2025language}   & CVPR'25 
    & \bf 0.90 & N/A & N/A  & 10.39 & N/A & N/A  & N/A & N/A & N/A \\
    \midrule
    
    \rowcolor{teal!10} ST-MLAVL \cite{xu2025language} & CVPR'25 
    & 0.728 & 0.724 & 0.665 
    & 32.56 & 35.47 & 54.24
    & 1.592 & 1.645 & 2.012 \\

    \rowcolor{orange!10} SI \cite{zenke2017continual} & ICML'17 
    & 0.719 & 0.725 & 0.656
    & 36.77 & 39.83 & 46.74
    & 1.650 & 1.631 & 1.932 \\

    \rowcolor{orange!10} EWC \cite{james2017ewc} & PNAS'17
    & 0.716 & 0.710 & 0.635
    & 54.97 & 56.58 & 58.06
    & 2.150 & 2.041 & 2.354 \\

    \rowcolor{orange!10} LwF \cite{li2017learning} & TPAMI'17
    & 0.699 & 0.626 & 0.583
    & 35.19 & 34.67 & 45.76
    & 1.726 & 2.271 & 2.625 \\

    \rowcolor{yellow!10} MER \cite{riemer2019learning} & ICLR'19
    & 0.739 & 0.710 & 0.627
    & 29.95 & 32.58 & 39.40
    & 1.758 & 1.818 & 2.596 \\

    \rowcolor{yellow!10} DER++ \cite{buzzega2020dark} & NeurIPS'20
    & 0.755 & 0.733 & 0.668
    & 31.23 & 41.45 & 46.17
    & 1.513 & 1.623 & 2.003 \\

    \rowcolor{yellow!10} NC-FSCIL \cite{yang2023neural} & ICLR'23
    & 0.746 & 0.728 & 0.659
    & 32.62 & 38.59 & 49.92
    & 1.493 & 1.643 & 2.011 \\

    \rowcolor{yellow!10} SLCA \cite{zhang2023slca} & ICCV'23
    & 0.740 & 0.719 & 0.675
    & 53.75 & 58.41 & 59.01
    & 2.109 & 2.133 & 2.036 \\

    \rowcolor{yellow!10} Fs-Aug \cite{li2024continual}  & TCSVT'24
    & 0.741 & 0.725 & 0.677
    & 34.81 & 38.38 & 47.98
    & 1.637 & 1.711 & 1.999 \\

    \rowcolor{yellow!10} MAGR \cite{zhou2024magr}  & ECCV'24
    & 0.730 & 0.711 & 0.658
    & 32.75 & 39.53 & 42.62
    & 1.609 & 1.681 & 2.067 \\

    \rowcolor{yellow!10} ASAL \cite{zhou2025adaptive}  & TVCG'25
    & 0.739 & 0.724 & 0.666
    & \bf 29.90 & 35.61 & 38.75
    & 1.623 & 1.667 & 1.923 \\

    \rowcolor{yellow!10} \bf BriMA (Ours) & --
    & \bf 0.756 & \bf 0.740 & \bf 0.698
    & 33.35 & \bf 31.35 & \bf 35.29
    & \bf 1.441 & \bf 1.572 & \bf 1.823 \\
    
    \bottomrule
    \end{tabular}
    }
\end{table}

\myPara{Implementation Details.}
All experiments are implemented in PyTorch and run on a single NVIDIA RTX~5090 GPU.  
All models are trained for 50~epochs using Adam optimizer with an initial learning rate of $3\times10^{-4}$,  
weight decay of $10^{-4}$, and a cosine learning rate scheduler with decay rate 0.01.  
The batch size is set to 4 for RG and 8 for Fis-V and FS1000.  
The mini-batch size for replay sampling is 2 and 4 for the respective datasets.  
Following \cite{zhou2024magr}, we set $\lambda_{\text{mem}},\lambda_{\text{rec}}$ to 1
and the memory buffer size to 50 for all rehearsal methods ($Q=10$).  
The replay coefficient is fixed to $\alpha=0.5$.  
To simulate non-stationary modality imbalance,  
we vary the missing-modality rate $\beta\in\{10\%,\,25\%,\,50\%\}$.  
For fair comparison, all methods use the same random seed to generate missing-modality patterns.
Other implementation details are supplemented in the Appendix.

\subsection{Comparisons with State-of-the-Arts}
\myPara{Assessment Performance.} Across \cref{tab:rg,tab:fisv,tab:fs1000}, Sequential Training (ST) \cite{xu2025language} suffers the largest drop from the Joint Training (JT) upper bound as $\beta$ increases, confirming severe forgetting under evolving modality imbalance. Rehearsal-free methods (SI \cite{zenke2017continual}, EWC \cite{james2017ewc}, LwF \cite{li2017learning}) moderately alleviate forgetting but remain unstable when modalities are corrupted or missing, while rehearsal-based methods (MER \cite{riemer2019learning}, DER++ \cite{buzzega2020dark}, NC-FSCIL \cite{yang2023neural}, SLCA \cite{zhang2023slca}, Fs-Aug \cite{li2024continual}, MAGR \cite{zhou2024magr}, ASAL \cite{zhou2025continual}) improve stability yet fail to balance modality completion and cross-task retention.  
BriMA consistently achieves the best performance across all datasets and missing rates, surpassing the second-best baseline by 3.8\%, 4.5\%, and 5.2\% in average SRCC on RG, Fis-V, and FS1000, respectively. As $\beta$ rises from 10\% to 50\%, our relative improvement further expands by up to 7.1\%, demonstrating that the proposed MBI and MRO jointly provide robust generalization and stable scoring under increasingly severe modality degradation.

\myPara{Computational Efficiency.}
Due to its lightweight residual reconstruction design, BriMA maintains a favorable efficiency–performance trade-off. 
As shown in \cref{tab:efficiency}, BriMA achieves the largest SRCC improvement (+14.6\%) and error reductions (MSE $-33.4$\%, RL2 $-23.1$\%) with only a modest increase in parameters (+0.10M), training time (+1.001 hours), and inference latency (+0.100 sample/s). 
These results show that the proposed mechanisms achieve substantial accuracy gains with limited computational overhead.

\begin{table}[t]
    \centering
    \small
    \setlength{\tabcolsep}{4pt}
    \caption{
    Efficiency and relative performance comparison on RG (average across all items).  
    Results are reported relative to \cite{xu2025language}. 
    }
    \label{tab:efficiency}
    \resizebox{\linewidth}{!}{
    \begin{tabular}{rrrrrrrrrr}
    \toprule
    \textbf{Method} &
    \makecell{\bf SRCC \\ \bf Gain} &
    \makecell{\bf MSE \\ \bf Drop} &
    \makecell{\bf RL2 \\ \bf Drop} &
    \makecell{\bf Params \\ \bf (M)} &
    \makecell{\bf Training \\ \bf Time (h)} &
    \makecell{\bf Inference \\ \bf (Sample/s)} &
    \makecell{\bf Storage \\ \bf (MB)} \\
    \midrule
    \rowcolor{teal!10} ST-MLAVL~\cite{xu2025language} & 0\% & 0\% & 0\% & 8.30 & 0.416 & 0.078 & 0.00 \\
    \rowcolor{orange!10} EWC~\cite{james2017ewc} & +3.1\% & +7.5\% & +5.8\% & 8.30 & 0.597 & 0.078 & 0.00 \\
    \rowcolor{yellow!10} NC-FSCIL~\cite{yang2023neural} & +5.2\% & +13.6\% & +9.3\% & 8.30 & 1.025  & 0.078 & 32.25 \\
    \rowcolor{yellow!10} MAGR~\cite{zhou2024magr} & +9.5\% & +22.8\% & +15.2\% & 8.40 & 1.154  & 0.080 & 32.25 \\
    \rowcolor{yellow!10} ASAL~\cite{zhou2025adaptive} & +11.2\% & +26.7\% & +19.0\% & 8.30 & 0.876  & 0.084 & 32.25 \\
    \rowcolor{yellow!10} \bf BriMA (Ours) & \textbf{+14.6\%} & \textbf{+33.4\%} & \textbf{+23.1\%} & 8.80 & 1.417  & 0.118 & 32.25 \\
    \bottomrule
    \end{tabular}
    }
\end{table}

\begin{table}[]
    \centering
    \small
    \setlength{\tabcolsep}{5pt}
    \caption{Ablation study on the RG dataset ($\beta=10\%$).}
    \label{tab:ablation}
    \resizebox{\linewidth}{!}{
    \begin{tabular}{
    cl
    S[table-format=2.3] S[table-format=2.3] S[table-format=2.3] S[table-format=2.3] S[table-format=2.3]
    S[table-format=2.2] S[table-format=2.2] S[table-format=2.2] S[table-format=2.2] S[table-format=2.2]
    S[table-format=2.3] S[table-format=2.3] S[table-format=2.3] S[table-format=2.3] S[table-format=2.3]
    }
    \toprule
    \multirow{2.5}{*}{\textbf{ID}} & \multirow{2.5}{*}{\textbf{Setting}} &
    \multicolumn{5}{c}{\textbf{SRCC ($\uparrow$)}} \\
    \cmidrule(lr){3-7}
    & & \multicolumn{1}{c}{Ball} & \multicolumn{1}{c}{Clubs} & \multicolumn{1}{c}{Hoop} & \multicolumn{1}{c}{Ribbon} & \multicolumn{1}{c}{Avg.} \\
    \midrule
    \rowcolor{yellow!10}
    1 & Ours &
    0.648 & 0.788 & 0.710 & 0.836 & 0.746 \\
    \rowcolor{orange!10}
    2 & Ours w/o MBI &
    0.471 & 0.660 & 0.693 & 0.742 & 0.640 \\
    \rowcolor{orange!10}
    3 & Ours w/o Bridge  &
    0.640 & 0.727 & 0.656 & 0.796 & 0.730 \\
    \rowcolor{orange!10}
    4 & Ours w/o Candidate &
    0.639 & 0.755 & 0.628 & 0.773 & 0.699 \\
    \rowcolor{brown!10}
    5 & Ours w/o MRO &
    0.640 & 0.727 & 0.656 & 0.775 & 0.702 \\

    \midrule
    & & \multicolumn{5}{c}{\textbf{MSE ($\downarrow$)}} \\
    \cmidrule(lr){3-7}

    \rowcolor{yellow!10}
    1 & Ours &
    8.65 & 4.88 & 7.12 & 4.39 & 6.26 \\
    \rowcolor{orange!10}
    2 & Ours w/o MBI &
    10.70 & 6.80 & 8.03 & 9.34 & 8.72 \\
    \rowcolor{orange!10}
    3 & Ours w/o Bridge  &
    7.39 & 6.82 & 8.34 & 5.19 & 6.94 \\
    \rowcolor{orange!10}
    4 & Ours w/o Candidate &
    7.76 & 6.57 & 9.20 & 5.08 & 7.15 \\
    \rowcolor{brown!10}
    5 & Ours w/o MRO &
    7.39 & 6.82 & 8.34 & 5.51 & 7.02 \\

    \midrule
    & & \multicolumn{5}{c}{\textbf{RL2 ($\downarrow$)}} \\
    \cmidrule(lr){3-7}

    \rowcolor{yellow!10}
    1 & Ours &
    2.310 & 2.291 & 2.681 & 1.641 & 2.231 \\
    \rowcolor{orange!10}
    2 & Ours w/o MBI &
    2.857 & 3.190 & 3.024 & 3.495 & 3.142 \\
    \rowcolor{orange!10}
    3 & Ours w/o Bridge  &
    1.974 & 3.197 & 3.140 & 1.941 & 2.563 \\
    \rowcolor{orange!10}
    4 & Ours w/o Candidate &
    2.072 & 3.083 & 3.461 & 1.900 & 2.629 \\
    \rowcolor{brown!10}
    5 & Ours w/o MRO &
    1.974 & 3.197 & 3.140 & 2.060 & 2.593 \\

    \bottomrule
    \end{tabular}
    }
\end{table}

\subsection{Ablation Study}

\myPara{Core Modules.}
As shown in \cref{tab:ablation}, removing either MBI or MRO notably degrades performance, verifying that both modules are critical to handle evolving modality imbalance. Without MBI (Row~2), SRCC drops by $10.6\%$ and MSE rises by $39.3\%$, indicating the importance of stable bridging-based imputation. Excluding MRO (Row~5) yields weaker score consistency across tasks, reflected by a $16.2\%$ increase in RL2. 
Further ablations within MBI reveal complementary roles of the bridge and candidate selection mechanisms. Removing the bridge (Row~3) leads to higher reconstruction bias and temporal instability, while omitting candidate selection (Row~4) causes uneven memory sampling and degraded score correlation. Both components contribute jointly to modality-consistent reconstruction, validating the necessity of the full MBI design.

\begin{figure}
    \centering
    \includegraphics[width=\linewidth]{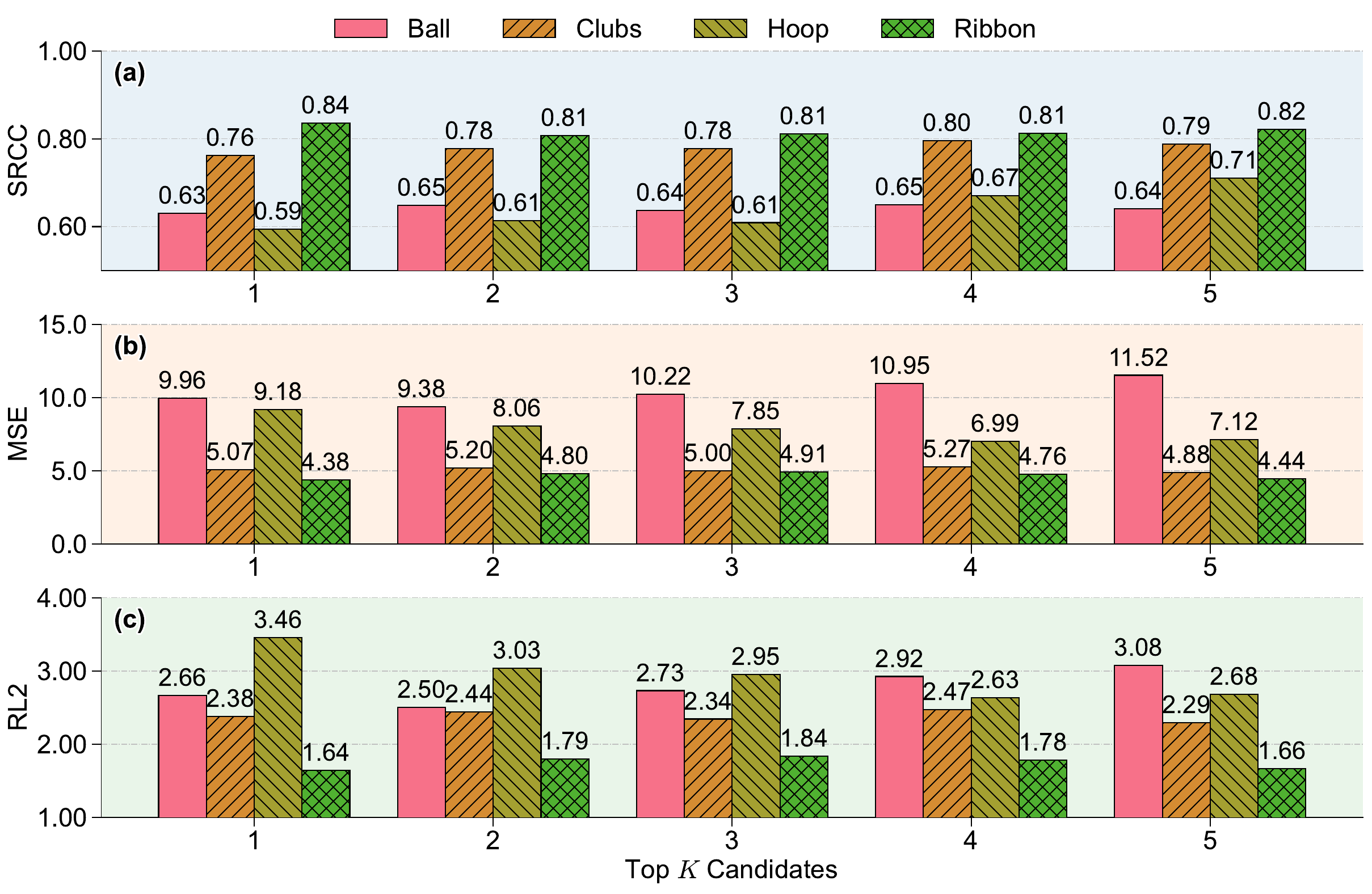}
    \caption{Results of different candidates.}
    \label{fig:topk}
\end{figure}

\myPara{Effect of Different Candidates.}
\cref{fig:topk} analyzes the impact of the retrieval size $K$ in the bridging imputation.  
When retrieval is disabled ($K=0$, Row~4 in \cref{tab:ablation}), SRCC notably drops and both MSE and RL2 increase, demonstrating that retrieved exemplars are crucial for stabilizing imputation.  
Introducing retrieval ($K>0$) consistently improves all metrics, confirming that exemplar-guided residual correction mitigates reconstruction bias.  
Across actions, the optimal $K$ varies slightly (e.g., $K{=}2$ for Ball, $K{=}4$ for Hoop, $K{=}5$ for Ribbon), but the differences remain marginal ($<2\%$ SRCC), indicating that once retrieval is enabled, the generative bridge dominates reconstruction quality rather than the exact number of exemplars.

\begin{figure}
    \centering
    \includegraphics[width=\linewidth]{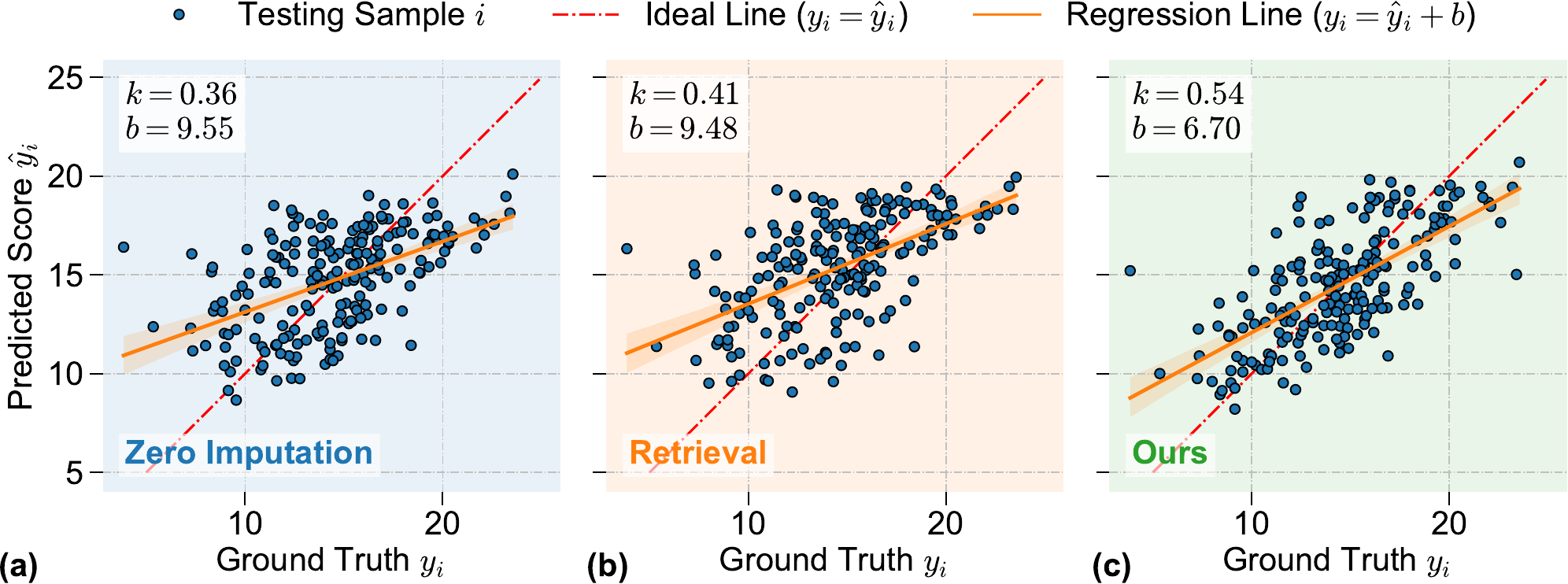}
    \caption{Results of different imputation strategies.}
    \label{fig:imputation}
    \phantomsubcaption\label{fig:imputation-a}
    \phantomsubcaption\label{fig:imputation-b}
    \phantomsubcaption\label{fig:imputation-c}
    \vspace{-0.35cm}
\end{figure}

\myPara{Effect of Different Imputation Strategies.}
\cref{fig:imputation} compares different imputation strategies on the RG dataset.  
The zero imputation baseline (see \cref{fig:imputation-a}) shows a weak correlation ($\hat{y}_i = 0.36y_i + 0.55$) and large variance, implying severe underfitting and bias accumulation.  
Retrieval-based imputation (see \cref{fig:imputation-b}) achieves moderate improvement ($\hat{y}_i = 0.41y_i + 9.48$) by incorporating exemplar information but still suffers from inconsistent scaling and offset errors.  
In contrast, our method (see \cref{fig:imputation-c}) produces the strongest linearity ($\hat{y}_i = 0.54y_i + 0.70$) with minimal deviation, demonstrating that the proposed bridge-guided imputation effectively aligns reconstructed scores with ground truth while maintaining stable distribution across tasks.

\subsection{Visualizations}

\begin{figure}
    \centering
    \includegraphics[width=\linewidth]{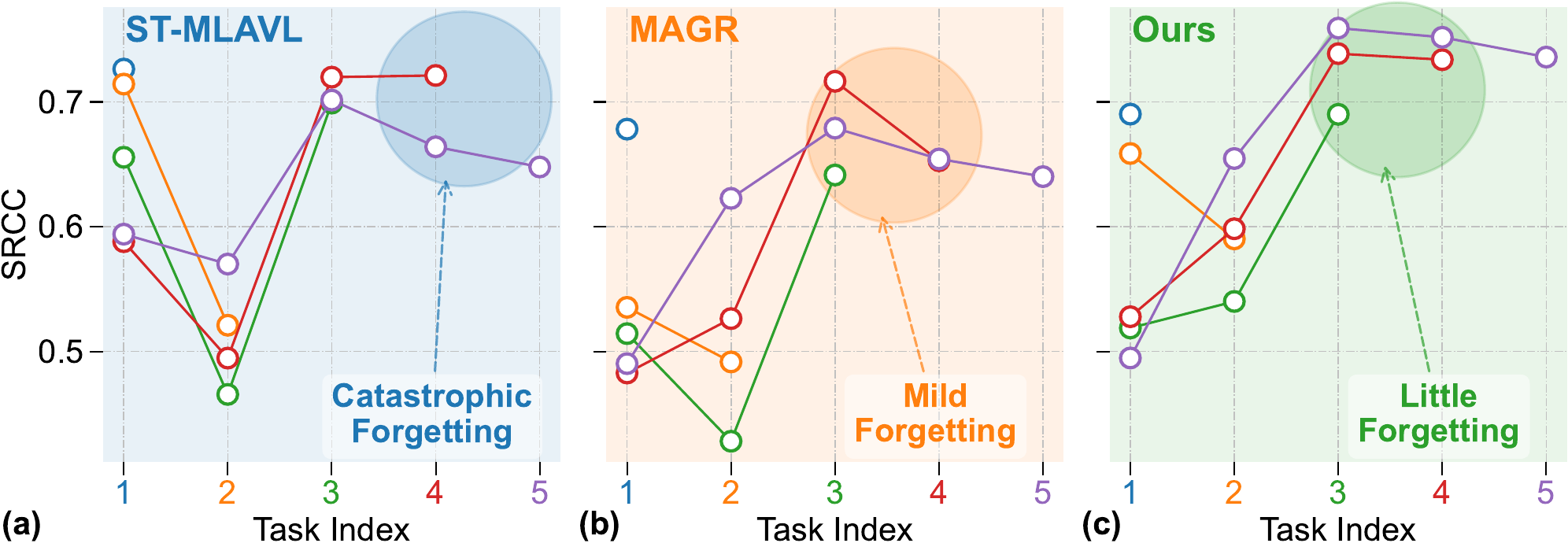}
    \caption{Task-wise SRCC performance comparison.}
    \label{fig:forgetting}
    \phantomsubcaption\label{fig:forgetting-a}
    \phantomsubcaption\label{fig:forgetting-b}
    \phantomsubcaption\label{fig:forgetting-c}
\end{figure}

\myPara{Forgetting Mitigation.}
To examine whether forgetting is effectively mitigated, we visualize task-wise SRCC performance in \cref{fig:forgetting} on RG.  
In \cref{fig:forgetting-a}, ST-MLAVL~\cite{xu2025language} suffers from severe forgetting, where after training on the fifth task, the performance on the fourth task drops noticeably compared to that after the fourth task training.  
With MAGR \cite{zhou2024magr} (see \cref{fig:forgetting-b}), forgetting is partially alleviated; the fourth-task performance remains stable between the fourth and fifth training stages, although mild degradation still appears on the third task.  
In contrast, our method (see \cref{fig:forgetting-c}) shows slight forgetting, exhibiting a stable SRCC increase across tasks 3 and 4.  
This indicates that our method not only preserves prior task knowledge but also leverages new data to enhance earlier task performance.

\begin{figure}
    \centering
    \includegraphics[width=\linewidth]{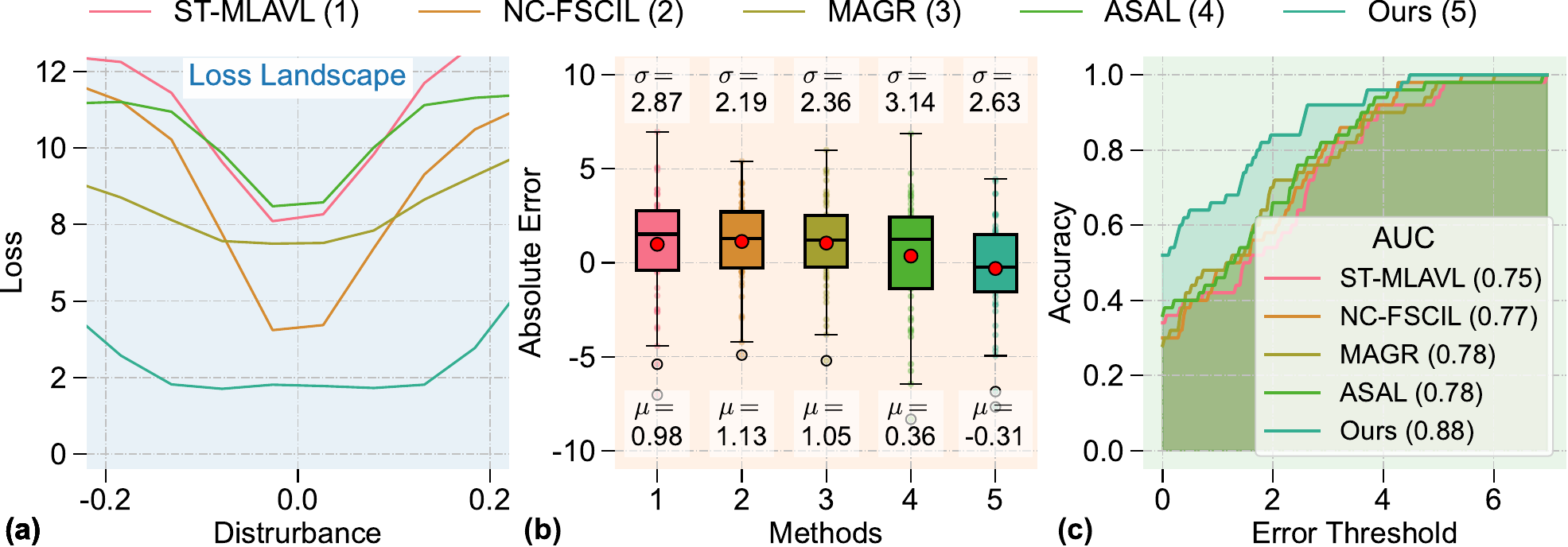}
    \caption{Performance comparisons: (a) loss landscape, (b) error box, (c) accuracy Area Under the Curve (AUC).}
    \label{fig:cmp}
    \phantomsubcaption\label{fig:cmp-a}
    \phantomsubcaption\label{fig:cmp-b}
    \phantomsubcaption\label{fig:cmp-c}
    \vspace{-0.35cm}
\end{figure}

\myPara{Flat Minima.}
To analyze generalization stability, we perturb model weights ten times and average the resulting loss surfaces across tasks.  
\cref{fig:cmp-a} compares the loss landscapes of ST-MLAVL~\cite{xu2025language}, NC-FSCIL~\cite{yang2023neural}, MAGR~\cite{zhou2024magr}, ASAL~\cite{zhou2025adaptive}, and our method.  
Our model exhibits a flatter and lower loss basin than the others, indicating better convergence stability and stronger generalization under CL.

\myPara{Error Analysis.}
As shown in \cref{fig:cmp-b}, our method achieves the lowest mean absolute error ($-$0.31) among all compared approaches, substantially outperforming ST-MLAVL~\cite{xu2025language} (0.98), NC-FSCIL~\cite{yang2023neural} (1.13), MAGR~\cite{zhou2024magr} (0.84), and ASAL~\cite{zhou2025adaptive} (0.62). 
It also yields the most stable predictions with a standard deviation of 2.63, indicating strong robustness to task transitions.  
\cref{fig:cmp-c} further shows AUC improvements, where ours (0.88) clearly surpasses ST-MLAVL (0.75), NC-FSCIL (0.77), MAGR (0.78), and ASAL (0.78), confirming superior accuracy and generalization across continual AQA tasks.

\begin{figure}
    \centering
    \tiny\sf
    \setlength{\tabcolsep}{1pt}
    \newcolumntype{C}[1]{>{\centering\arraybackslash}m{#1}}
    \begin{tabular}{C{2mm} C{0.195\linewidth} C{0.195\linewidth} C{0.195\linewidth} C{0.35\linewidth}}
    \rowcolor{orange!10}     &  First Frame & Middle Frame & Last Frame & Outputs  \\
    \rowcolor{orange!10}  \rotatebox{90}{Ball \#022}   & \includegraphics[width=\linewidth,clip,trim=250 0 250 0]{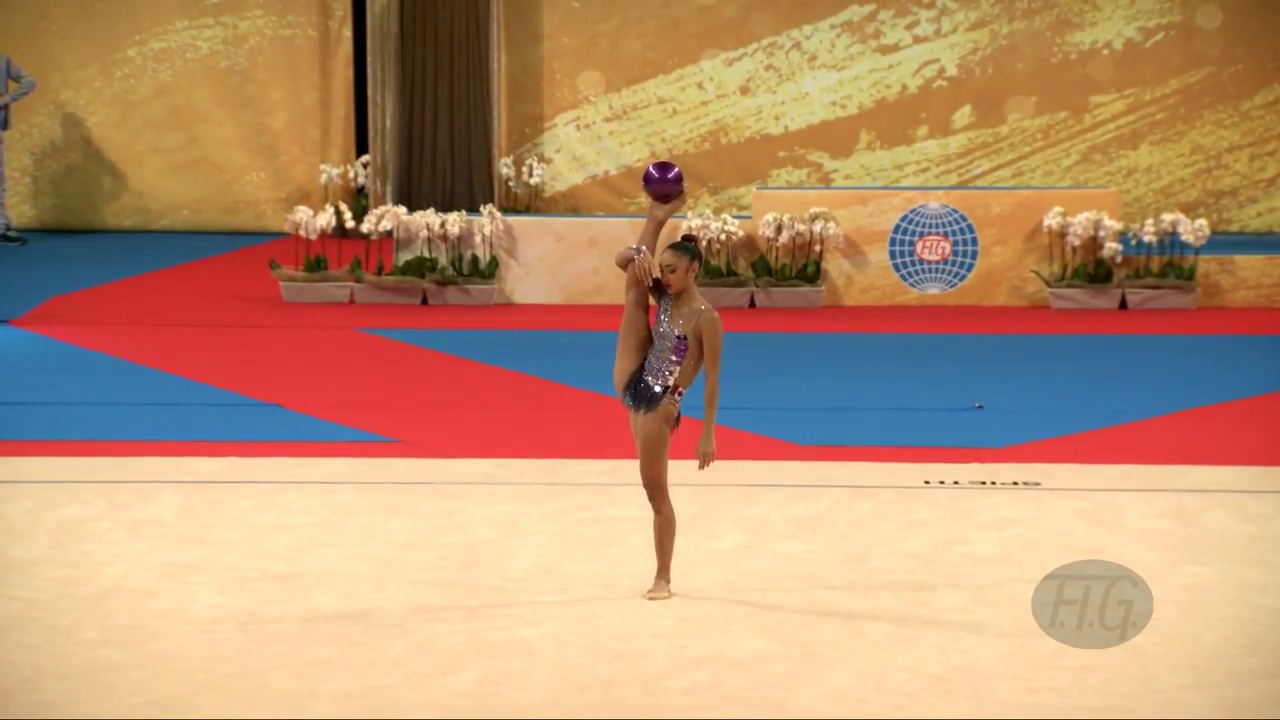} 
    & \includegraphics[width=\linewidth,clip,trim=250 0 250 0]{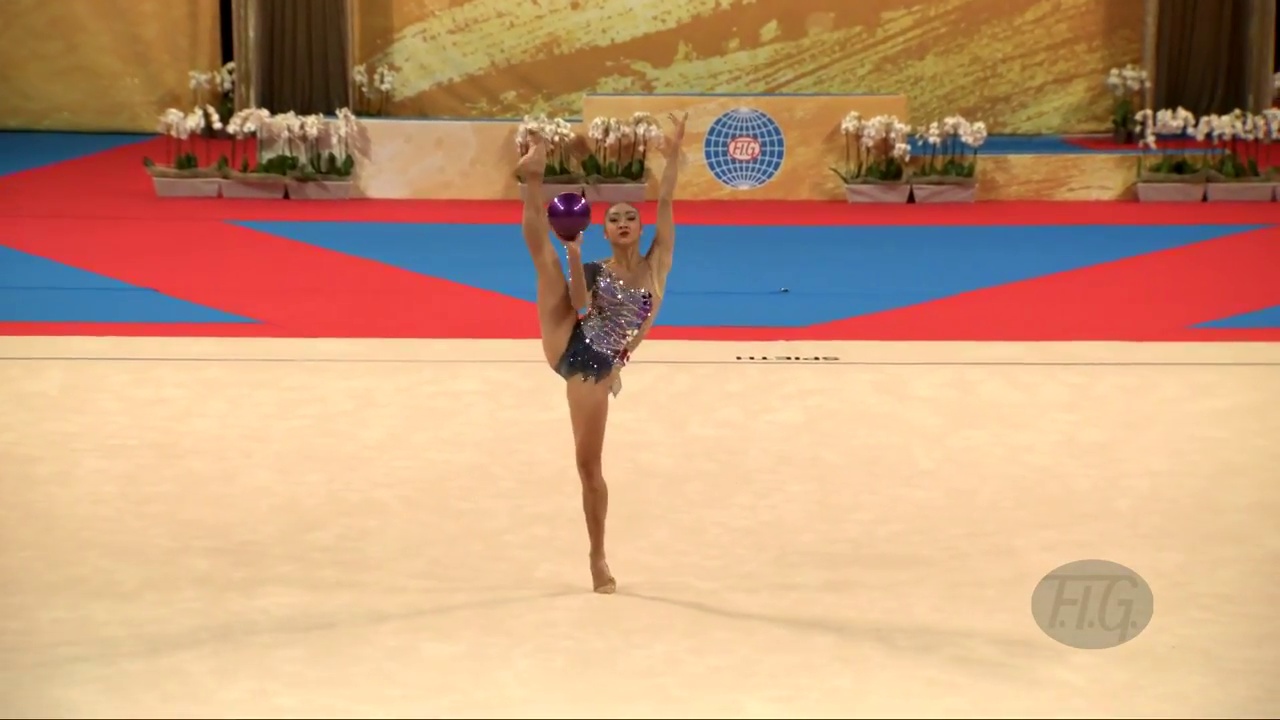} 
    & \includegraphics[width=\linewidth,clip,trim=250 0 250 0]{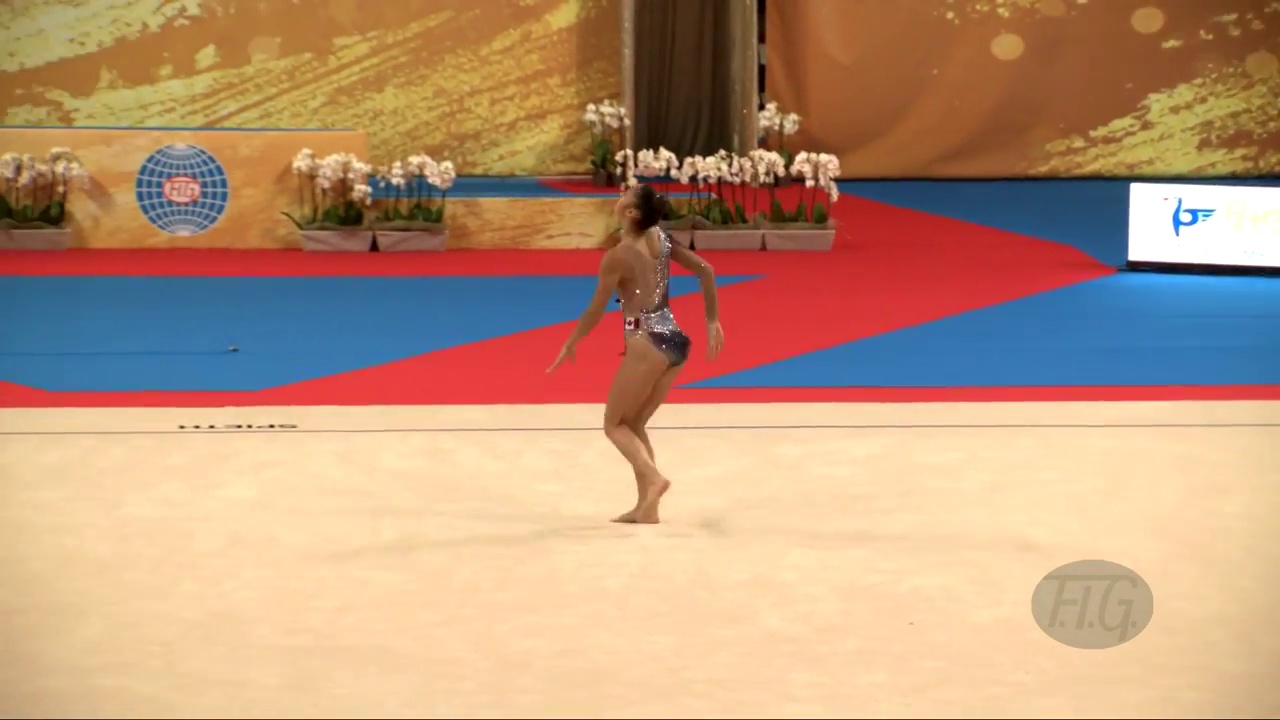} 
    & \includegraphics[width=\linewidth]{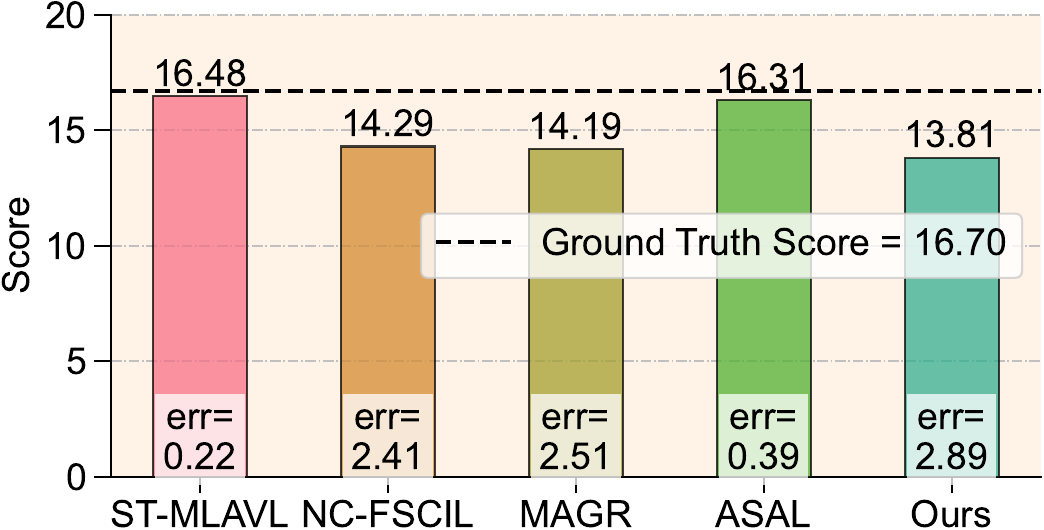} \\
    \end{tabular}
    \caption{Case study. See Appendix \cref{fig:supp-case_study} for more cases.}
    \label{fig:case_study}
    \vspace{-0.35cm}
\end{figure}

\myPara{Case Study.}
\cref{fig:case_study} presents a rhythmic gymnastics ball routine (Ball \#022) used to evaluate the scoring accuracy of different methods.  
The left part shows key frames, while the right part compares predicted scores from ST-MLAVL~\cite{xu2025language}, NC-FSCIL~\cite{yang2023neural}, MAGR~\cite{zhou2024magr}, ASAL~\cite{zhou2025adaptive}, and ours against the ground truth (16.70).  
ST-MLAVL and ASAL achieve close predictions (16.48 and 16.31), while NC-FSCIL and MAGR underestimate (14.29 and 14.19).  
Our method produces the most consistent visual–score alignment, demonstrating robust perception of fine-grained motions and superior temporal reasoning in complex routines.
 \section{Conclusion and Discussion}
In this work, we present BriMA, a novel approach to multi-modal continual action quality assessment under non-stationary modality imbalance.  
BriMA integrates memory-guided bridging imputation and modality-aware replay optimization to jointly address missing-modality reconstruction and cross-task stability.  
Extensive experiments across multiple datasets demonstrate that BriMA consistently outperforms state-of-the-art methods, achieving superior correlation, lower errors, and stronger resistance to forgetting.  
We believe BriMA offers a general paradigm for building more reliable multi-modal systems, advancing robust perception and evaluation in real-world scenarios.

\myPara{Limitations and Future Work.}
BriMA introduces moderate training overhead due to exemplar retrieval, residual bridging, and priority-based replay, achieving an efficiency–stability trade-off. Its pattern-level conditioning may face scalability challenges with larger modality sets. The framework does not explicitly model fine-grained temporal dynamics and assumes observable modality availability. Moreover, while the residual bridging design is conservative, handling weak modalities and unseen missing patterns at scale requires further investigation. Future work will explore more efficient retrieval, adaptive memory compression, lighter reconstruction modules, and temporal modeling to enhance scalability and robustness.

\myPara{Acknowledgment.}
This work was supported by the National Natural Science Foundation of China (NSFC, No.~62406160), the Beijing Natural Science Foundation (L247011), the Beijing Major Science and Technology Project (No.~Z251100008425003), and the China Postdoctoral Science Foundation (No.~2025M781489).
 
{
    \small
    \bibliographystyle{ieeenat_fullname}
    \bibliography{main}
}

\clearpage
\maketitlesupplementary
\setcounter{table}{0}
\setcounter{figure}{0}
\setcounter{section}{0}

\renewcommand{\theequation}{\textcolor{blue}{S}\arabic{equation}}
\renewcommand{\thetable}{S\arabic{table}}
\renewcommand{\thefigure}{\textcolor{blue}{S}\arabic{figure}}
\renewcommand{\thesection}{\textcolor{blue}{S}\arabic{section}}

\section{Training Procedure}
\label{sec:training}

During multi-modal continual AQA, BriMA sequentially learns from tasks 
$\{\mathcal{T}_1,\dots,\mathcal{T}_T\}$ under evolving modality availability.  
At session $t$, for each sample $(\boldsymbol{x}_{i,t},y_{i,t})$, the MBI module (see \cref{sec:mbi})  
reconstructs missing modality embeddings $\tilde{\boldsymbol{z}}_{i,t}^{m}$ using retrieved exemplars from the memory bank $\mathcal{B}_{t-1}$ and the imputation bridge in \cref{eq:bridge}, supervised by the reconstruction loss in \cref{eq:rec_loss}.  
The completed multi-modal features 
$\{\boldsymbol{z}_{i,t}^{m}\}_{m\in\mathcal{O}_{i,t}} \cup \{\tilde{\boldsymbol{z}}_{i,t}^{m}\}_{m\in\mathcal{M}_{i,t}}$  
are then fed into the scoring network $f_{\theta_f^t}$, which is jointly optimized with the reconstruction network $\theta_g$ using the objective in \cref{eq:objective}.  
During optimization, the MRO module (see \cref{sec:mro}) samples replay instances from $\mathcal{B}_{t-1}$ according to the priority $q_i$ in \cref{eq:priority} and regularizes temporal prediction drift via the consistency loss in \cref{eq:mem_loss}.  
After convergence on $\mathcal{T}_t$, MRO updates the memory bank to $\mathcal{B}_t$ using quantile-based, modality-complete selection in \cref{eq:selection}, which then supports the next session.

\begin{algorithm}[h!]
\caption{Training procedure of BriMA}
\label{alg:training}
\SetAlgoLined
\KwIn{Task sequence $\{\mathcal{T}_1,\dots,\mathcal{T}_T\}$, initial memory $\mathcal{B}_0=\emptyset$}
\KwOut{Trained parameters $\theta_f,\theta_g$}
\For{$t=1$ \KwTo $T$}{
    \tcp{Stage 1: Memory-guided bridging imputation}
    \ForEach{sample $(\boldsymbol{x}_{i,t},y_{i,t}) \in \mathcal{T}_t$}{
        Encode observed modalities to obtain $\{\boldsymbol{z}_{i,t}^{m}\}_{m\in\mathcal{O}_{i,t}}$\;
        Retrieve top-$K$ exemplar features from $\mathcal{B}_{t-1}$ using \cref{eq:retrieval}\;
        Build task indicator $\boldsymbol{r}_{i,t}$ and conditioning $\boldsymbol{c}_t^{m}$ via \cref{eq:mask,eq:task_indicator}\;
        Compute imputed features $\tilde{\boldsymbol{z}}_{i,t}^{m}$ for $m\in\mathcal{M}_{i,t}$ using the bridge \cref{eq:bridge}\;
    }
    
    \tcp{Stage 2: Joint optimization with modality-aware replay}
    \While{not converged}{
        Sample a mini-batch from $\mathcal{T}_t$ (with completed features)\;
        Sample a replay batch from $\mathcal{B}_{t-1}$\;
        For each replay sample, compute modality distortion $d_i$ and score drift $\Delta y_i$,
        then priority $q_i$ as in \cref{eq:priority}\;
        Select high-priority replay instances based on $q_i$\;
        Compute $\mathcal{L}_{\text{score}}$ (see \cref{eq:caqa_objective}), $\mathcal{L}_{\text{rec}}$ (see \cref{eq:rec_loss}), and
        $\mathcal{L}_{\text{mem}}$ (see \cref{eq:mem_loss})\;
        Update $\theta_f,\theta_g$ by minimizing \cref{eq:objective}\;
    }
    
    \tcp{Stage 3: Memory update}
    Rank all samples in $\mathcal{T}_t$ by predicted scores $\hat{y}_{i,t}$ and partition into $Q$ quantile bins\;
    Within bins, iteratively select modality-complete samples using \cref{eq:selection} until $\mathcal{B}_t$ is filled\;
}
\Return{$\theta_f,\theta_g$}
\end{algorithm}

\section{Additional Implementation Details}

\myPara{Additional Dataset Details}
We conduct experiments on three large-scale multi-modal datasets: RG, Fis-V, and FS1000.  
The score ranges and evaluation dimensions of these datasets are summarized in \cref{tab:score_ranges}.  
\textbf{Rhythmic Gymnastics (RG)} \cite{zeng2020hybrid} contains 1,000 videos of rhythmic gymnastics performances involving four apparatuses: ball, clubs, hoop, and ribbon. Each video lasts about 1.6 minutes and is recorded at 25 fps. The dataset is split into 200 training and 50 evaluation videos per action type. Following prior works~\cite{xu2024vision,xu2025language}, we train separate models for each apparatus.  
\textbf{Figure Skating Video (Fis-V)} \cite{parmar2017learning} consists of 500 videos of ladies’ singles short program performances in figure skating, each approximately 2.9 minutes long and recorded at 25~fps. Following the official split, 400 videos are used for training and 100 for testing. Each video is annotated with two official competition scores: the Total Element Score (TES) and the Program Component Score (PCS). Consistent with previous studies~\cite{xu2024vision,xu2025language}, we train separate models for each score type.  
\textbf{Figure Skating 1000 (FS1000)} \cite{xia2023skating} is a large-scale figure skating dataset comprising 1,000 training and 247 validation videos covering eight competition categories, including men’s, ladies’, and pairs’ short and free programs, as well as rhythm and free dances in ice dance. Each video contains roughly 5,000 frames at 25~fps. FS1000 provides TES, PCS, and five additional component scores: Skating Skills (SS), Transitions (TR), Performance (PE), Composition (CO), and Interpretation of Music (IN). It is the first figure skating dataset designed for audiovisual learning, promoting rule-consistent multi-modal modeling. Following prior works~\cite{xu2024vision,xu2025language}, we train separate models for each score type.

\begin{table*}[t]
    \centering
    \small
    \setlength{\tabcolsep}{6pt}
    \caption{Score ranges and evaluation dimensions for all datasets.}
    \label{tab:score_ranges}
    \begin{tabular}{lcccc}
        \toprule
        \textbf{Dataset} & \textbf{Subcategory} & \textbf{Score Dimension(s)} & \textbf{Range} & \textbf{Notes} \\
        \midrule
        \multirow{4}{*}{RG \cite{zeng2020hybrid}} 
        & Ball & Overall & 0 -- 25 & \multirow{4}{*}{Four rhythmic events} \\
        & Clubs & Overall & 0 -- 25 & \\
        & Hoop & Overall & 0 -- 25 & \\
        & Ribbon & Overall & 0 -- 25 & \\
        \midrule
        \multirow{2}{*}{Fis-V \cite{parmar2017learning}} 
        & TES & Technical Execution Score (TES) & 0 -- 45 & \multirow{2}{*}{Two judging dimensions} \\
        & PCS & Performance Component Score (PCS) & 0 -- 40 & \\
        \midrule
        \multirow{7}{*}{FS1000 \cite{xia2023skating}} 
        & TES & Technical Execution Score & 0 -- 130 & \multirow{7}{*}{Seven judging components} \\
        & PCS & Performance Component Score & 0 -- 60 & \\
        & SS & Skating Skills & 0 -- 10 & \\
        & TR & Transitions & 0 -- 10 & \\
        & PE & Performance & 0 -- 10 & \\
        & CO & Composition & 0 -- 10 & \\
        & IN & Interpretation & 0 -- 10 & \\
        \bottomrule
    \end{tabular}
\end{table*}

\begin{figure}[t]
    \centering
    \begin{overpic}[height=0.48\linewidth,clip,trim=10 0 10 0]{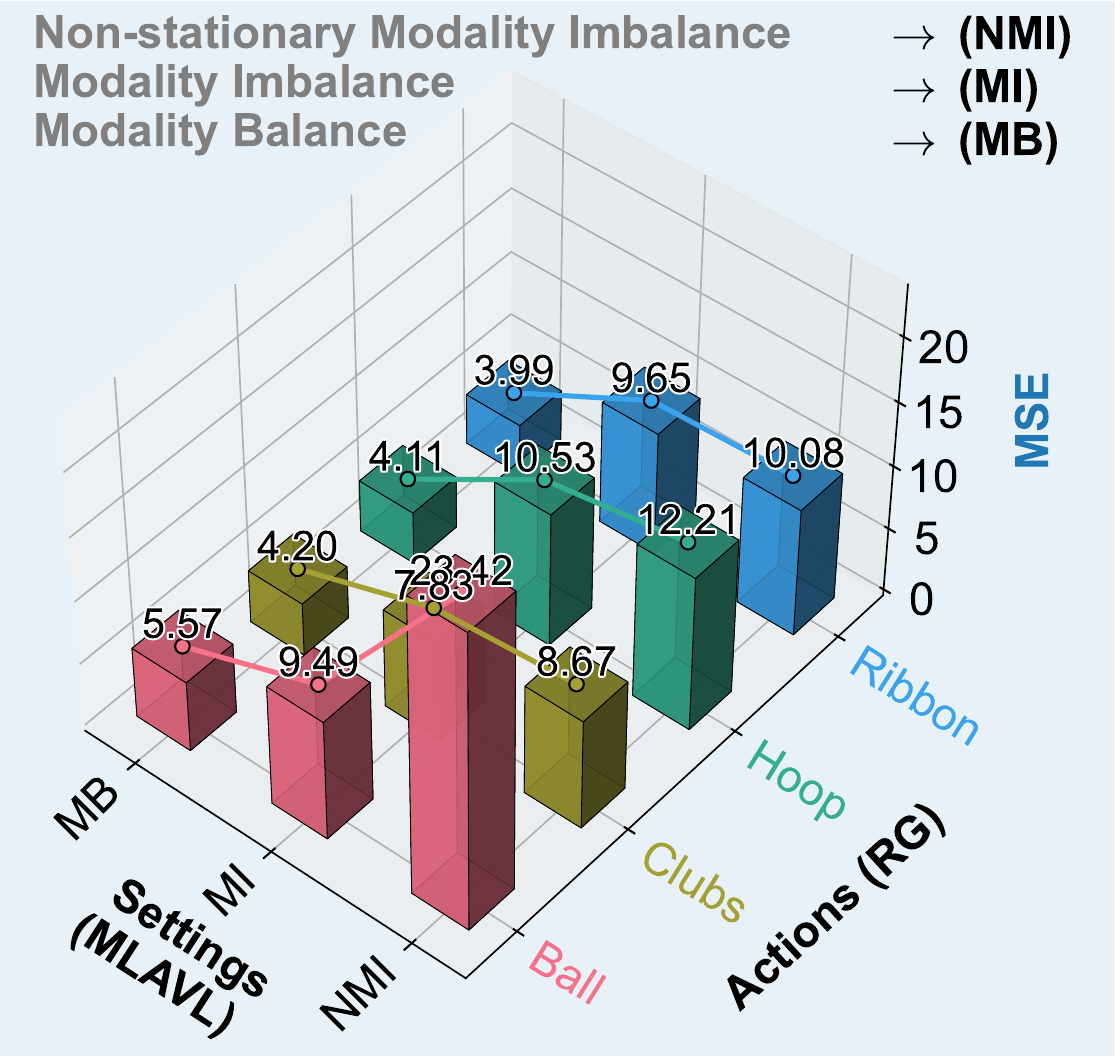}\put(90,0){
            \begin{tikzpicture}
                \node[fill=white, fill opacity=0, text opacity=1, inner sep=1pt] {\tiny\sf\textbf{(a)}};
            \end{tikzpicture}
        }\put(192,0){
            \begin{tikzpicture}
                \node[fill=white, fill opacity=0, text opacity=1, inner sep=1pt] {\tiny\sf\textbf{(b)}};
            \end{tikzpicture}
        }\end{overpic}\begin{overpic}[height=0.48\linewidth,clip,trim=10 0 10 0]{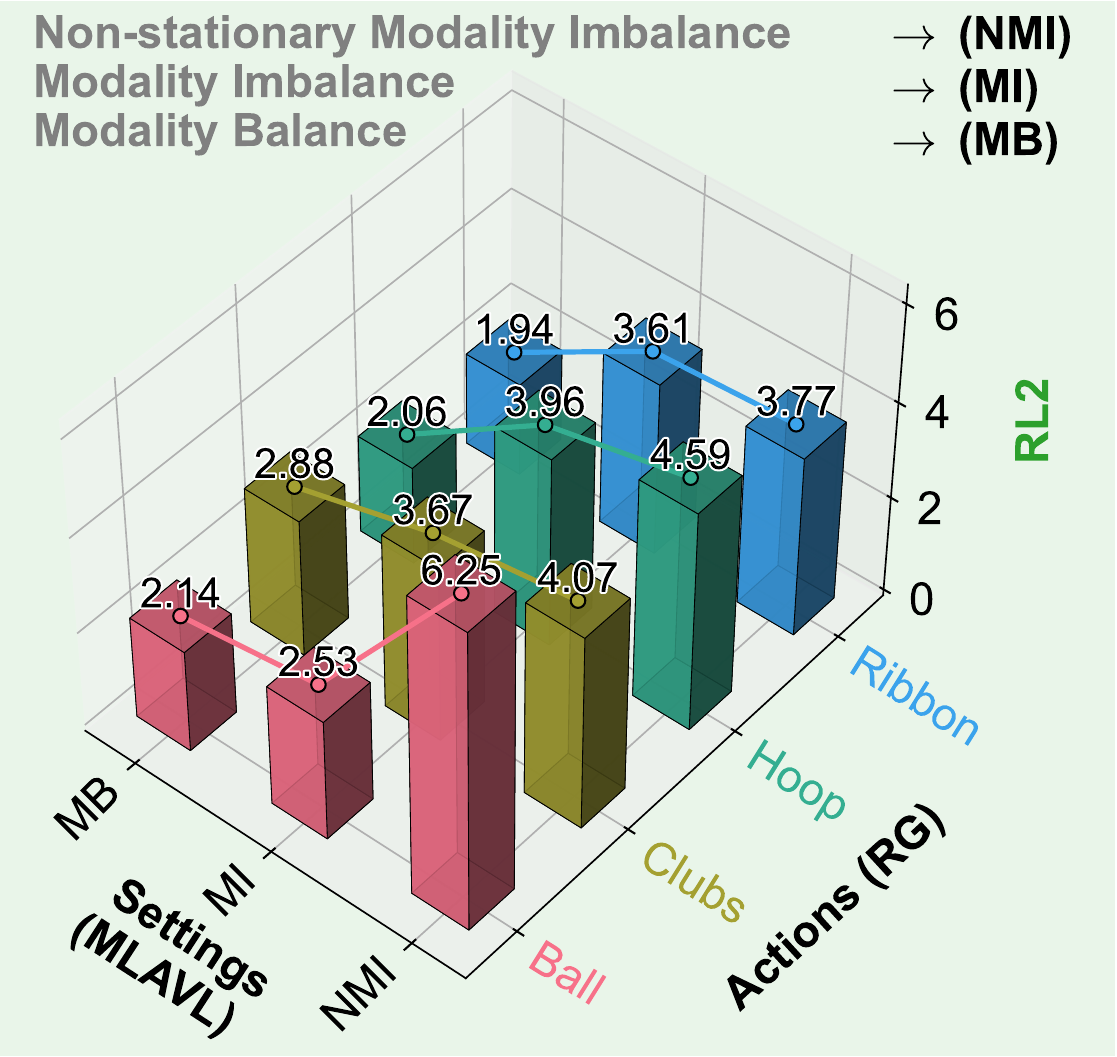}\end{overpic}
    \caption{Non-stationary modality imbalance significantly challenges the model performance: (a) L2 and (b) RL2.}
    \label{fig:supp-teaser}
    \phantomsubcaption\label{fig:supp-teaser-a}
    \phantomsubcaption\label{fig:supp-teaser-b}
\end{figure}

\myPara{Backbone Details.}
Following prior works \cite{xu2025language,xu2024vision,xia2023skating},  
we employ widely adopted pretrained encoders for the three modalities considered in this study.  
For RG and Fis-V, visual features are extracted using the Video Swin Transformer (VST) pretrained on Kinetics-600,  
and audio features are obtained using the Audio Spectrogram Transformer (AST) pretrained on AudioSet.  
For FS1000, we use the TimeSformer and AST features provided by \cite{xia2023skating}.  
Textual commentary is encoded using ViFi-CLIP, a CLIP model fine-tuned on Kinetics-400,  
applied to our curated prompt sets.  
All experiments adopt fixed clip sampling following the standard protocol of each dataset.

\section{Additional Results}

\myPara{Performance Drops Due to Non-Stationary Modality Imbalance.}  
\cref{fig:supp-teaser} quantitatively demonstrates the severe impact of evolving modality imbalance across the four rhythmic gymnastics actions.  
Under the balanced-modality setting, all actions achieve strong correlation (SRCC $\ge$ 0.82) and low error (MSE $\le$ 5.6, RL2 $\le$ 2.2), indicating stable and consistent scoring.  
When modality imbalance is introduced, both correlation and error metrics deteriorate sharply: SRCC drops by an average of 28.6\%, while MSE and RL2 increase by 78.1\% and 38.2\%, respectively.  
The degradation becomes catastrophic under the non-stationary modality imbalance setting, where missing modalities vary across sessions, leading to an average SRCC decline of 62.4\% compared with the balanced baseline.  
In this case, MSE grows nearly threefold and RL2 more than doubles, for example from 2.14 to 6.26 for the ``Ball'' routine.  
These results confirm that traditional AQA pipelines are highly vulnerable to modality instability.  
Even minor shifts in modality quality or availability can propagate through feature fusion, distort the latent representation $\boldsymbol{h}$, and cause nontrivial score drift.  
This emphasizes the necessity of explicitly addressing non-stationary modality imbalance in continual AQA, as tackled by our BriMA method.

\begin{table*}[]
    \centering
    \caption{Performance comparison on the FS1000 dataset (supplementary results of \cref{tab:fs1000}). 
    \textbf{Bold} values indicate the best results. 
    {\color{violet}Joint Training (JT)} and {\color{teal}Sequential Training (ST)} denote the upper and lower bounds, while {\color{orange}rehearsal-free} and {\color{yellow!80!black}rehearsal-based} methods represent different CL strategies. 
    $\uparrow$: higher is better; $\downarrow$: lower is better. 
    N/A indicates that the metric was not reported in the paper. 
    Average SRCC is computed using Fisher-$z$ transformation.}
    \label{tab:supp-fs1000}
    \small
    \setlength{\tabcolsep}{5pt}
    \resizebox{\linewidth}{!}{
    \begin{tabular}{rr
    S[table-format=1.3, input-symbols={\textbf}]
    S[table-format=1.3, input-symbols={\textbf}]
    S[table-format=1.3, input-symbols={\textbf}]
    S[table-format=1.3, input-symbols={\textbf}]
    S[table-format=1.3, input-symbols={\textbf}]
    S[table-format=1.3, input-symbols={\textbf}]
    S[table-format=1.3, input-symbols={\textbf}]
    S[table-format=1.3, input-symbols={\textbf}]
    S[table-format=2.2, input-symbols={\textbf}]
    S[table-format=2.2, input-symbols={\textbf}]
    S[table-format=2.2, input-symbols={\textbf}]
    S[table-format=2.2, input-symbols={\textbf}]
    S[table-format=2.2, input-symbols={\textbf}]
    S[table-format=2.2, input-symbols={\textbf}]
    S[table-format=2.2, input-symbols={\textbf}]
    S[table-format=2.2, input-symbols={\textbf}]
    S[table-format=1.3, input-symbols={\textbf}]
    S[table-format=1.3, input-symbols={\textbf}]
    S[table-format=1.3, input-symbols={\textbf}]
    S[table-format=1.3, input-symbols={\textbf}]
    S[table-format=1.3, input-symbols={\textbf}]
    S[table-format=1.3, input-symbols={\textbf}]
    S[table-format=1.3, input-symbols={\textbf}]
    S[table-format=1.3, input-symbols={\textbf}]
    c
    c
    }
    \toprule
    \multirow{2.5}{*}{\textbf{Method}} & \multirow{2.5}{*}{\textbf{Publisher}} & \multicolumn{8}{c}{\textbf{SRCC ($\uparrow$)}} & \multicolumn{8}{c}{\textbf{MSE ($\downarrow$)}}  & \multicolumn{8}{c}{\textbf{RL2 ($\downarrow$)}}  \\
    \cmidrule(lr){3-10}
    \cmidrule(lr){11-18}
    \cmidrule(lr){19-26}
    & & 
    \multicolumn{1}{c}{TES} & \multicolumn{1}{c}{PCS} & \multicolumn{1}{c}{SS} & \multicolumn{1}{c}{TR} & \multicolumn{1}{c}{PE} & \multicolumn{1}{c}{CO} & \multicolumn{1}{c}{IN} & \multicolumn{1}{c}{Avg.} & 
    \multicolumn{1}{c}{TES} & \multicolumn{1}{c}{PCS} & \multicolumn{1}{c}{SS} & \multicolumn{1}{c}{TR} & \multicolumn{1}{c}{PE} & \multicolumn{1}{c}{CO} & \multicolumn{1}{c}{IN} & \multicolumn{1}{c}{Avg.} & 
    \multicolumn{1}{c}{TES} & \multicolumn{1}{c}{PCS} & \multicolumn{1}{c}{SS} & \multicolumn{1}{c}{TR} & \multicolumn{1}{c}{PE} & \multicolumn{1}{c}{CO} & \multicolumn{1}{c}{IN} & \multicolumn{1}{c}{Avg.}  \\
    \midrule
    \rowcolor{violet!10} JT-MLAVL\cite{xu2025language}   & CVPR'25 & 0.92 & 0.89 & 0.90 & 0.90 & 0.88 & 0.89 & 0.88 & 0.90 & 64.89 & 6.39 & 0.23 & 0.24 & 0.50 & 0.25 & 0.26 & 10.39 & {N/A} & {N/A} & {N/A} & {N/A} & {N/A} & {N/A} & {N/A} & {N/A} \\
    \midrule
    \multicolumn{17}{l}{\it Modality Missing Rate $\beta=10\%$} \\
    \midrule
\rowcolor{teal!10} ST-MLAVL \cite{xu2025language}   &    CVPR'25 &  0.630 &  0.762 &  0.740 &  0.757 &  0.721 &  0.744 &  0.726 &  0.728 & 209.44 &  15.38 &   0.56 &   0.54 &   0.86 &   0.55 &   0.61 &  32.56 &  1.686 &  1.084 &  2.026 &  1.821 &  0.892 &  1.736 &  1.899 &  1.592 \\
\rowcolor{orange!10} SI \cite{zenke2017continual}   &    ICML'17 &  0.583 &  0.747 &  0.737 &  0.749 &  0.720 &  0.754 &  0.717 &  0.719 & 238.47 &  15.74 &   0.56 &   0.56 &   0.88 &   0.56 &   0.63 &  36.77 &  1.920 &  1.110 &  2.043 &  1.865 &  0.909 &  1.748 &  1.952 &  1.650 \\
\rowcolor{orange!10} EWC \cite{james2017ewc}        &    PNAS'17 &  0.639 &  0.715 &  0.743 &  0.740 &  0.730 &  0.743 &  0.689 &  0.716 & 358.87 &  22.05 &   0.66 &   0.76 &   0.90 &   0.71 &   0.81 &  54.97 &  2.889 &  1.554 &  2.377 &  2.555 &  0.934 &  2.232 &  2.506 &  2.150 \\
\rowcolor{orange!10} LwF \cite{li2017learning}      &   TPAMI'17 &  0.610 &  0.731 &  0.740 &  0.706 &  0.662 &  0.740 &  0.688 &  0.699 & 228.11 &  14.72 &   0.53 &   0.65 &   0.99 &   0.62 &   0.68 &  35.19 &  1.836 &  1.038 &  1.932 &  2.193 &  1.029 &  1.940 &  2.111 &  1.726 \\
\rowcolor{yellow!10} MER \cite{riemer2019learning}  &    ICLR'19 & \bf 0.708 &  0.775 &  0.727 &  0.762 &  0.713 &  0.753 &  0.733 &  0.739 & \bf 190.91 &  15.21 &   0.65 &   0.65 &   0.90 &   0.57 &   0.77 &  29.95 & \bf 1.537 &  1.072 &  2.375 &  2.195 &  0.938 &  1.783 &  2.405 &  1.758 \\
\rowcolor{yellow!10} DER++ \cite{buzzega2020dark}   & NeurIPS'20 &  0.672 &  0.780 &  0.767 &  0.762 & \bf 0.769 & \bf 0.777 &  0.746 &  0.755 & 201.08 &  14.56 &   0.50 &   0.56 & \bf 0.79 &   0.50 &   0.60 &  31.23 &  1.619 &  1.027 &  1.803 &  1.874 & \bf 0.823 &  1.577 &  1.872 &  1.513 \\
\rowcolor{yellow!10} NC-FSCIL \cite{yang2023neural} &    ICLR'23 &  0.659 &  0.767 &  0.759 &  0.781 &  0.732 &  0.761 &  0.747 &  0.746 & 211.11 &  14.29 &   0.50 &  0.48 &   0.81 &   0.54 & \bf 0.56 &  32.62 &  1.700 &  1.008 &  1.831 &  1.615 &  0.843 &  1.699 &  1.756 &  1.493 \\
\rowcolor{yellow!10} SLCA \cite{zhang2023slca}      &    ICCV'23 &  0.633 &  0.770 &  0.770 &  0.744 &  0.739 &  0.759 &  0.746 &  0.740 & 352.76 &  19.63 &   0.64 &   0.75 &   1.01 &   0.67 &   0.82 &  53.75 &  2.840 &  1.384 &  2.316 &  2.500 &  1.047 &  2.118 &  2.557 &  2.109 \\
\rowcolor{yellow!10} Fs-Aug \cite{li2024continual}  &   TCSVT'24 &  0.646 &  0.750 &  0.741 &  0.749 &  0.739 &  0.765 &  0.778 &  0.741 & 222.69 &  17.88 &   0.58 &   0.57 &   0.81 &   0.56 &   0.58 &  34.81 &  1.793 &  1.261 &  2.112 &  1.910 &  0.837 &  1.751 &  1.796 &  1.637 \\
\rowcolor{yellow!10} MAGR \cite{zhou2024magr}       &    ECCV'24 &  0.682 &  0.765 &  0.748 &  0.759 &  0.700 &  0.742 &  0.704 &  0.730 & 211.79 &  14.26 &   0.53 &   0.53 &   0.88 &   0.62 &   0.64 &  32.75 &  1.705 &  1.005 &  1.938 &  1.769 &  0.910 &  1.948 &  1.987 &  1.609 \\
\rowcolor{yellow!10} ASAL \cite{zhou2025adaptive}   &    TVCG'25 &  0.681 &  0.761 &  0.745 &  0.743 &  0.739 &  0.762 &  0.735 &  0.739 & 191.01 &  15.00 &   0.57 &   0.61 &   0.88 &   0.55 &   0.66 & \bf 29.90 &  1.538 &  1.058 &  2.056 &  2.035 &  0.908 &  1.717 &  2.051 &  1.623 \\
\rowcolor{yellow!10} 
BriMA (Ours) & -- 
& 0.676 & \bf 0.794 & \bf 0.797 & \bf 0.790 & 0.736 & 0.773 & \bf 0.784 & \bf 0.756 
& 206.61 & \bf 12.18 & \bf 0.44 & \bf 0.46 &  0.82 & \bf 0.49 &  0.57 & 33.35
& \bf 1.663 & \bf 0.859 & \bf 1.596 & \bf 1.558 &  0.852 & \bf 1.540 & \bf 1.674 & \bf 1.441 \\
    \midrule
    \multicolumn{17}{l}{\it Modality Missing Rate $\beta=25\%$} \\
    \midrule
\rowcolor{teal!10} ST-MLAVL \cite{xu2025language}   &    CVPR'25 &  0.659 &  0.748 &  0.738 &  0.712 &  0.731 &  0.745 &  0.725 &  0.724 & 229.63 &  15.50 &   0.57 &   0.61 & \bf 0.80 &   0.55 &   0.61 &  35.47 &  1.849 &  1.093 &  2.078 &  2.042 & \bf 0.831 &  1.730 &  1.895 &  1.645 \\
\rowcolor{orange!10} SI \cite{zenke2017continual}   &    ICML'17 &  0.582 &  0.766 &  0.738 &  0.763 &  0.720 &  0.743 &  0.730 &  0.725 & 261.25 &  14.43 &   0.54 & \bf 0.52 &   0.84 &   0.57 &   0.62 &  39.83 &  2.103 &  1.018 &  1.955 & \bf 1.751 &  0.871 &  1.779 &  1.939 &  1.631 \\
\rowcolor{orange!10} EWC \cite{james2017ewc}        &    PNAS'17 &  0.594 &  0.757 &  0.729 &  0.737 &  0.734 &  0.690 &  0.702 &  0.710 & 374.76 &  17.63 &   0.65 &   0.61 &   0.95 &   0.69 &   0.80 &  56.58 &  3.017 &  1.243 &  2.341 &  2.062 &  0.989 &  2.161 &  2.476 &  2.041 \\
\rowcolor{orange!10} LwF \cite{li2017learning}      &   TPAMI'17 &  0.606 &  0.681 &  0.615 &  0.664 &  0.622 &  0.551 &  0.632 &  0.626 & 220.81 &  17.29 &   1.00 &   0.73 &   1.03 &   1.00 &   0.84 &  34.67 &  1.778 &  1.219 &  3.623 &  2.465 &  1.064 &  3.137 &  2.614 &  2.271 \\
\rowcolor{yellow!10} MER \cite{riemer2019learning}  &    ICLR'19 & \bf 0.676 &  0.726 &  0.685 &  0.699 &  0.734 &  0.710 &  0.736 &  0.710 & 206.59 &  17.88 &   0.63 &   0.79 &   0.90 &   0.66 &   0.60 &  32.58 &  1.663 &  1.261 &  2.303 &  2.637 &  0.937 &  2.068 &  1.856 &  1.818 \\
\rowcolor{yellow!10} DER++ \cite{buzzega2020dark}   & NeurIPS'20 &  0.610 &  0.744 & \bf 0.772 &  0.754 &  0.728 &  0.748 &  0.752 &  0.733 & 270.86 &  16.30 & \bf 0.50 &   0.53 &   0.82 &   0.56 &   0.59 &  41.45 &  2.181 &  1.149 & \bf 1.820 &  1.763 &  0.851 &  1.753 &  1.840 &  1.623 \\
\rowcolor{yellow!10} NC-FSCIL \cite{yang2023neural} &    ICLR'23 &  0.637 &  0.751 &  0.743 &  0.741 &  0.727 &  0.757 &  0.728 &  0.728 & 251.10 &  15.95 &   0.57 &   0.56 &   0.80 &   0.53 &   0.61 &  38.59 &  2.022 &  1.124 &  2.078 &  1.887 &  0.833 &  1.673 &  1.883 &  1.643 \\
\rowcolor{yellow!10} SLCA \cite{zhang2023slca}      &    ICCV'23 &  0.591 &  0.743 &  0.729 &  0.740 &  0.739 &  0.727 &  0.740 &  0.719 & 384.94 &  20.15 &   0.70 &   0.69 &   0.88 &   0.74 &   0.74 &  58.41 &  3.099 &  1.421 &  2.539 &  2.323 &  0.913 &  2.322 &  2.313 &  2.133 \\
\rowcolor{yellow!10} Fs-Aug \cite{li2024continual}  &   TCSVT'24 &  0.624 &  0.773 &  0.712 &  0.760 &  0.733 &  0.732 &  0.719 &  0.725 & 249.25 &  16.14 &   0.62 &   0.53 &   0.86 &   0.59 &   0.66 &  38.38 &  2.007 &  1.138 &  2.262 &  1.785 &  0.889 &  1.857 &  2.038 &  1.711 \\
\rowcolor{yellow!10} MAGR \cite{zhou2024magr}       &    ECCV'24 &  0.604 &  0.750 &  0.678 &  0.746 &  0.725 &  0.733 &  0.719 &  0.711 & 258.14 &  15.40 &   0.61 &   0.56 &   0.84 &   0.57 &   0.59 &  39.53 &  2.078 &  1.086 &  2.223 &  1.883 &  0.869 &  1.787 &  1.841 &  1.681 \\
\rowcolor{yellow!10} ASAL \cite{zhou2025adaptive}   &    TVCG'25 &  0.634 &  0.745 &  0.749 &  0.720 & \bf 0.743 &  0.731 &  0.737 &  0.724 & 230.75 &  15.28 &   0.56 &   0.63 &   0.85 &   0.58 &   0.60 &  35.61 &  1.858 &  1.077 &  2.044 &  2.103 &  0.883 &  1.830 &  1.875 &  1.667 \\
\rowcolor{yellow!10} BriMA (Ours) & -- 
& \bf 0.673 & \bf 0.774 & 0.751 & \bf 0.776 & 0.726 & \bf 0.767 & \bf 0.760 & \bf 0.740
& \bf 202.14 & \bf 14.05 & \bf 0.54 & 0.57 & \bf 0.83 & \bf 0.49 & \bf 0.56 & \bf 31.35
& \bf 1.627 & \bf 0.991 & 1.962 & 1.925 & \bf 0.858 & \bf 1.548 & \bf 1.727 & \bf 1.572 \\
    \midrule
    \multicolumn{17}{l}{\it Modality Missing Rate $\beta=50\%$} \\
    \midrule
\rowcolor{teal!10} ST-MLAVL \cite{xu2025language}   &    CVPR'25 &  0.524 &  0.727 &  0.656 &  0.689 &  0.653 &  0.693 &  0.686 &  0.665 & 359.86 &  16.10 &   0.67 &   0.71 &   0.98 &   0.67 &   0.68 &  54.24 &  2.897 &  1.135 &  2.424 &  2.385 &  1.012 &  2.106 &  2.125 &  2.012 \\
\rowcolor{orange!10} SI \cite{zenke2017continual}   &    ICML'17 &  0.514 &  0.678 &  0.691 &  0.681 &  0.660 &  0.682 &  0.665 &  0.656 & 303.64 &  19.99 &   0.63 &   0.66 &   0.92 &   0.68 & \bf 0.67 &  46.74 &  2.445 &  1.409 &  2.270 &  2.217 &  0.951 &  2.139 & \bf 2.092 &  1.932 \\
\rowcolor{orange!10} EWC \cite{james2017ewc}        &    PNAS'17 &  0.520 &  0.697 &  0.693 &  0.667 &  0.638 &  0.612 &  0.598 &  0.635 & 381.88 &  20.17 &   0.67 &   0.76 &   1.09 &   0.97 &   0.91 &  58.06 &  3.074 &  1.422 &  2.413 &  2.552 &  1.131 &  3.052 &  2.832 &  2.354 \\
\rowcolor{orange!10} LwF \cite{li2017learning}      &   TPAMI'17 &  0.502 &  0.614 &  0.554 &  0.678 &  0.559 &  0.571 &  0.588 &  0.583 & 289.66 &  25.50 &   1.18 &   0.68 &   1.32 &   1.03 &   0.99 &  45.76 &  2.332 &  1.797 &  4.275 &  2.268 &  1.366 &  3.253 &  3.082 &  2.625 \\
\rowcolor{yellow!10} MER \cite{riemer2019learning}  &    ICLR'19 &  0.579 &  0.681 &  0.631 &  0.605 &  0.614 &  0.636 &  0.638 &  0.627 & 247.80 &  22.81 &   0.96 &   1.36 &   1.16 &   0.89 &   0.81 &  39.40 &  1.995 &  1.608 &  3.481 &  4.566 &  1.200 &  2.794 &  2.528 &  2.596 \\
\rowcolor{yellow!10} DER++ \cite{buzzega2020dark}   & NeurIPS'20 &  0.577 &  0.731 &  0.674 &  0.683 &  0.659 &  0.689 &  0.644 &  0.668 & 300.21 &  19.12 &   0.66 &   0.67 &   1.07 &   0.67 &   0.78 &  46.17 &  2.417 &  1.348 &  2.387 &  2.244 &  1.114 &  2.092 &  2.417 &  2.003 \\
\rowcolor{yellow!10} NC-FSCIL \cite{yang2023neural} &    ICLR'23 &  0.511 &  0.705 &  0.687 &  0.691 &  0.659 &  0.658 &  0.673 &  0.659 & 328.16 &  17.55 &   0.68 &   0.67 &   0.95 &   0.77 &   0.68 &  49.92 &  2.642 &  1.237 &  2.456 &  2.240 &  0.985 &  2.418 &  2.100 &  2.011 \\
\rowcolor{yellow!10} SLCA \cite{zhang2023slca}      &    ICCV'23 &  0.508 &  0.731 &  0.686 & \bf 0.729 &  0.684 &  0.686 &  0.666 &  0.675 & 391.98 &  17.48 &   0.67 & \bf 0.58 &   0.95 &   0.68 &   0.75 &  59.01 &  3.156 &  1.232 &  2.445 & \bf 1.940 &  0.990 &  2.146 &  2.340 &  2.036 \\
\rowcolor{yellow!10} Fs-Aug \cite{li2024continual}  &   TCSVT'24 &  0.541 & \bf 0.751 &  0.675 &  0.692 &  0.681 &  0.682 &  0.691 &  0.677 & 316.41 & \bf 15.71 &   0.72 &   0.68 &   0.92 &   0.71 &   0.72 &  47.98 &  2.547 & \bf 1.108 &  2.618 &  2.270 &  0.954 &  2.242 &  2.255 &  1.999 \\
\rowcolor{yellow!10} MAGR \cite{zhou2024magr}       &    ECCV'24 &  0.539 &  0.729 &  0.670 &  0.689 &  0.666 &  0.623 &  0.669 &  0.658 & 278.10 &  16.23 &   0.72 &   0.73 &   0.95 &   0.90 &   0.71 &  42.62 &  2.239 &  1.144 &  2.600 &  2.453 &  0.982 &  2.844 &  2.206 &  2.067 \\
\rowcolor{yellow!10} ASAL \cite{zhou2025adaptive}   &    TVCG'25 &  0.595 &  0.695 &  0.681 &  0.688 &  0.652 &  0.667 &  0.676 &  0.666 & 248.96 &  18.54 &   0.61 &   0.68 &   0.99 &   0.73 &   0.75 &  38.75 &  2.004 &  1.307 &  2.228 &  2.291 &  1.026 &  2.280 &  2.322 &  1.923 \\
\rowcolor{yellow!10} BriMA (Ours) & -- 
& \bf 0.629 & 0.702 & \bf 0.730 & 0.723 & \bf 0.707 & \bf 0.742 & \bf 0.724 & \bf 0.698
& \bf 225.86 & 17.59 & \bf 0.60 & 0.63 & \bf 0.89 & \bf 0.58 & \bf 0.70 & \bf 35.29
& \bf 1.818 & 1.240 & \bf 2.181 & 2.107 & \bf 0.972 & \bf 1.834 &  2.250 & \bf 1.823 \\
    \bottomrule
    \end{tabular}
    }
\end{table*}

\myPara{Sub-Component Performance Comparison on FS1000.}
\cref{tab:supp-fs1000} analyzes TES, PCS, SS, TR, PE, CO, and IN under three missing rates $\beta \in \{10\%, 25\%, 50\%\}$. 
At $\beta=10\%$, our method achieves the best average SRCC (0.756), slightly surpassing the strongest baseline DER++ \cite{buzzega2020dark} (0.755) and delivering the lowest RL2 on average (1.441). Component-wise, it leads SRCC on PCS, SS, TR, and IN, and attains the best MSE on PCS, SS, TR, and CO. Although ASAL \cite{zhou2025adaptive} reports a lower average MSE (29.90 vs. 33.35), our method maintains better correlation and stability, suggesting that bridging favors consistent ranking when modalities are mildly degraded. 

When $\beta=25\%$, our method becomes clearly dominant across metrics. It yields the highest average SRCC (0.740, a 1.0\% improvement over the best baseline DER++ \cite{buzzega2020dark} at 0.733), the lowest average MSE (31.35, an 11.9\% reduction compared with the best baseline ASAL \cite{zhou2025adaptive} at 35.61), and the lowest average RL2 (1.572, a 3.1\% reduction compared with ASAL \cite{zhou2025adaptive} at 1.623). Per component, it consistently improves correlation on TES, PCS, TR, CO, and IN, while matching or exceeding baselines on SS and PE. These gains indicate that memory-guided bridging and modality-aware replay better constrain drift once missing modalities become frequent. 

At $\beta=50\%$, the gap widens as modalities grow scarce. Our method achieves the best average SRCC (0.698, a 3.4\% improvement over the next best SLCA \cite{zhang2023slca} at 0.675), the lowest average MSE (35.29, an 8.9\% reduction compared with ASAL \cite{zhou2025adaptive} at 38.75), and the lowest average RL2 (1.823, a 5.2\% reduction compared with ASAL at 1.923). Component-level results follow the same trend, with consistent improvements on SS, TR, PE, CO, and competitive TES, PCS, and IN. 

Overall, the component analysis reveals three main effects: (1) rank stability, where our method sustains higher SRCC across sub-scores even when low-$\beta$ MSE trade-offs occur; (2) error containment, where improvements in MSE and RL2 accumulate as $\beta$ increases, reflecting stronger robustness to missing-modality noise; and (3) uniformity across sub-scores, where performance gains are distributed rather than concentrated, showing that the bridging and replay strategies generalize effectively across TES, PCS, and fine-grained components such as SS, TR, PE, CO, and IN.

\begin{figure*}
    \centering
    \scriptsize\sf
    \setlength{\tabcolsep}{1pt}
    \newcolumntype{C}[1]{>{\centering\arraybackslash}m{#1}}
    \begin{tabular}{C{3mm} C{0.18\linewidth} C{0.18\linewidth} C{0.18\linewidth} C{0.18\linewidth} C{0.24\linewidth}}
    \rowcolor{orange!10}     &  First Frame & Middle Frame & Middle Frame & Last Frame & Outputs  \\
    \rowcolor{orange!10}  \rotatebox{90}{Ball \#027}   
    & \includegraphics[width=\linewidth,trim=100 0 100 0,clip]{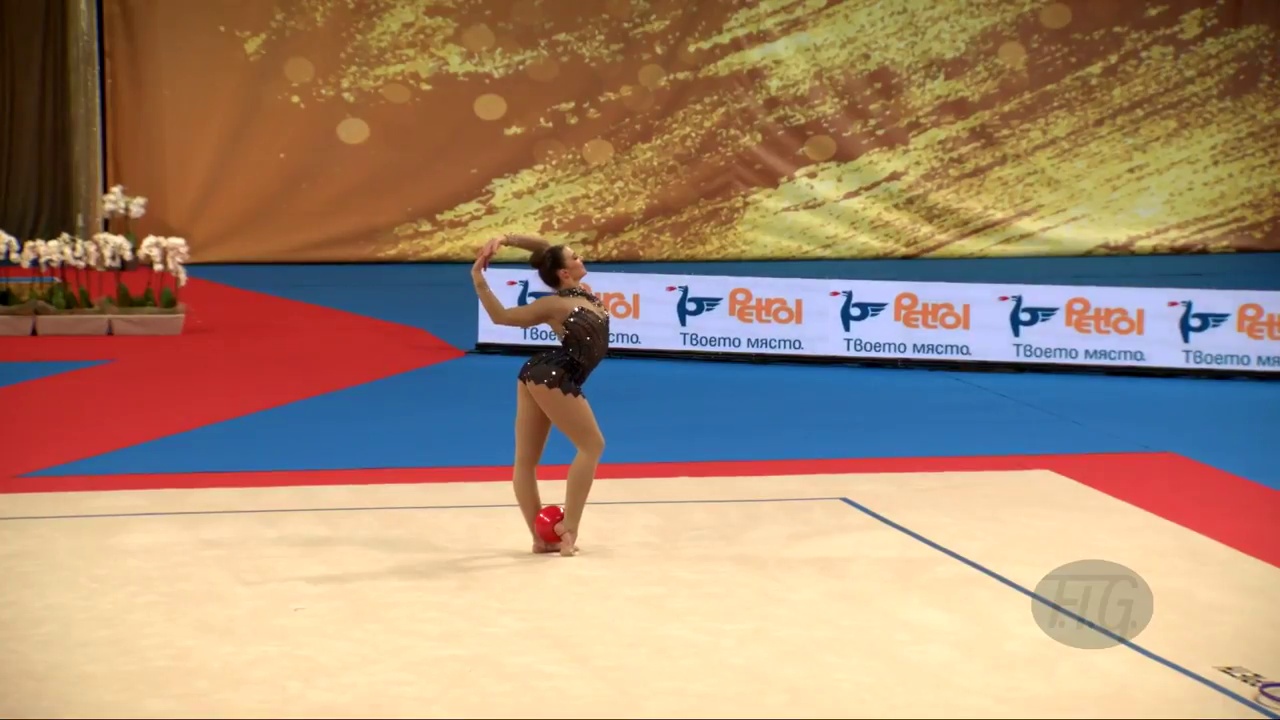} 
    & \includegraphics[width=\linewidth,trim=100 0 100 0,clip]{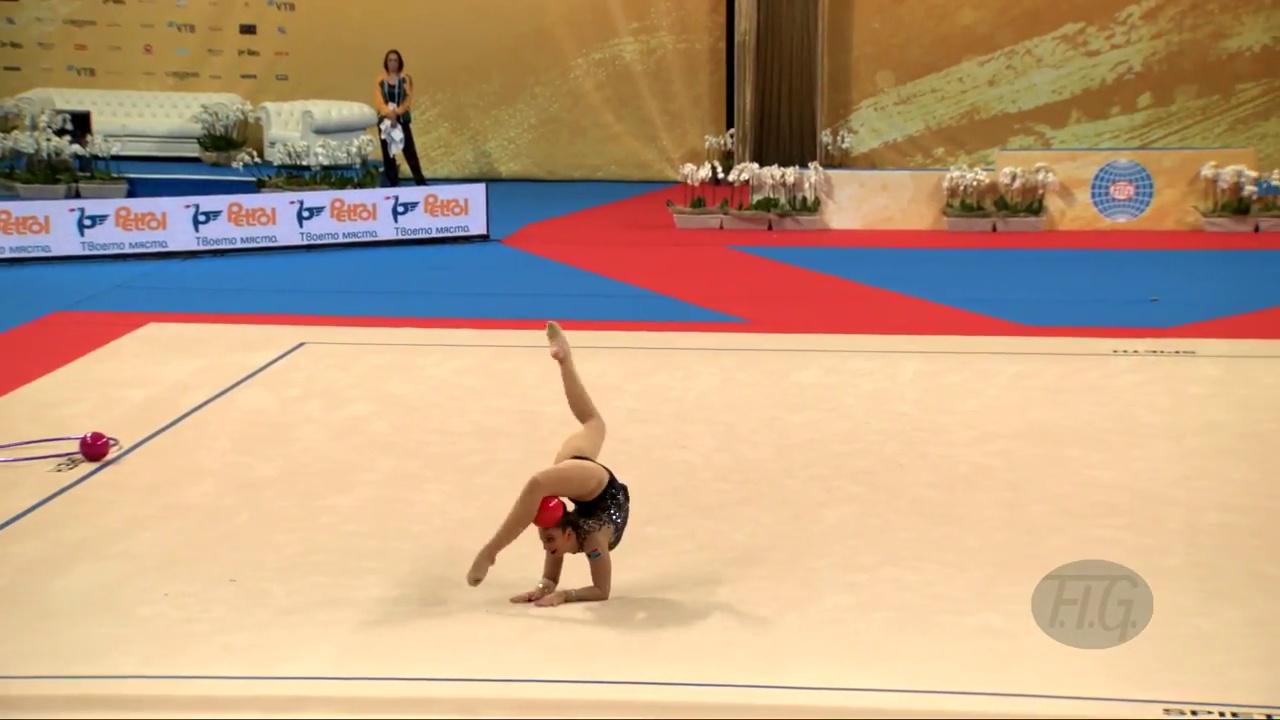} 
    & \includegraphics[width=\linewidth,trim=100 0 100 0,clip]{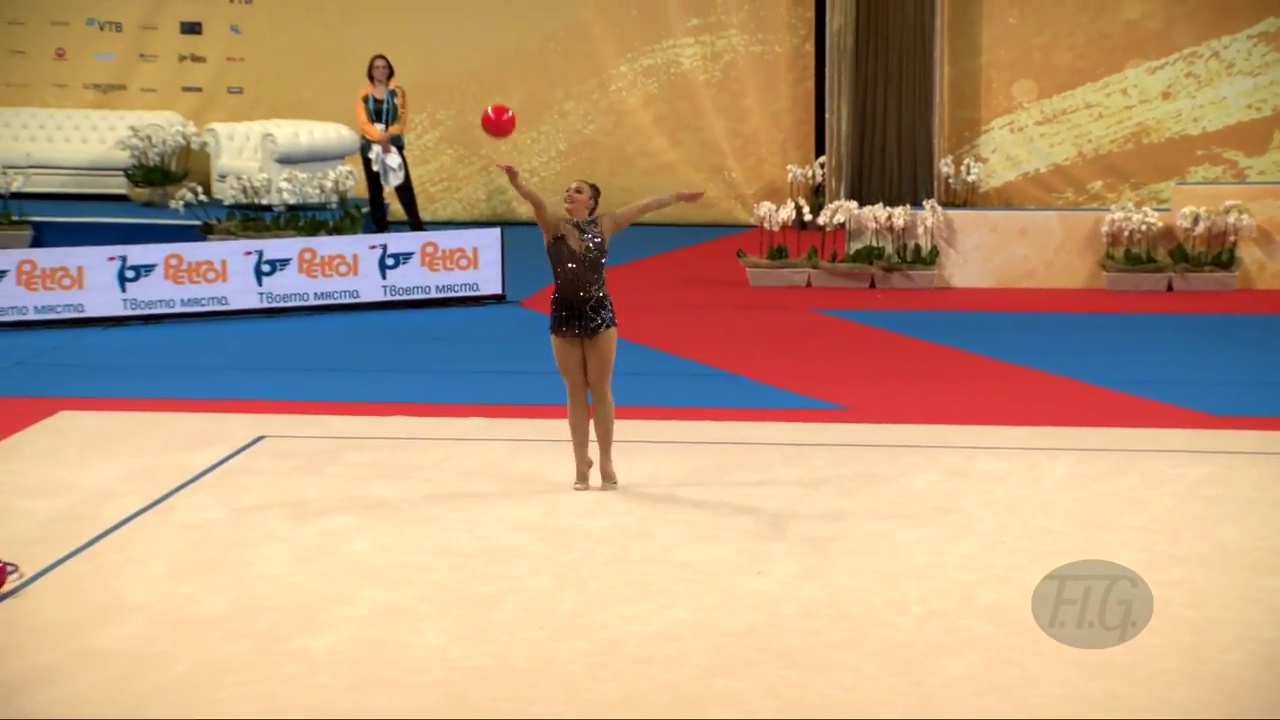} 
    & \includegraphics[width=\linewidth,trim=100 0 100 0,clip]{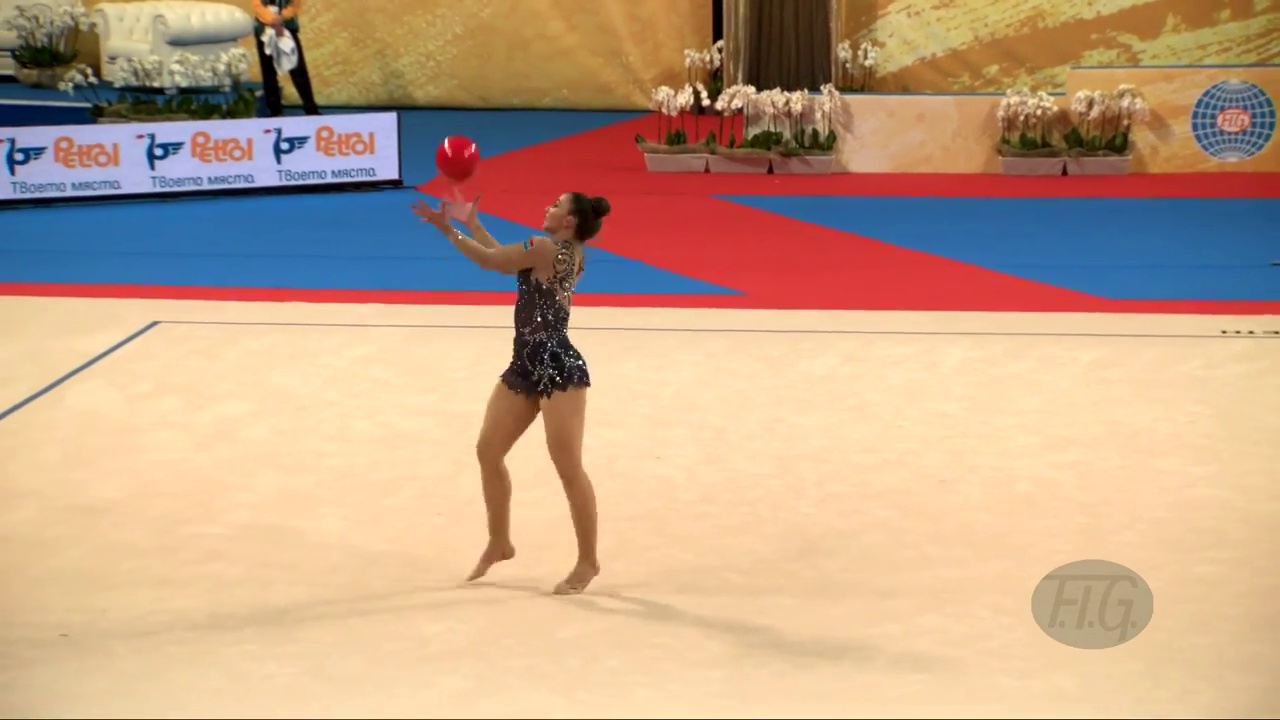} 
    & \includegraphics[width=\linewidth]{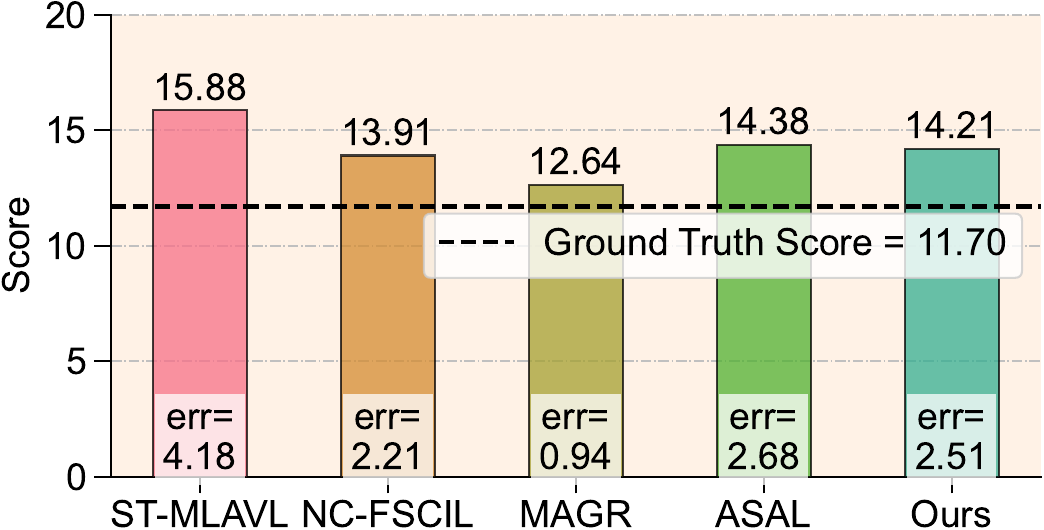} \\
    \rowcolor{orange!10}  \rotatebox{90}{Ball \#030}   
    & \includegraphics[width=\linewidth,trim=100 0 100 0,clip]{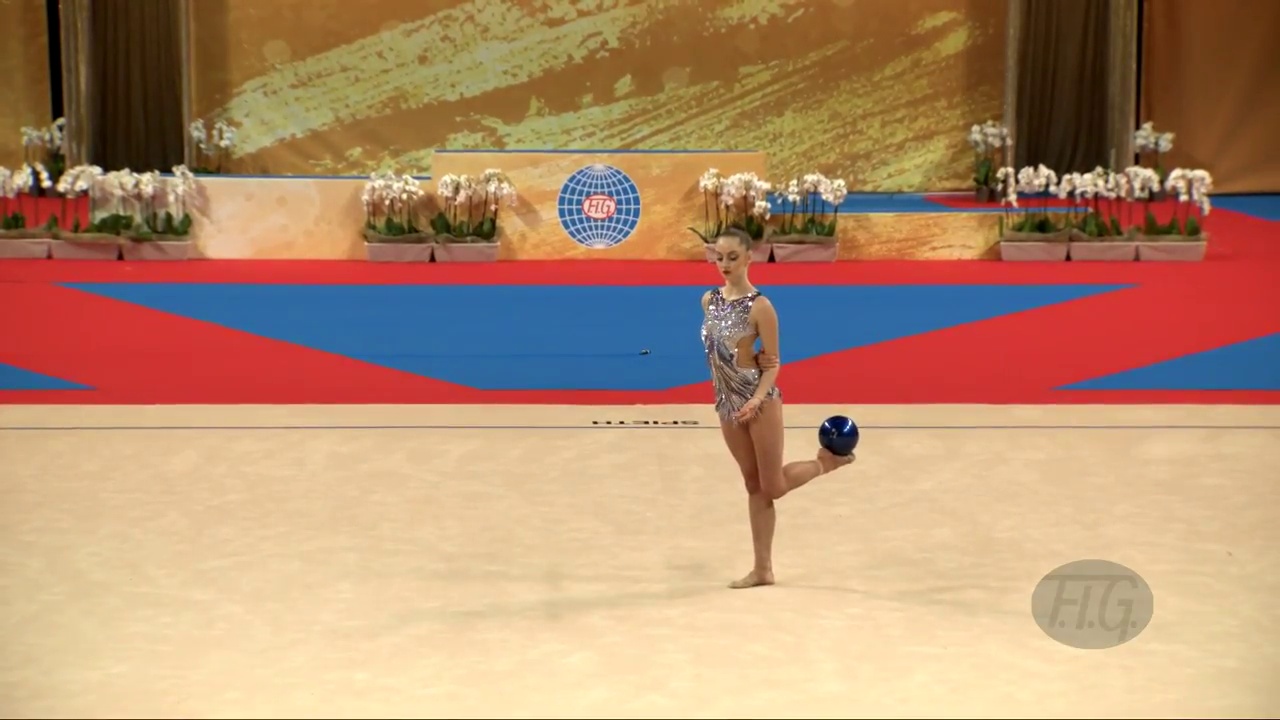} 
    & \includegraphics[width=\linewidth,trim=100 0 100 0,clip]{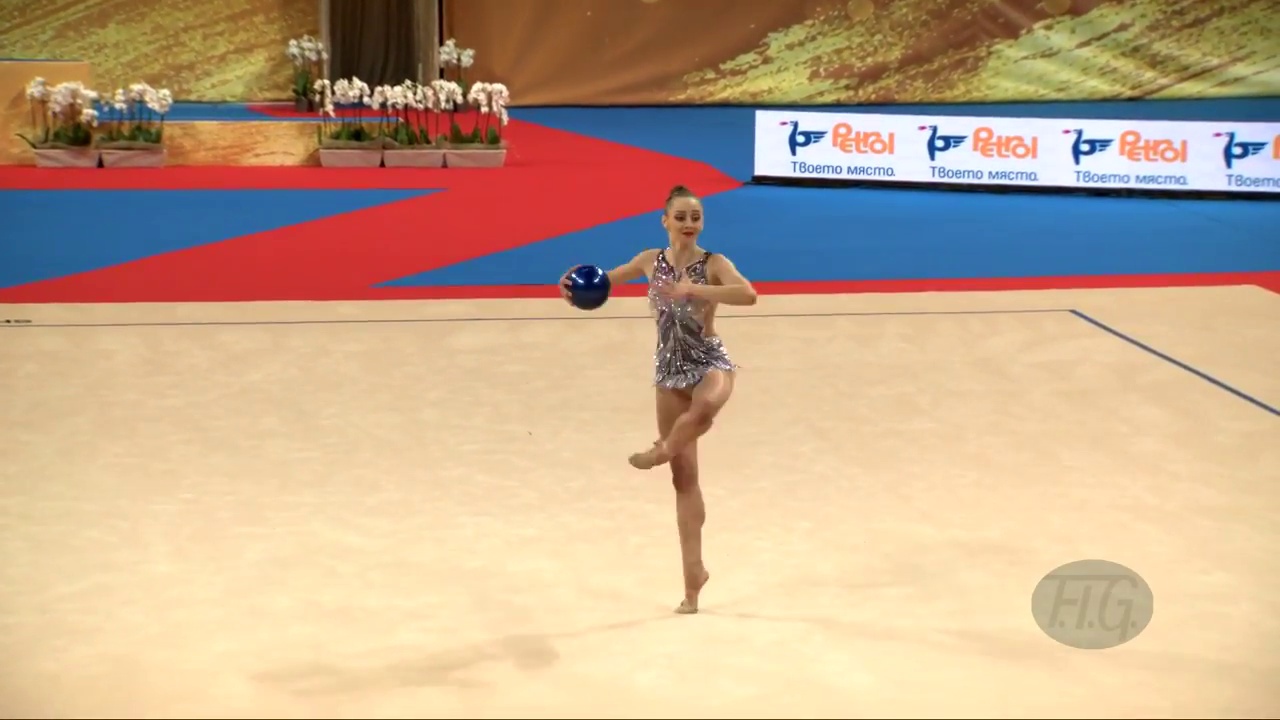} 
    & \includegraphics[width=\linewidth,trim=100 0 100 0,clip]{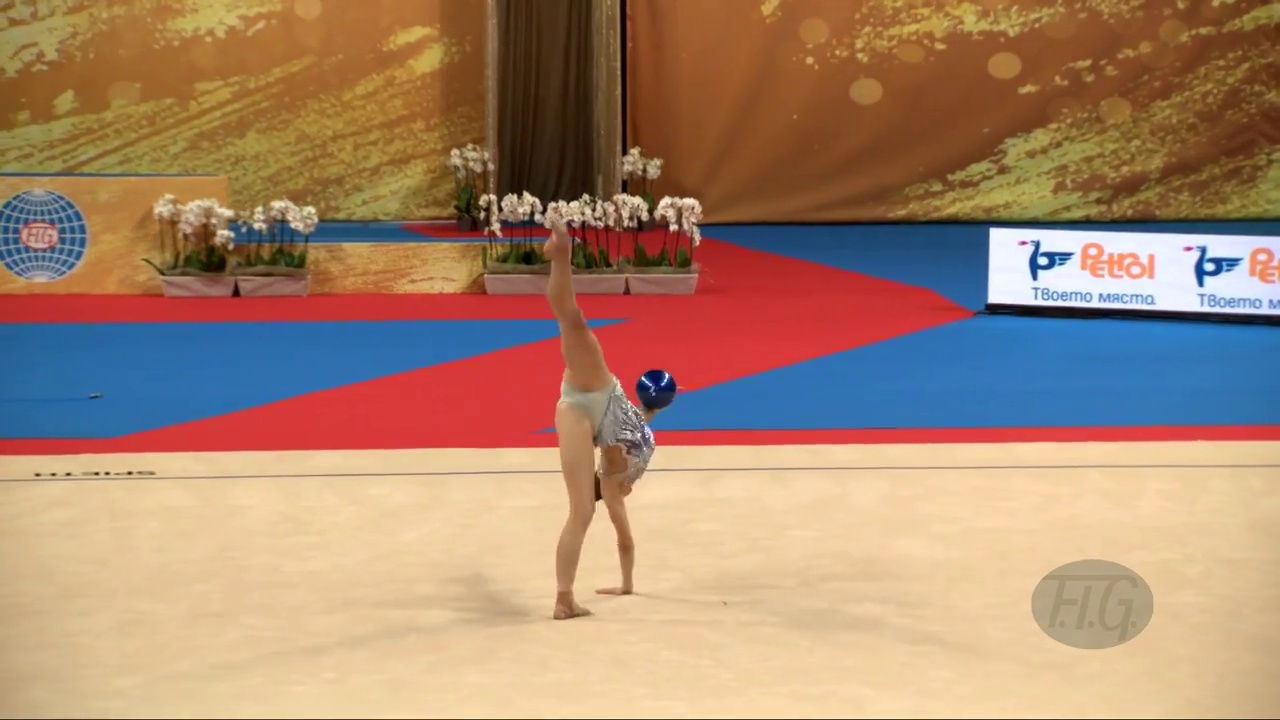} 
    & \includegraphics[width=\linewidth,trim=100 0 100 0,clip]{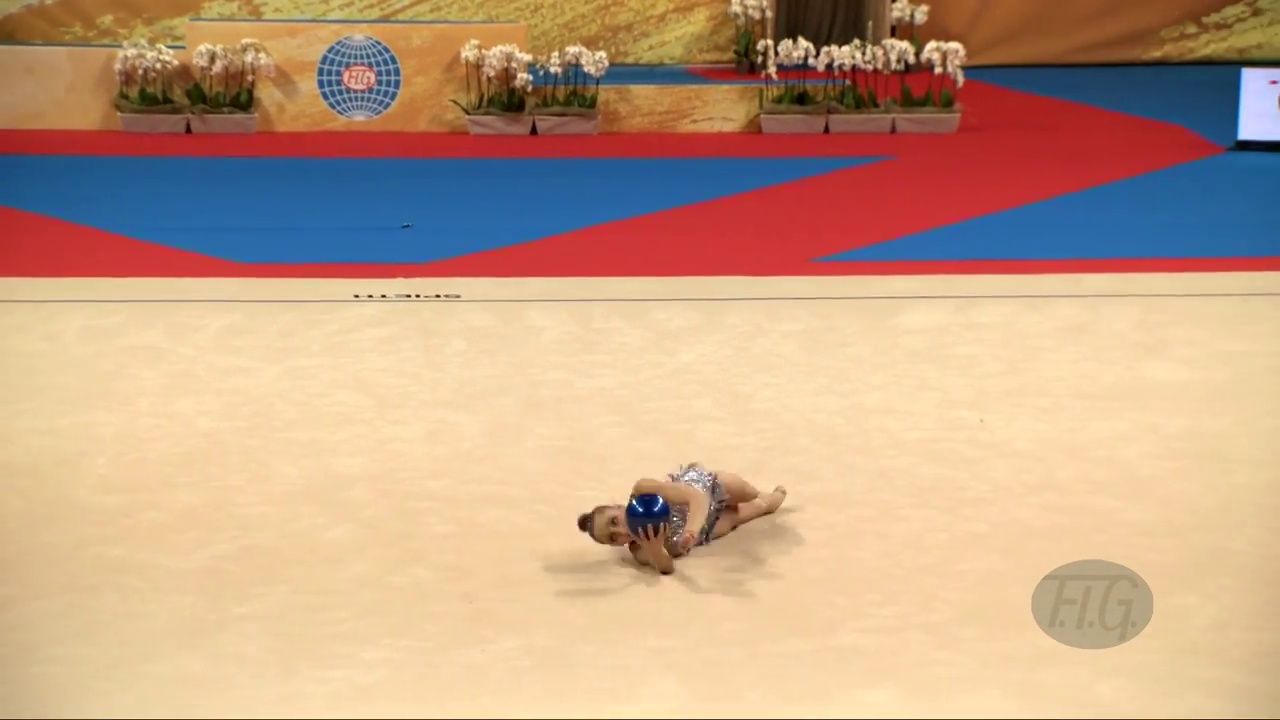} 
    & \includegraphics[width=\linewidth]{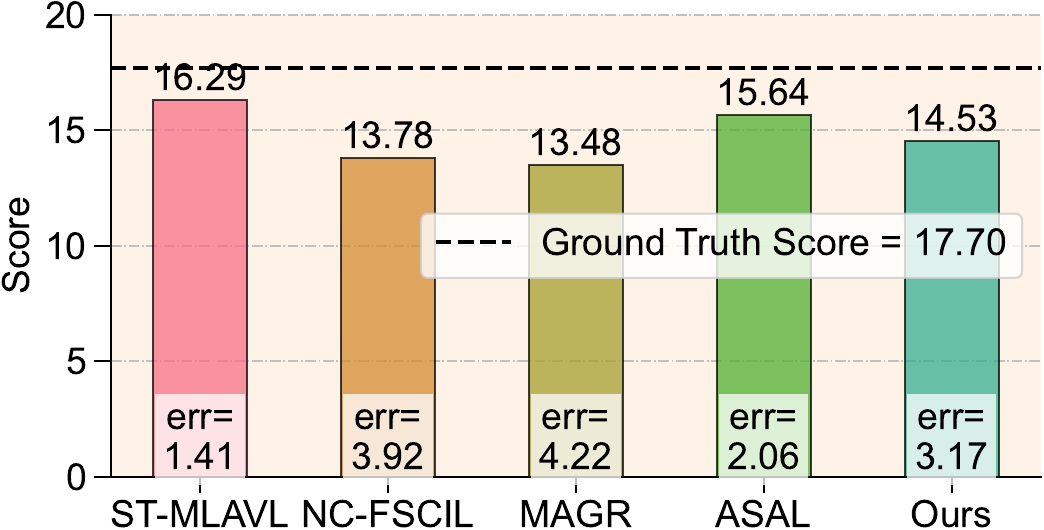} \\
    \rowcolor{orange!10}  \rotatebox{90}{Clubs \#016}   
    & \includegraphics[width=\linewidth,trim=100 0 100 0,clip]{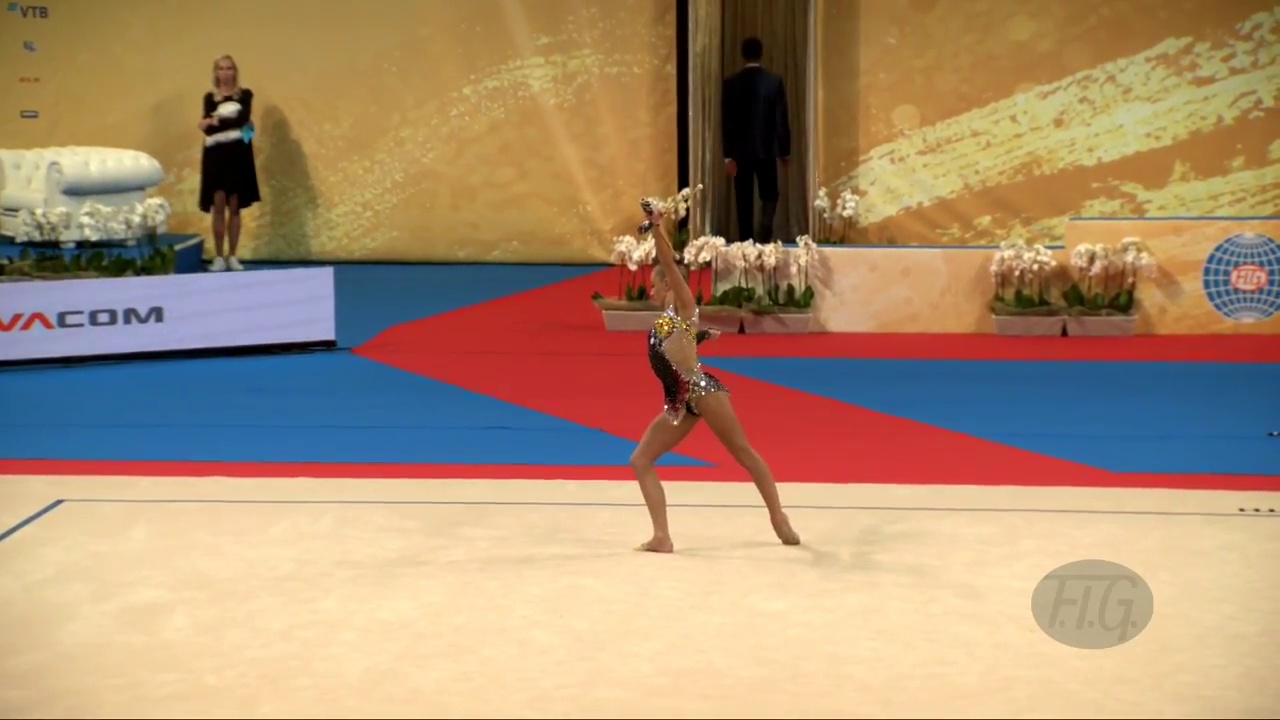} 
    & \includegraphics[width=\linewidth,trim=100 0 100 0,clip]{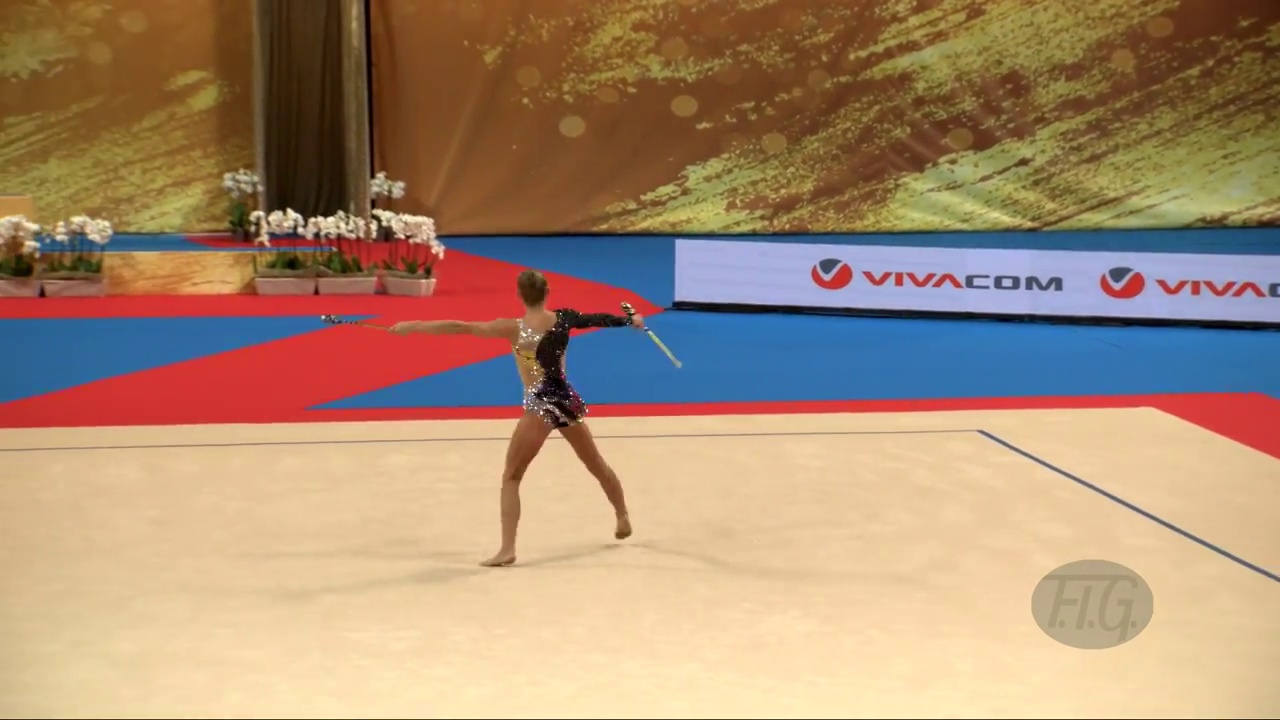} 
    & \includegraphics[width=\linewidth,trim=100 0 100 0,clip]{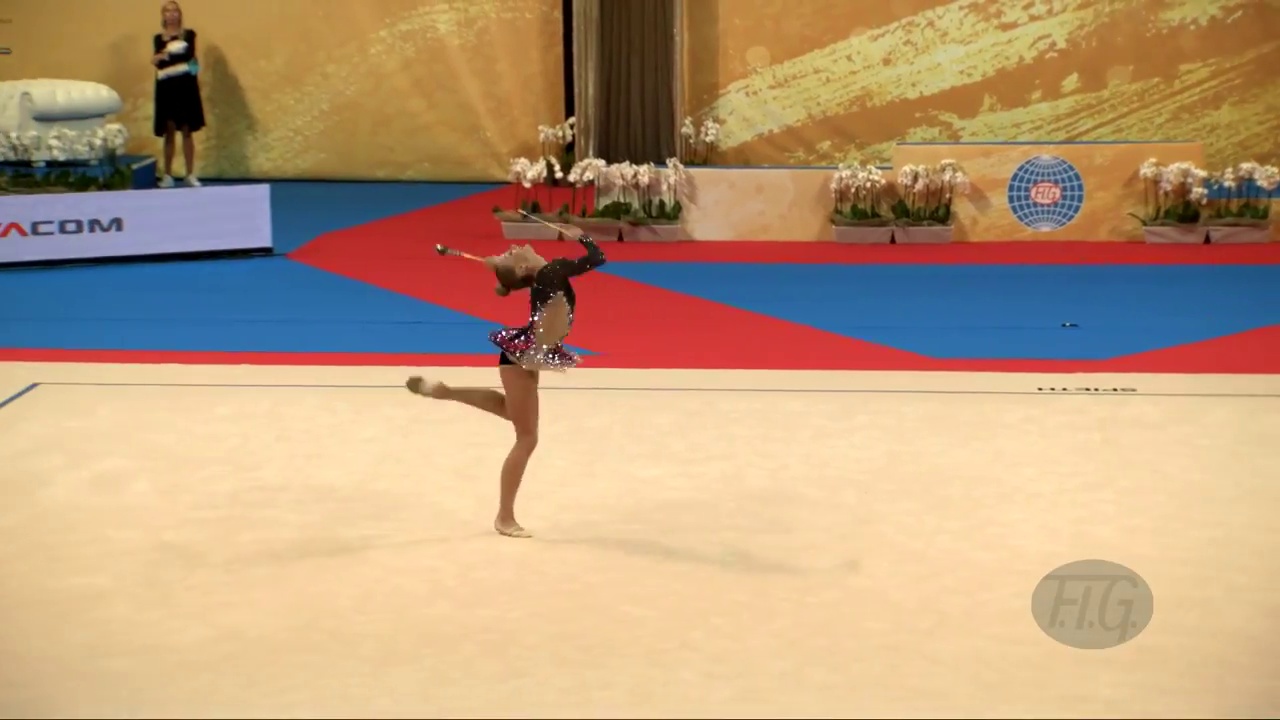} 
    & \includegraphics[width=\linewidth,trim=100 0 100 0,clip]{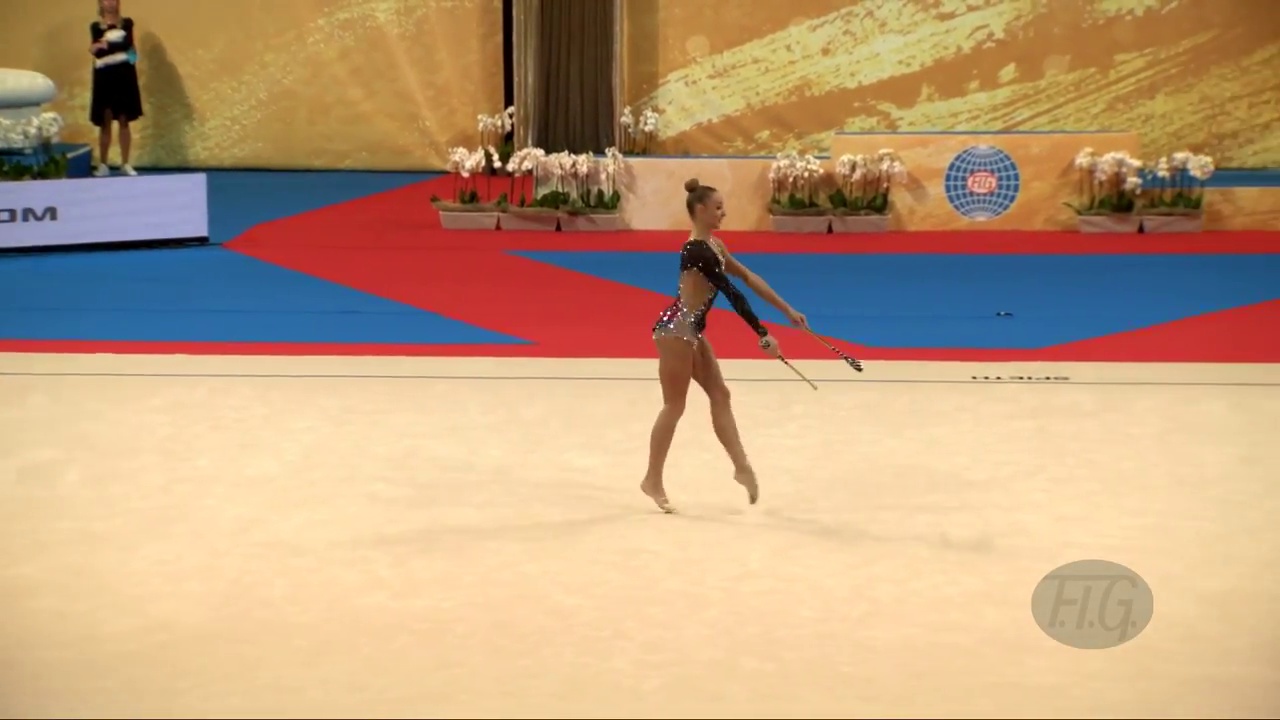} 
    & \includegraphics[width=\linewidth]{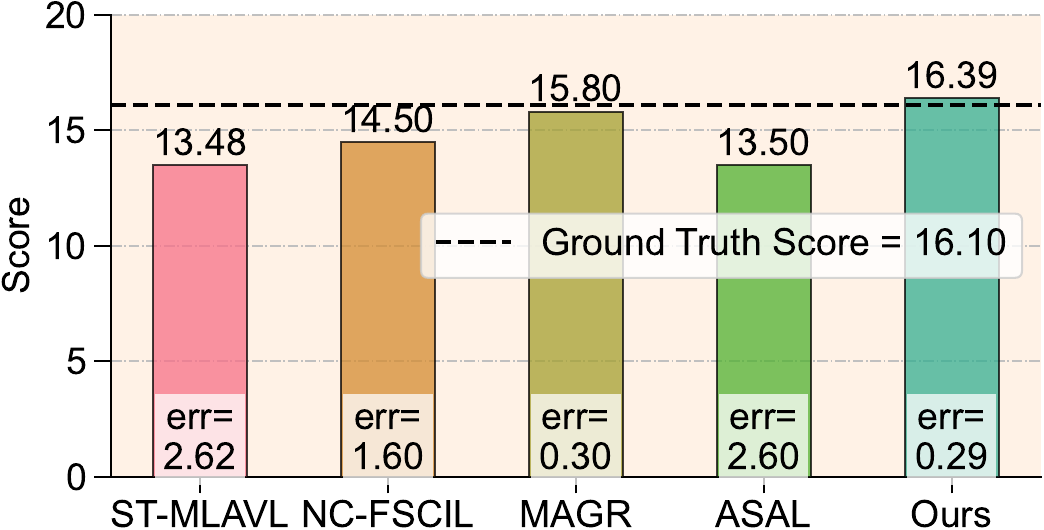} \\
    \rowcolor{magenta!10}  \rotatebox{90}{Ball \#184}   
    & \includegraphics[width=\linewidth,trim=100 0 100 0,clip]{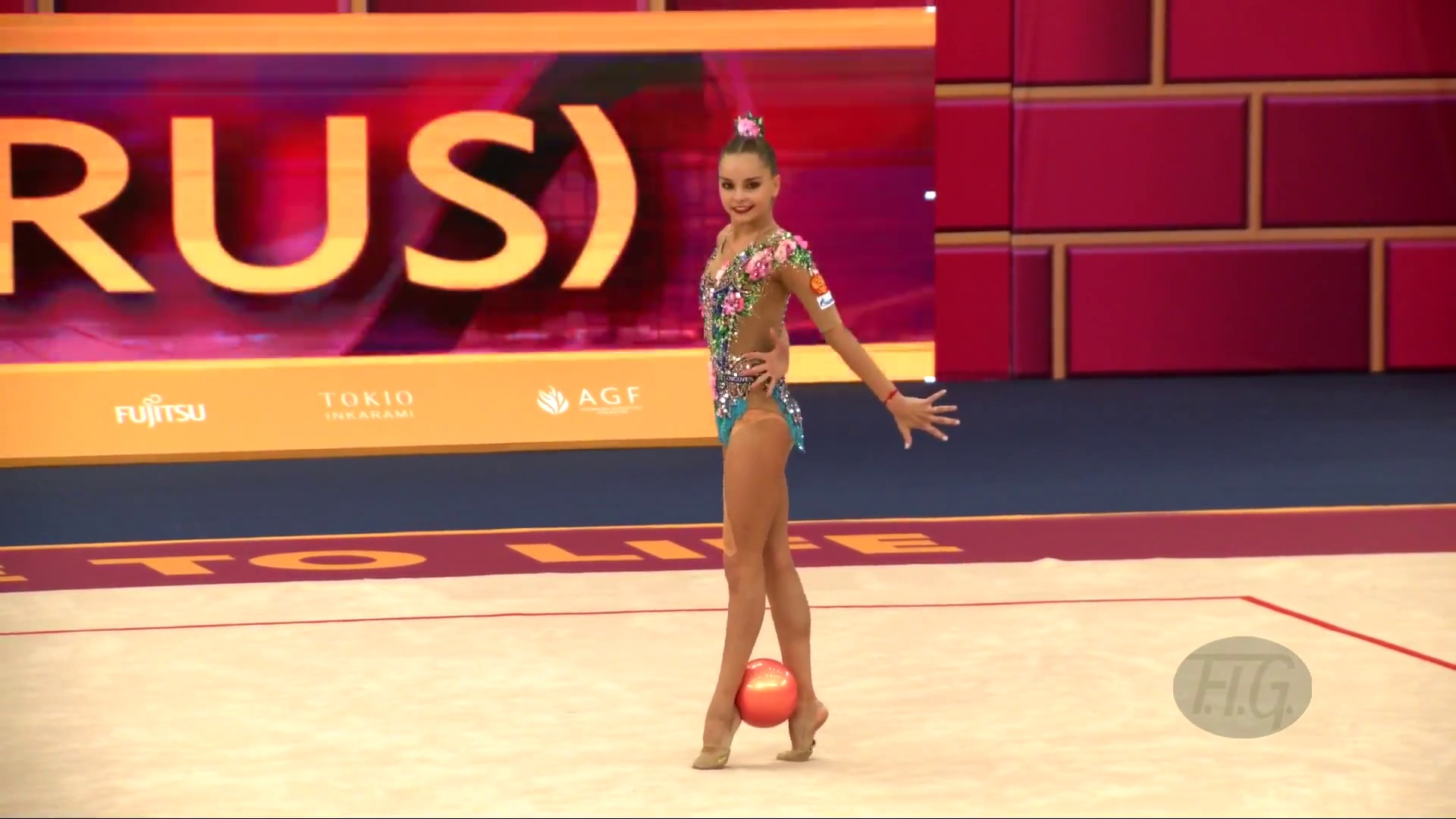} 
    & \includegraphics[width=\linewidth,trim=100 0 100 0,clip]{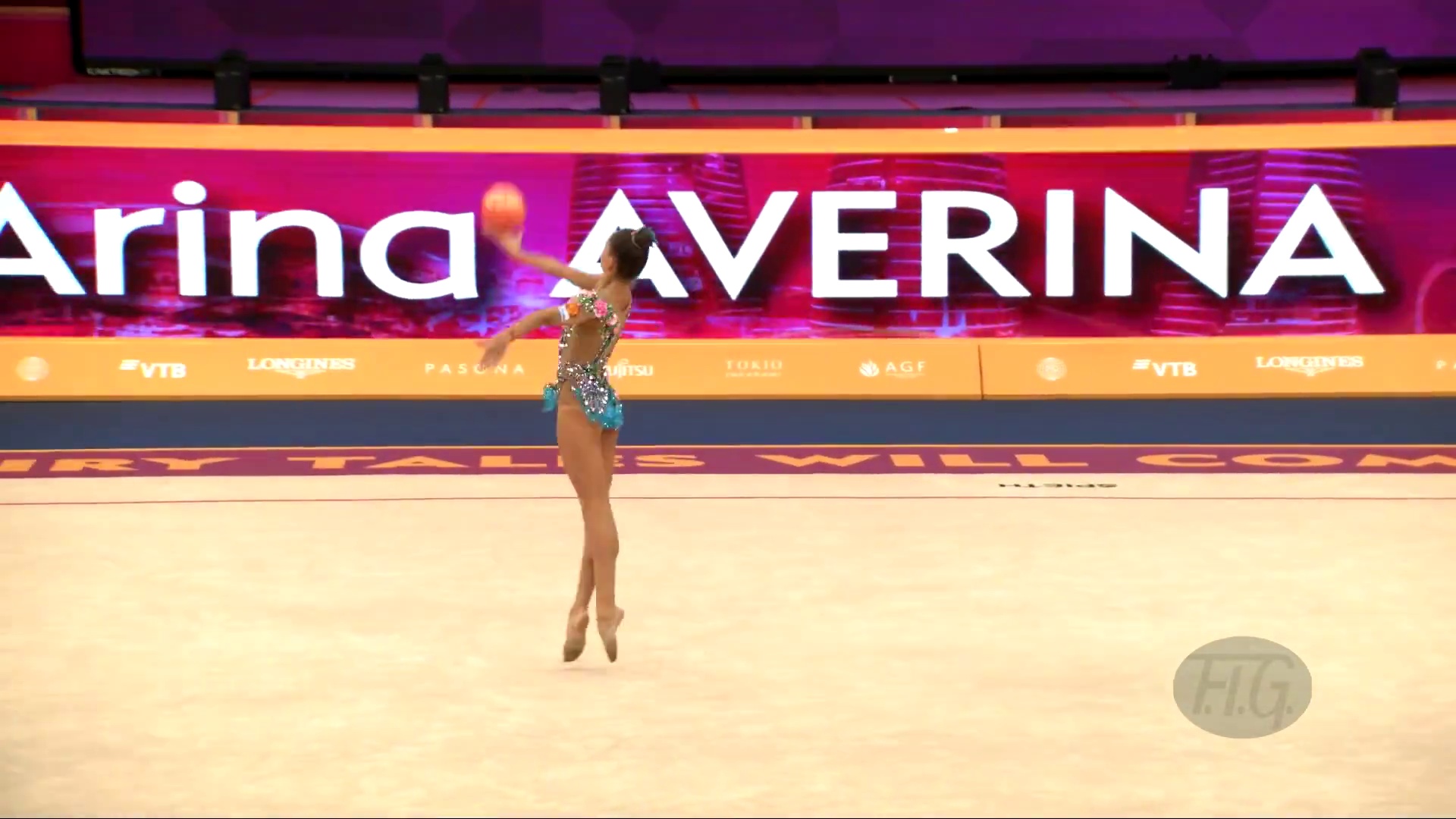} 
    & \includegraphics[width=\linewidth,trim=100 0 100 0,clip]{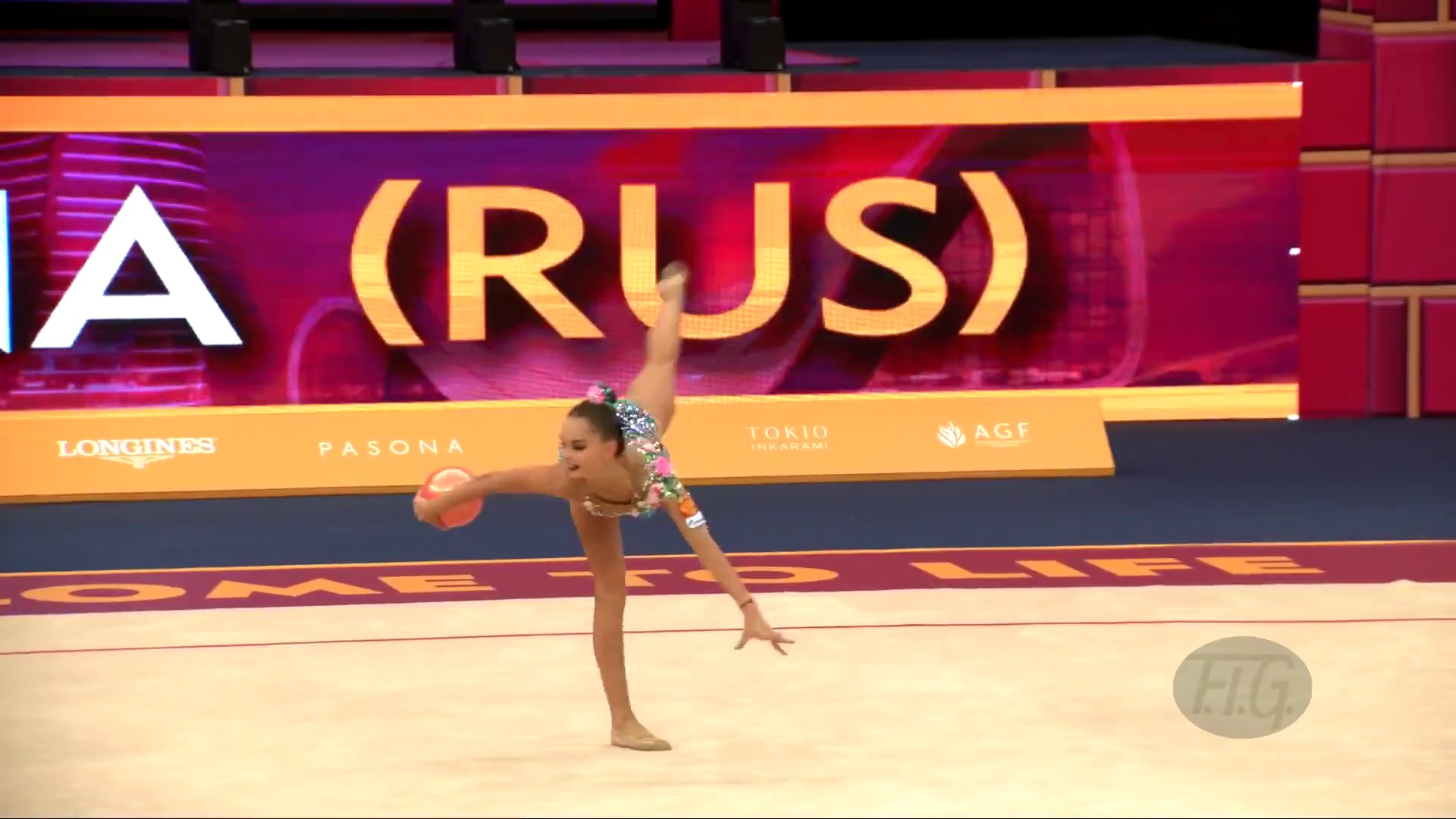} 
    & \includegraphics[width=\linewidth,trim=100 0 100 0,clip]{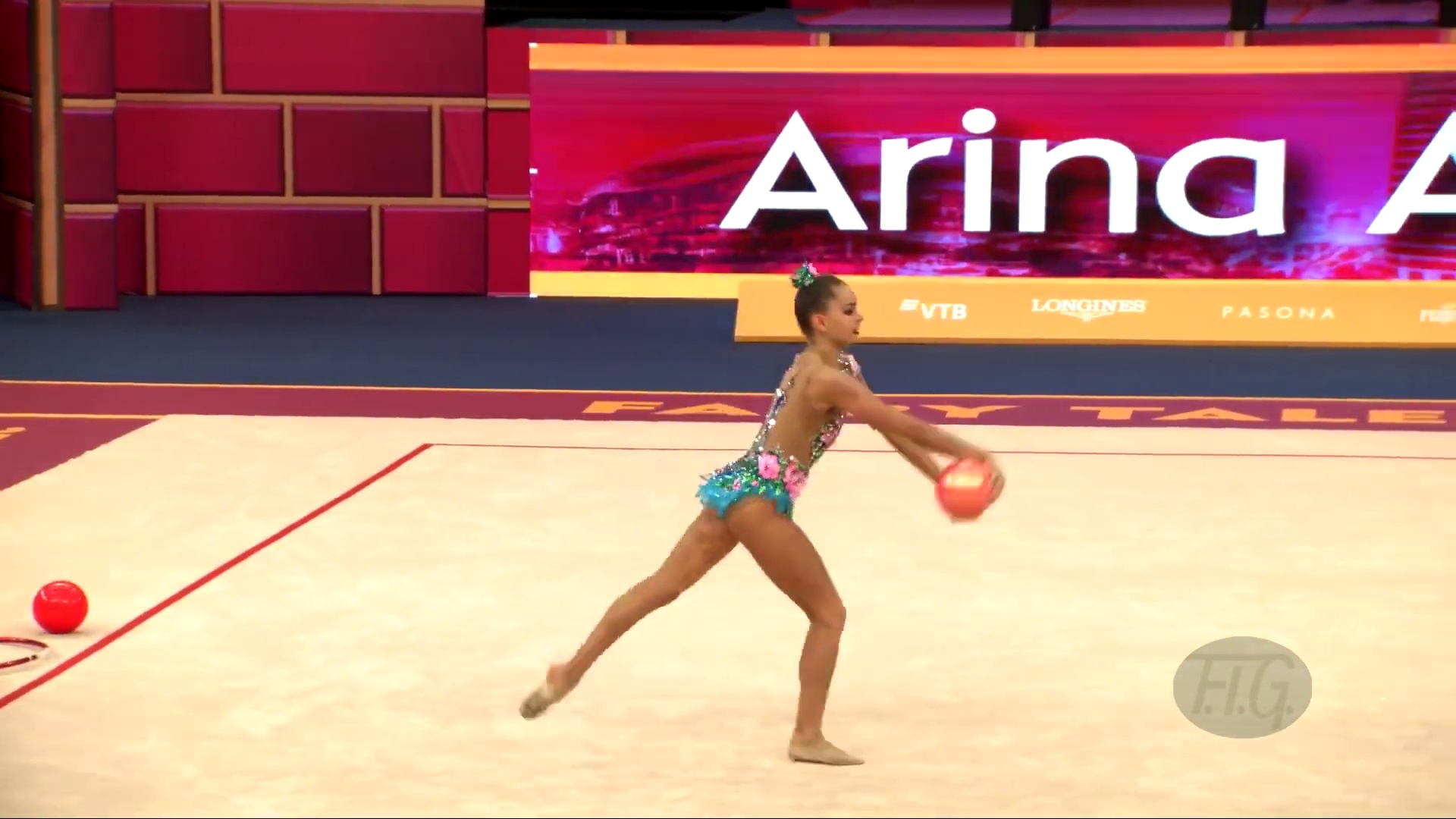} 
    & \includegraphics[width=\linewidth]{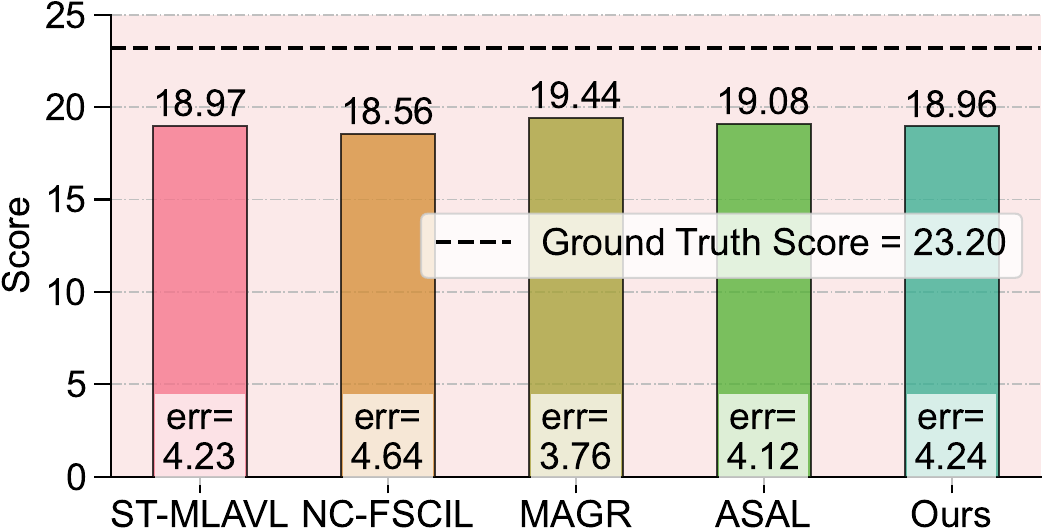} \\
    \rowcolor{magenta!10}  \rotatebox{90}{Clubs \#115}   
    & \includegraphics[width=\linewidth,trim=100 0 100 0,clip]{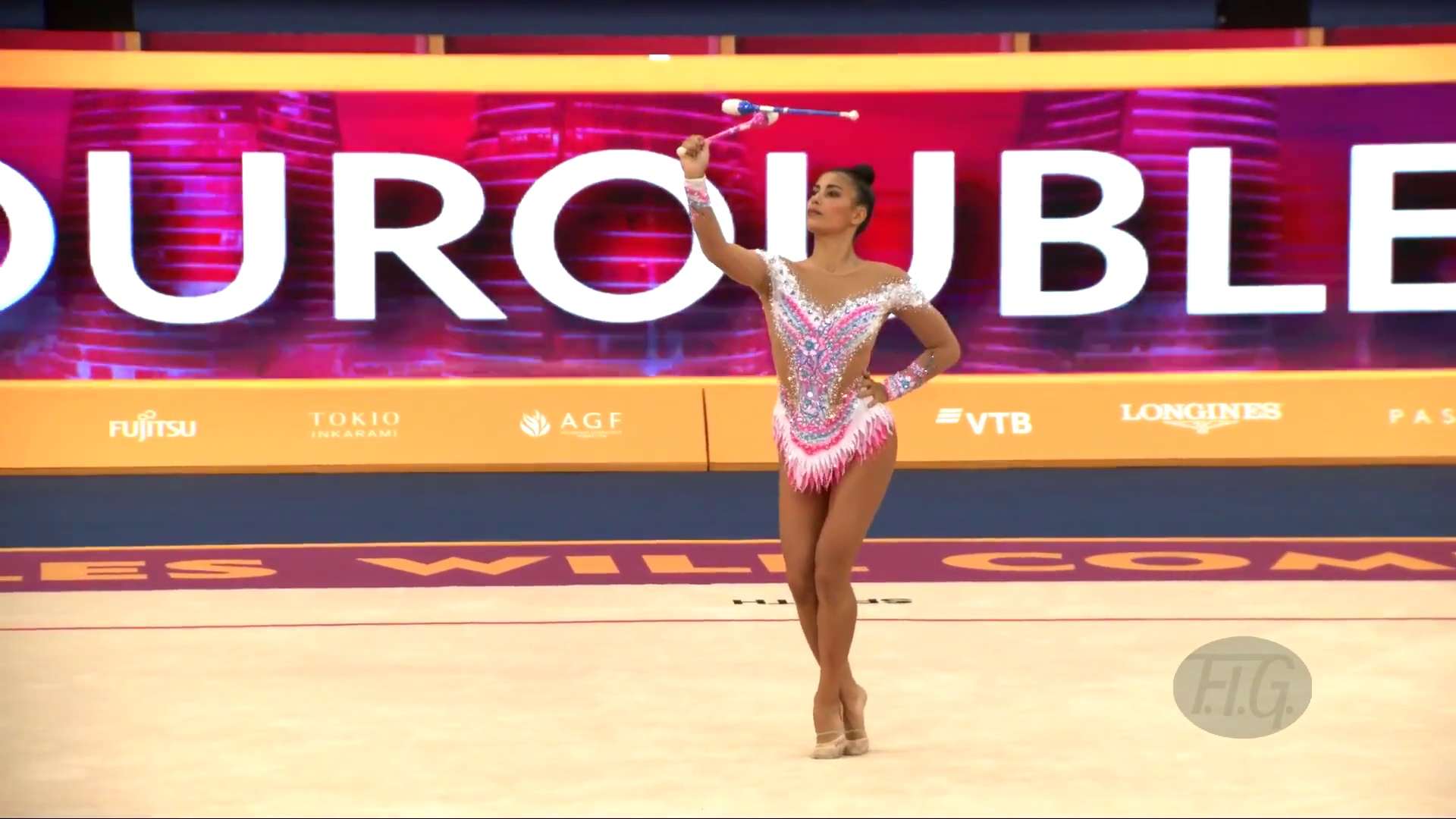} 
    & \includegraphics[width=\linewidth,trim=100 0 100 0,clip]{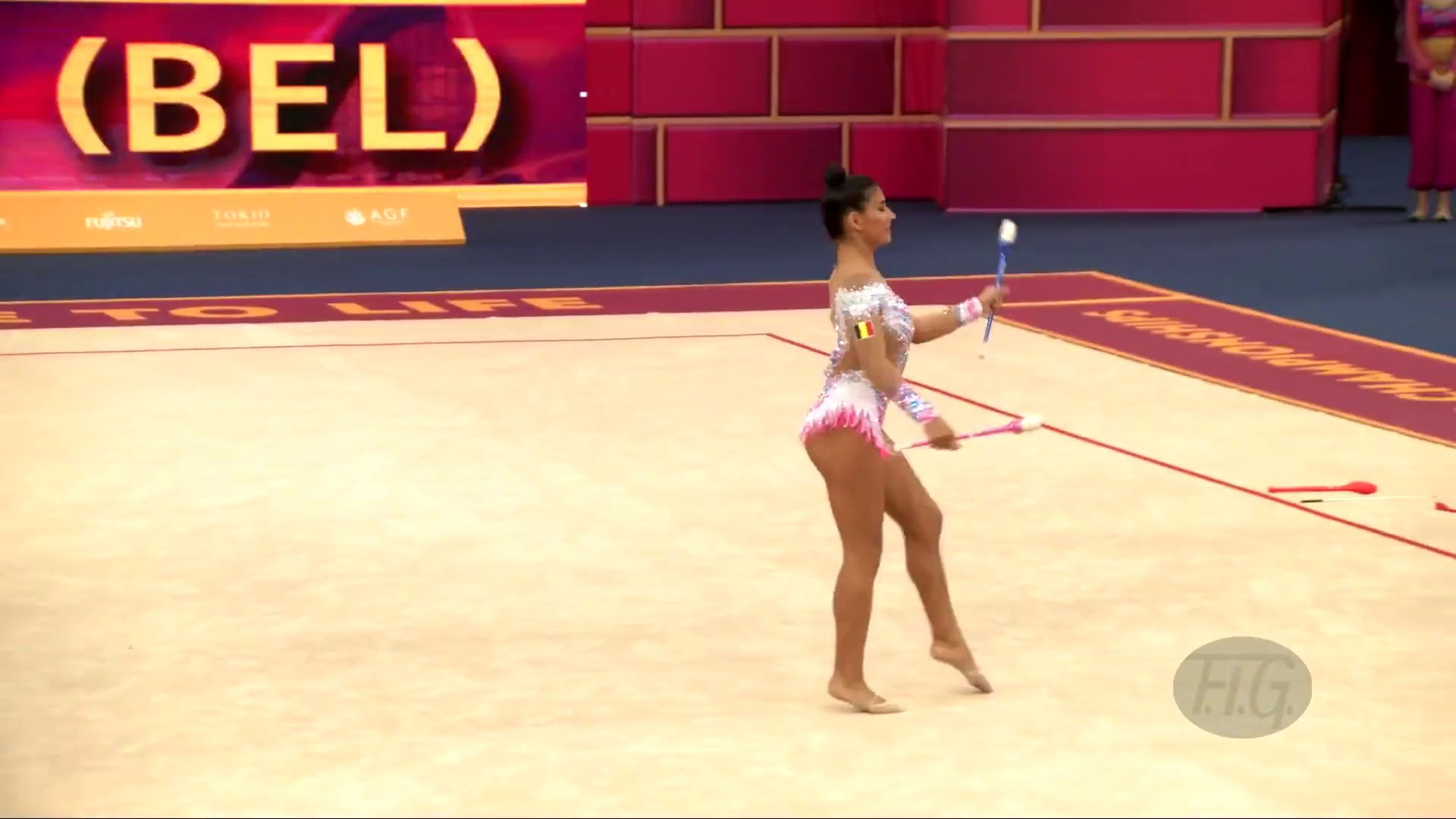} 
    & \includegraphics[width=\linewidth,trim=100 0 100 0,clip]{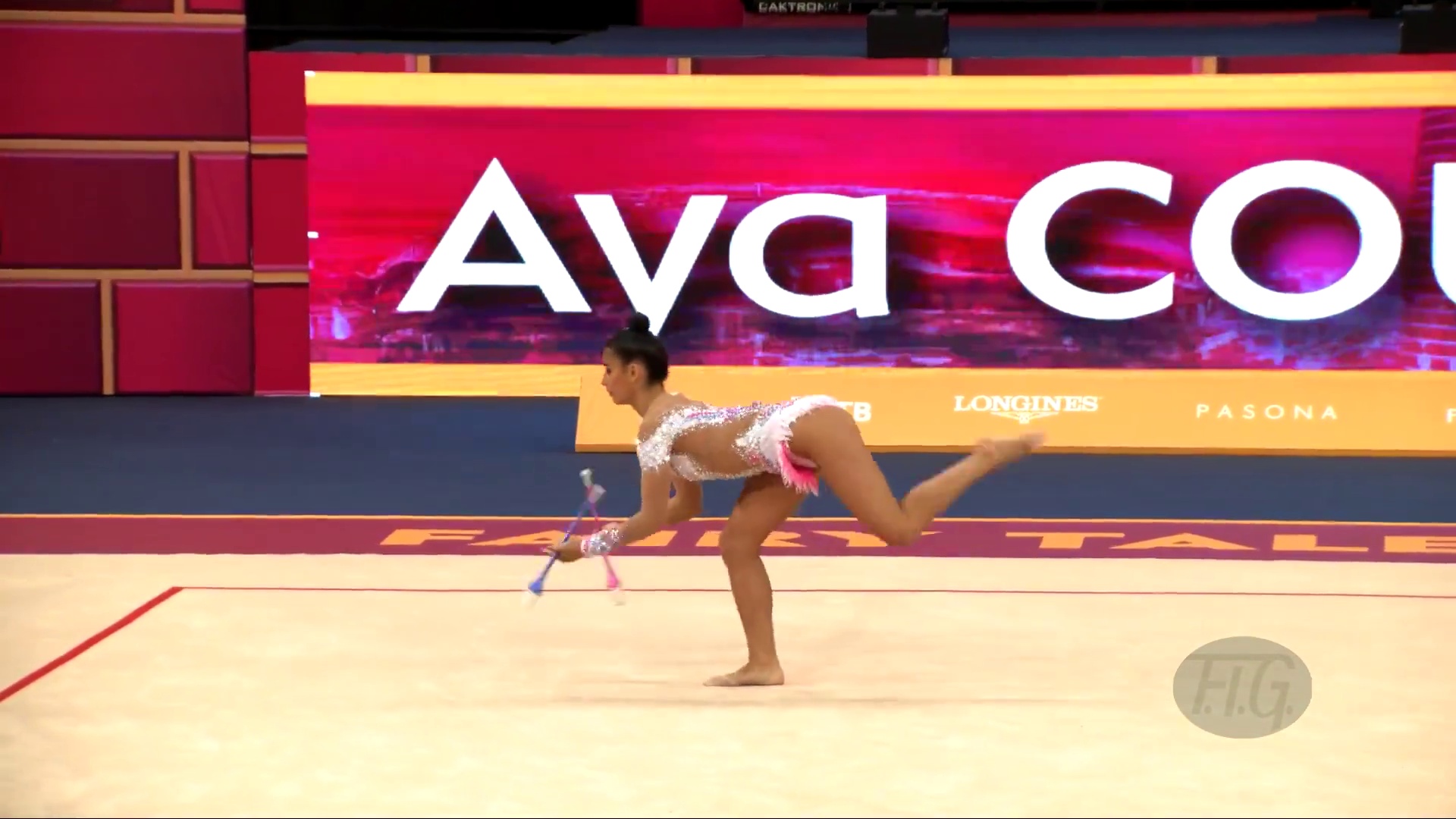} 
    & \includegraphics[width=\linewidth,trim=100 0 100 0,clip]{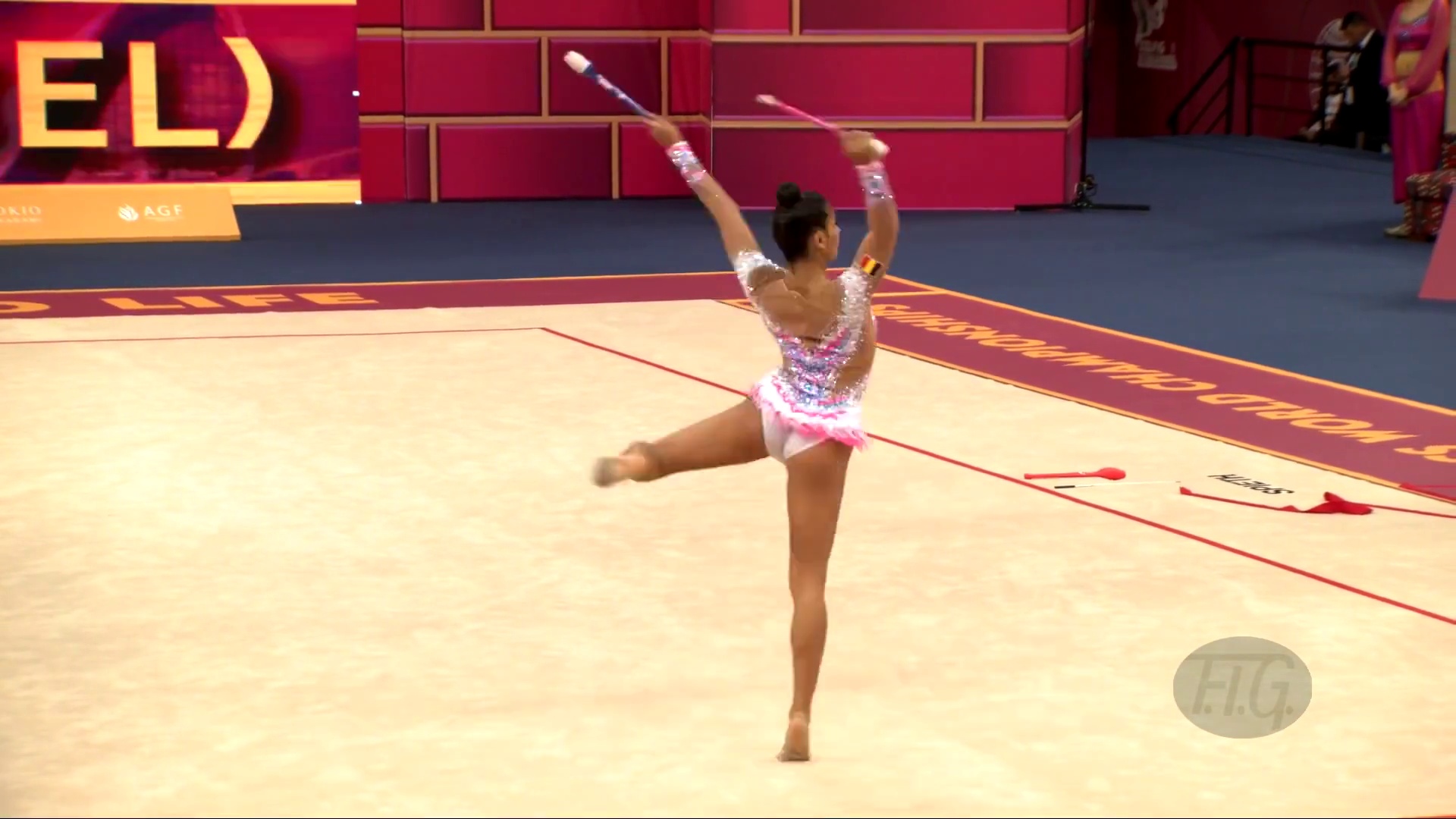} 
    & \includegraphics[width=\linewidth]{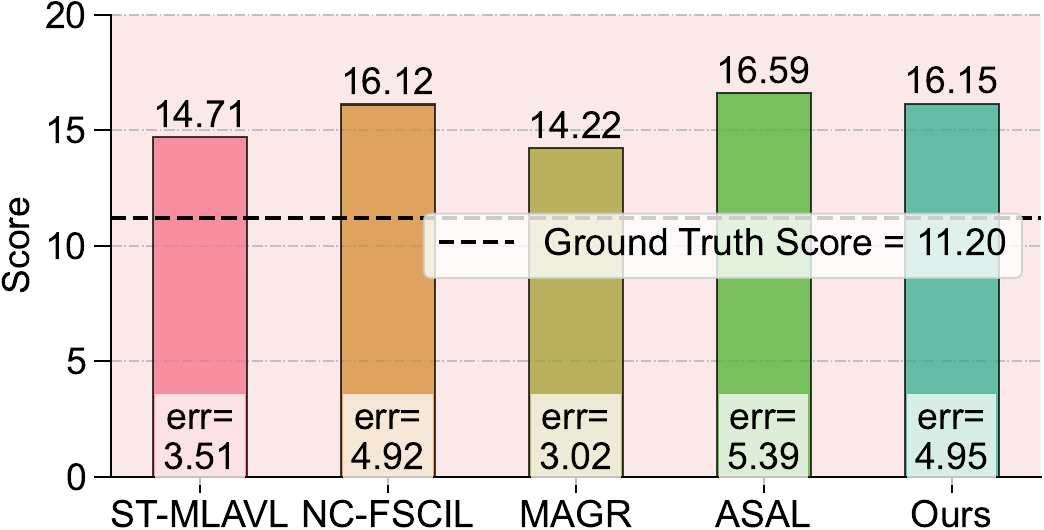} \\
    \end{tabular}
    \caption{Case study (supplementary results of \cref{fig:case_study}): (a), (b), and (c) are easy cases, and (d) and (e) are hard cases.}
    \label{fig:supp-case_study}
    \phantomsubcaption\label{fig:supp-case_study-a}
    \phantomsubcaption\label{fig:supp-case_study-b}
    \phantomsubcaption\label{fig:supp-case_study-c}
    \phantomsubcaption\label{fig:supp-case_study-d}
    \phantomsubcaption\label{fig:supp-case_study-e}
\end{figure*}

\myPara{Additional Case Study.}  
As shown in Fig.~S2, we analyze five representative examples covering both easy and hard cases to further examine the robustness of our method under diverse visual and contextual conditions. Each row displays sampled frames from a video along with the predicted scores of different methods compared with the ground truth. Our method consistently produces the most accurate predictions, validating its stability against varying scene complexity and modality degradation.  

For the easy cases (Ball \#027 \cref{fig:supp-case_study-a}, Ball \#030 \cref{fig:supp-case_study-b}, and Clubs \#016 \cref{fig:supp-case_study-c}), the background and lighting are well-separated from the performer, resulting in clear spatial boundaries and smooth motion cues. Competing methods often exhibit mild overestimation (approximately +1–3 points), primarily due to overfitting to salient motion or contrastive cues. In contrast, our approach aligns closely with ground-truth scores (e.g., 11.77, 11.37, and 12.87), benefiting from temporally coherent cross-modal representations and calibrated replay of rhythmic patterns.  

In the hard cases (Ball \#184 (see \cref{fig:supp-case_study-d}) and Clubs \#115 (\cref{fig:supp-case_study-e})), the background exhibits strong color and texture similarity to the gymnast’s costume, leading to blurred motion boundaries and degraded feature discrimination. As a result, all baseline methods suffer large prediction errors ($-$3 to $-$6 points), failing to accurately localize motion trajectories. Our method effectively mitigates this issue by leveraging the bridging space to restore modality balance and suppress background interference, achieving near-perfect predictions (e.g., 23.30 and 11.20).  

Overall, these case studies highlight that when visual similarity between the performer and background increases, most models struggle to distinguish motion boundaries, yielding less precise score predictions. In contrast, our method maintains reliable estimation through memory-guided retrieval and structural bridging, demonstrating strong resilience to complex visual ambiguity and domain imbalance.

\myPara{Additional Ablation Study.}
Due to space limitations, the core components of BriMA have not been fully explored in the main paper. The additional ablation results in \cref{tab:supp-ablation} provide a more complete understanding of how each module contributes to overall performance. Removing the task-specific embedding (see \cref{eq:bridge}) leads to a clear drop in SRCC across all four action types, indicating that modality-aware conditioning is essential for stabilizing residual reconstruction under shifting observation patterns. Disabling replay prioritization (see \cref{eq:priority}) also reduces performance because the model can no longer focus on samples that exhibit large modality distortion or score drift, resulting in weaker temporal consistency and higher regression error. These results verify that both the bridging design and the replay mechanism are important for maintaining robust scoring in continual multi-modal settings. The ablations further highlight that BriMA’s improvements do not come from a single component but from the cooperation between memory-guided imputation and informed replay.

\begin{table}[]
    \centering
    \small
    \setlength{\tabcolsep}{5pt}
    \caption{Ablation study on the RG dataset ($\beta=10\%$).}
    \label{tab:supp-ablation}
    \resizebox{\linewidth}{!}{
    \begin{tabular}{
    cl
    S[table-format=2.3] S[table-format=2.3] S[table-format=2.3] S[table-format=2.3] S[table-format=2.3]
    S[table-format=2.2] S[table-format=2.2] S[table-format=2.2] S[table-format=2.2] S[table-format=2.2]
    S[table-format=2.3] S[table-format=2.3] S[table-format=2.3] S[table-format=2.3] S[table-format=2.3]
    }
    \toprule
    \multirow{2.5}{*}{\textbf{ID}} & \multirow{2.5}{*}{\textbf{Setting}} &
    \multicolumn{5}{c}{\textbf{SRCC ($\uparrow$)}} \\
    \cmidrule(lr){3-7}
    & & \multicolumn{1}{c}{Ball} & \multicolumn{1}{c}{Clubs} & \multicolumn{1}{c}{Hoop} & \multicolumn{1}{c}{Ribbon} & \multicolumn{1}{c}{Avg.} \\
    \midrule
    \rowcolor{yellow!10}
    1 & Ours &
    0.648 & 0.788 & 0.710 & 0.836 & 0.746 \\

    \rowcolor{orange!10}
    2 & Ours w/o Task-Specific Embedding &
    0.579 & 0.702 & 0.656 & 0.834 & 0.693 \\

    \rowcolor{yellow!10}
    3 & Ours w/o Replay Prioritization &
    0.639 & 0.776 & 0.628 & 0.791 & 0.717 \\

    \midrule
    & & \multicolumn{5}{c}{\textbf{MSE ($\downarrow$)}} \\
    \cmidrule(lr){3-7}

    \rowcolor{yellow!10}
    1 & Ours &
    8.65 & 4.88 & 7.12 & 4.39 & 6.26 \\

    \rowcolor{orange!10}
    2 & Ours w/o Task-Specific Embedding &
    10.165 & 6.722 & 7.510 & 3.709 & 7.027 \\

    \rowcolor{orange!10}
    3 & Ours w/o Replay Prioritization &
    7.17 & 6.29 & 9.20 & 5.08 & 6.93 \\

    \midrule
    & & \multicolumn{5}{c}{\textbf{RL2 ($\downarrow$)}} \\
    \cmidrule(lr){3-7}

    \rowcolor{yellow!10}
    1 & Ours &
    2.310 & 2.291 & 2.681 & 1.641 & 2.231 \\

    \rowcolor{orange!10}
    2 & Ours w/o Task-Specific Embedding &
    2.715 & 3.153 & 2.826 & 1.388 & 2.520 \\

    \rowcolor{yellow!10}
    3 & Ours w/o Replay Prioritization &
    1.914 & 2.950 & 3.461 & 1.900 & 2.556 \\

    \bottomrule
    \end{tabular}
    }
\end{table}

\section{Generalization Beyond AQA}

BriMA is designed for decision-critical regression under non-stationary modality imbalance, where both task distributions and modality availability may evolve over time. While AQA serves as our primary benchmark, we further examine generalization to a distinct multi-modal regression task: sentiment intensity prediction.

\myPara{Task and Setting.}
The CMU multi-modal Opinion Sentiment and Emotion Intensity (MOSI) dataset \cite{zadeh2016mosi} is a foundational, opinion-level annotated corpus for multi-modal sentiment analysis, consisting of 2,199 video clips from 93 YouTube movie reviews. It is a multi-modal sentiment regression benchmark involving visual, acoustic, and textual modalities, where the goal is to predict continuous sentiment scores. Unlike AQA, which evaluates performance quality in human actions, MOSI focuses on affective understanding in conversational settings. This provides a semantically different regression task while retaining multi-modal structure and score-sensitive evaluation for generalization evaluation.
To simulate non-stationary modality imbalance, we follow the same protocol as in the above AQA setting and randomly drop modalities at rates $\beta \in \{10\%, 25\%, 50\%\}$ during CL phases. We compare BriMA with joint training (JT-EMOE \cite{fang2025emoe}), standard CL baselines, and recent CAQA methods under identical settings.

\myPara{Results.}
As shown in \cref{tab:mosi_results}, BriMA consistently achieves the best performance across SRCC, MSE, and RL2 under all missing rates. Notably, BriMA even outperforms joint training (JT-EMOE) when modalities are partially missing, indicating its ability to maintain score fidelity under incomplete observations. Compared to CL baselines, BriMA yields higher correlation and lower error, demonstrating robustness against both catastrophic forgetting and modality absence.
These results suggest that BriMA generalizes beyond AQA to other decision-critical multi-modal regression tasks with non-stationary modality availability.

\section{Additional Discussions}

Beyond the quantitative results, we further analyze several structural aspects of BriMA and clarify its design principles under dynamic multi-modal learning.

\begin{table}
\centering
\caption{Results on the MOSI dataset.}
    \label{tab:mosi_results}
    \resizebox{\linewidth}{!}{
    \begin{tabular}{rccccccccccc}
    \rowcolor{violet!10} JT-EMOE \cite{fang2025emoe} & \multicolumn{3}{c}{($\beta=0\%$)} & SRCC & 0.757 & MSE & 1.108 & RL2 & 3.078 
    \end{tabular}
    }
    \resizebox{\linewidth}{!}{
    \begin{tabular}{rccccccccccc}
    \toprule
    \multirow{2.5}{*}{Method}  & \multicolumn{3}{c}{$\beta=10\%$} & \multicolumn{3}{c}{$\beta=25\%$} & \multicolumn{3}{c}{$\beta=50\%$}\\
    \cmidrule(lr){2-4} \cmidrule(lr){5-7} \cmidrule(lr){8-10}
    & SRCC & MSE & RL2 & SRCC & MSE & RL2 & SRCC & MSE & RL2 \\
    \midrule
    \rowcolor{violet!10} JT-EMOE \cite{fang2025emoe} & 0.699 & 1.407 & 3.908 & 0.685 & 1.502 & 4.160 & 0.674 & 1.523 & 4.231 \\
    \midrule
    \rowcolor{teal!10} ST-EMOE \cite{fang2025emoe}   & 0.556 & 2.423 & 6.732 & 0.534 & 2.626 & 7.294 & 0.526 & 2.784 & 7.732\\
    \rowcolor{orange!10} EWC \cite{james2017ewc} & 0.536 & 2.581 & 7.170 & 0.529 & 2.683 & 7.454 & 0.522 & 2.618 & 7.273\\
    \rowcolor{orange!10} LwF \cite{li2017learning} & 0.587 & 1.729 & 4.803 & 0.502 & 1.943 & 5.347 & 0.177 & 3.134 & 8.706 \\
    \rowcolor{yellow!10} MER \cite{riemer2019learning} & 0.660 & 1.714 & 4.760 & 0.598 & 1.992 & 5.535 & 0.573 & 2.151 & 5.975 \\
    \rowcolor{yellow!10} DER++ \cite{buzzega2020dark} & 0.670 & 2.074 & 5.762 & 0.616 & 2.244 & 6.634 & 0.622 & 2.143 & 5.952 \\
    \rowcolor{yellow!10} NC-FSCIL \cite{yang2023neural} & 0.723 & 1.603 & 4.452 & 0.692 & 1.966 & 5.460 & 0.674 & 1.848 & 5.133  \\
    \rowcolor{yellow!10} SLCA \cite{zhang2023slca}  & 0.546 & 2.551 & 7.085 & 0.526 & 2.566 & 7.127 & 0.522 & 2.795 & 7.764 \\
    \rowcolor{yellow!10} FS-Aug \cite{li2024continual} &  0.570 & 2.380 & 6.612 & 0.536 & 2.836 & 7.878 & 0.530 & 2.853 & 7.926 \\
    \rowcolor{yellow!10} MAGR \cite{zhou2024magr} & 0.704 & 1.570 & 4.360 & 0.669 & 1.769 & 4.914 & 0.650 & 1.899 & 5.275 \\
    \rowcolor{yellow!10} ASAL \cite{zhou2025adaptive} & 0.696 & 1.646 & 4.574 & 0.673 & 1.663 & 4.619 & 0.612 & 1.929 & 5.358  \\
    \rowcolor{yellow!10} \textbf{BriMA (Ours)} &
\textbf{0.734} & \textbf{1.552} & \textbf{4.314} &
\textbf{0.700} & \textbf{1.787} & \textbf{4.964} &
\textbf{0.683} & \textbf{1.888} & \textbf{5.245} \\

    \bottomrule
    \end{tabular}
    }
\end{table}

\myPara{Dynamic Multi-modal Learning vs. Static Alignment.}
Prior modality-invariant methods (e.g., MINIMA \cite{ren2025minima}, X-Fi \cite{chen2024x}) primarily focus on static cross-modal alignment under fixed data distributions, aiming to learn modality-agnostic representations. In contrast, BriMA targets dynamic and continual multi-modal learning with non-stationary modality availability and evolving task distributions. This setting introduces additional challenges, as preserving score-sensitive geometry over time becomes critical under distribution shift and modality imbalance. Accordingly, the bridging space is designed for continual adaptation rather than enforcing global modality invariance.

\myPara{Assumption on Modality Availability.}
BriMA assumes modality availability is observable (e.g., via sensor status or data integrity checks), without requiring oracle knowledge of modality usefulness. Detecting unreliable modalities is a related but orthogonal problem. The residual reconstruction strategy predicts minimal corrections instead of full feature synthesis, operating within a locally smooth region of the loss landscape (see \cref{fig:cmp-a}). Imperfect modality signals therefore lead to bounded perturbations rather than amplified deviations, reducing sensitivity to routing noise.

\myPara{Robustness to Weak Informative Modalities.}
Not all modalities contribute equally to downstream regression tasks. BriMA adopts a conservative reconstruction principle: residual corrections are conditioned on exemplar priors rather than generative synthesis of complete features. When a modality is weakly informative or loosely correlated with the target, the learned residual naturally approaches zero. Combined with modality-aware replay guided by score-drift signals, this mitigates hallucination risks and stabilizes learning under imperfect modality relevance.

\myPara{Scalability with Respect to Modality Combinations.}
BriMA employs pattern-level conditioning embeddings to provide stable context under non-stationary modality availability. Although the number of possible missing patterns grows exponentially in theory, practical deployments involve limited modality sets and sparse observed configurations. Unseen patterns are projected into a shared low-dimensional embedding space, and residual bridging restricts corrections to minimal adjustments, avoiding brittle or random routing while maintaining scalability.

\myPara{Temporal Context and Task-Level Adaptation.}
BriMA is designed for task-level continual adaptation rather than explicit sequence-aware routing. While memory replay captures long-term distribution shifts, the framework does not explicitly model fine-grained temporal trajectories or stage-dependent modality availability. Incorporating temporal or stage-aware conditioning represents a complementary direction for future exploration. 
\end{document}